\documentclass{article} % For LaTeX2e
\usepackage{iclr2024_conference,times}

\usepackage{amsmath,amsfonts,bm}

\def\eqref#1{equation~\ref{#1}}
\def\1{\bm{1}}

\def\mA{{\bm{A}}}
\def\mB{{\bm{B}}}
\def\mC{{\bm{C}}}
\def\mD{{\bm{D}}}
\def\mE{{\bm{E}}}
\def\mF{{\bm{F}}}
\def\mG{{\bm{G}}}
\def\mH{{\bm{H}}}
\def\mI{{\bm{I}}}
\def\mJ{{\bm{J}}}
\def\mK{{\bm{K}}}
\def\mL{{\bm{L}}}
\def\mM{{\bm{M}}}
\def\mN{{\bm{N}}}
\def\mO{{\bm{O}}}
\def\mP{{\bm{P}}}
\def\mQ{{\bm{Q}}}
\def\mR{{\bm{R}}}
\def\mS{{\bm{S}}}
\def\mT{{\bm{T}}}
\def\mU{{\bm{U}}}
\def\mV{{\bm{V}}}
\def\mW{{\bm{W}}}
\def\mX{{\bm{X}}}
\def\mY{{\bm{Y}}}
\def\mZ{{\bm{Z}}}

\DeclareMathAlphabet{\mathsfit}{\encodingdefault}{\sfdefault}{m}{sl}
\SetMathAlphabet{\mathsfit}{bold}{\encodingdefault}{\sfdefault}{bx}{n}

\newcommand{\E}{\mathbb{E}}

\newcommand{\R}{\mathbb{R}}

\usepackage{hyperref}
\usepackage{url}
\usepackage{booktabs}       % professional-quality tables
\usepackage{amsfonts}       % blackboard math symbols
\usepackage{nicefrac}       % compact symbols for 1/2, etc.
\usepackage{microtype}      % microtypography
\usepackage{xcolor}         % colors
\usepackage{caption}
\usepackage{graphicx}
\usepackage{amsmath}
\usepackage{amssymb}
\usepackage{booktabs}
\usepackage{verbatim}
\usepackage{color}
\usepackage{float}
\usepackage{epsfig}
\usepackage{caption,subcaption}
\usepackage{soul}
\usepackage{colortbl}
\usepackage[symbol]{footmisc}
\def\eg{{\textit e.g.}}

\usepackage{multirow}
\usepackage{wrapfig}

\title{Exposing Text-Image Inconsistency Using Diffusion Models}

\author{Mingzhen Huang, Shan Jia, Zhou Zhou, Yan Ju, Jialing Cai, Siwei Lyu \\
{University at Buffalo, State University of New York} \\ 
}

\iclrfinalcopy % Uncomment for camera-ready version, but NOT for submission.
\begin{document}

\def\mA{\mathcal{A}}
\def\mB{\mathcal{B}}
\def\mC{\mathcal{C}}
\def\mD{\mathcal{D}}
\def\mE{\mathcal{E}}
\def\mF{\mathcal{F}}
\def\mG{\mathcal{G}}
\def\mH{\mathcal{H}}
\def\mI{\mathcal{I}}
\def\mJ{\mathcal{J}}
\def\mK{\mathcal{K}}
\def\mL{\mathcal{L}}
\def\mM{\mathcal{M}}
\def\mN{\mathcal{N}}
\def\mO{\mathcal{O}}
\def\mP{\mathcal{P}}
\def\mQ{\mathcal{Q}}
\def\mR{\mathcal{R}}
\def\mS{\mathcal{S}}
\def\mT{\mathcal{T}}
\def\mU{\mathcal{U}}
\def\mV{\mathcal{V}}
\def\mW{\mathcal{W}}
\def\mX{\mathcal{X}}
\def\mY{\mathcal{Y}}
\def\mZ{\mathcal{Z}} 

\def\bbN{\mathbb{N}} 
\def\bbR{\mathbb{R}} 
\def\bbP{\mathbb{P}} 
\def\bbQ{\mathbb{Q}} 
\def\bbE{\mathbb{E}}

\def\1n{\mathbf{1}_n}
\def\0{\mathbf{0}}
\def\1{\mathbf{1}}

\def\A{{\bf A}}
\def\B{{\bf B}}
\def\C{{\bf C}}
\def\D{{\bf D}}
\def\E{{\bf E}}
\def\F{{\bf F}}
\def\G{{\bf G}}
\def\H{{\bf H}}
\def\I{{\bf I}}
\def\J{{\bf J}}
\def\K{{\bf K}}
\def\L{{\bf L}}
\def\M{{\bf M}}
\def\N{{\bf N}}
\def\O{{\bf O}}
\def\P{{\bf P}}
\def\Q{{\bf Q}}
\def\R{{\bf R}}
\def\S{{\bf S}}
\def\T{{\bf T}}
\def\U{{\bf U}}
\def\V{{\bf V}}
\def\W{{\bf W}}
\def\X{{\bf X}}
\def\Y{{\bf Y}}
\def\Z{{\bf Z}}

\def\a{{\bf a}}
\def\b{{\bf b}}
\def\c{{\bf c}}
\def\d{{\bf d}}
\def\e{{\bf e}}
\def\f{{\bf f}}
\def\g{{\bf g}}
\def\h{{\bf h}}
\def\i{{\bf i}}
\def\j{{\bf j}}
\def\k{{\bf k}}
\def\l{{\bf l}}
\def\m{{\bf m}}
\def\n{{\bf n}}
\def\o{{\bf o}}
\def\p{{\bf p}}
\def\q{{\bf q}}
\def\r{{\bf r}}
\def\s{{\bf s}}
\def\t{{\bf t}}
\def\u{{\bf u}}
\def\v{{\bf v}}
\def\w{{\bf w}}
\def\x{{\bf x}}
\def\y{{\bf y}}
\def\z{{\bf z}}

\def\balpha{\mbox{\boldmath{$\alpha$}}}
\def\bbeta{\mbox{\boldmath{$\beta$}}}
\def\bdelta{\mbox{\boldmath{$\delta$}}}
\def\bgamma{\mbox{\boldmath{$\gamma$}}}
\def\blambda{\mbox{\boldmath{$\lambda$}}}
\def\bsigma{\mbox{\boldmath{$\sigma$}}}
\def\btheta{\mbox{\boldmath{$\theta$}}}
\def\bomega{\mbox{\boldmath{$\omega$}}}
\def\bxi{\mbox{\boldmath{$\xi$}}}
\def\bnu{\mbox{\boldmath{$\nu$}}}                                  
\def\bphi{\mbox{\boldmath{$\phi$}}}
\def\bmu{\mbox{\boldmath{$\mu$}}}

\def\bDelta{\mbox{\boldmath{$\Delta$}}}
\def\bOmega{\mbox{\boldmath{$\Omega$}}}
\def\bPhi{\mbox{\boldmath{$\Phi$}}}
\def\bLambda{\mbox{\boldmath{$\Lambda$}}}
\def\bSigma{\mbox{\boldmath{$\Sigma$}}}
\def\bGamma{\mbox{\boldmath{$\Gamma$}}}
                                  
\newcommand{\myprob}[1]{\mathop{\mathbb{P}}_{#1}}

\newcommand{\myexp}[1]{\mathop{\mathbb{E}}_{#1}}

\newcommand{\mydelta}[1]{1_{#1}}

\newcommand{\myminimum}[1]{\mathop{\textrm{minimum}}_{#1}}
\newcommand{\mymaximum}[1]{\mathop{\textrm{maximum}}_{#1}}    
\newcommand{\mymin}[1]{\mathop{\textrm{minimize}}_{#1}}
\newcommand{\mymax}[1]{\mathop{\textrm{maximize}}_{#1}}
\newcommand{\mymins}[1]{\mathop{\textrm{min.}}_{#1}}
\newcommand{\mymaxs}[1]{\mathop{\textrm{max.}}_{#1}}  
\newcommand{\myargmin}[1]{\mathop{\textrm{argmin}}_{#1}} 
\newcommand{\myargmax}[1]{\mathop{\textrm{argmax}}_{#1}} 
\newcommand{\myst}{\textrm{s.t. }}

\newcommand{\denselist}{\itemsep -1pt}
\newcommand{\sparselist}{\itemsep 1pt}

\definecolor{pink}{rgb}{0.9,0.5,0.5}
\definecolor{purple}{rgb}{0.5, 0.4, 0.8}   
\definecolor{gray}{rgb}{0.3, 0.3, 0.3}
\definecolor{mygreen}{rgb}{0.2, 0.6, 0.2}

\newcommand{\cyan}[1]{\textcolor{cyan}{#1}}
\newcommand{\red}[1]{\textcolor{red}{#1}}  
\newcommand{\blue}[1]{\textcolor{blue}{#1}}
\newcommand{\magenta}[1]{\textcolor{magenta}{#1}}
\newcommand{\pink}[1]{\textcolor{pink}{#1}}
\newcommand{\green}[1]{\textcolor{green}{#1}} 
\newcommand{\gray}[1]{\textcolor{gray}{#1}}    
\newcommand{\mygreen}[1]{\textcolor{mygreen}{#1}}    
\newcommand{\purple}[1]{\textcolor{purple}{#1}}       

\definecolor{greena}{rgb}{0.4, 0.5, 0.1}
\newcommand{\greena}[1]{\textcolor{greena}{#1}}

\definecolor{bluea}{rgb}{0, 0.4, 0.6}
\newcommand{\bluea}[1]{\textcolor{bluea}{#1}}
\definecolor{reda}{rgb}{0.6, 0.2, 0.1}
\newcommand{\reda}[1]{\textcolor{reda}{#1}}

\def\changemargin#1#2{\list{}{\rightmargin#2\leftmargin#1}\item[]}
\let\endchangemargin=\endlist
                                               
\newcommand{\cm}[1]{}

\newcommand{\mhoai}[1]{{\color{blue}{[MH: #1]}}}

\newcommand{\mtodo}[1]{{\color{red}$\blacksquare$\textbf{[TODO: #1]}}}
\newcommand{\myheading}[1]{\vspace{0 ex}\noindent \textbf{#1}}
\newcommand{\htimesw}[2]{\mbox{$#1$$\times$$#2$}}

% The following are useful for creating homework or exams

\newif\ifshowsolution
%\showsolutionfalse
\showsolutiontrue

\ifshowsolution  
\newcommand{\Comment}[1]{\paragraph{\bf $\bigstar $ COMMENT:} {\sf #1} \bigskip}
\newcommand{\Solution}[2]{\paragraph{\bf $\bigstar $ SOLUTION:} {\sf #2} }
\newcommand{\Mistake}[2]{\paragraph{\bf $\blacksquare$ COMMON MISTAKE #1:} {\sf #2} \bigskip}
\else
\newcommand{\Solution}[2]{\vspace{#1}}
\fi

\newcommand{\truefalse}{
\begin{enumerate}
	\item True
	\item False
\end{enumerate}
}

\newcommand{\yesno}{
\begin{enumerate}
	\item Yes
	\item No
\end{enumerate}
}

\newcommand{\Sref}[1]{Sec.~\ref{#1}}
\newcommand{\Eref}[1]{Eq.~(\ref{#1})}
\newcommand{\Fref}[1]{Fig.~\ref{#1}}
\newcommand{\Tref}[1]{Table~\ref{#1}}

\definecolor{mygray}{gray}{.9}
\makeatletter
\newcommand{\thickhline}{%
    \noalign {\ifnum 0=`}\fi \hrule height 0.8pt
    \futurelet \reserved@a \@xhline
}

%Define a reference depth. 
%You can choose either relative or absolute.
%--------------------------
\newlength{\DepthReference}
\settodepth{\DepthReference}{g}%relative to a depth of a letter.
%\setlength{\DepthReference}{6pt}%absolute value.

%Define a reference Height. 
%You can choose either relative or absolute.
%--------------------------
\newlength{\HeightReference}
\settoheight{\HeightReference}{T}
%\setlength{\HeightReference}{6pt}

%--------------------------
\newlength{\Width}%

\newcommand{\MyColorBox}[2][red]%
{%
    \settowidth{\Width}{#2}%
    \colorbox{#1}%
    {%      
        \raisebox{-\DepthReference}%
        {%
                \parbox[b][\HeightReference\DepthReference][c]{\Width}{\centering#2}%
        }%
    }%
}
\maketitle
\begin{abstract}
In the battle against widespread online misinformation, a growing problem is text-image inconsistency, where images are misleadingly paired with texts with different intent or meaning. Existing classification-based methods for text-image inconsistency can identify contextual inconsistencies but fail to provide explainable justifications for their decisions that humans can understand. Although more nuanced, human evaluation is impractical at scale and susceptible to errors. To address these limitations, this study introduces D-TIIL (Diffusion-based Text-Image Inconsistency Localization), which employs text-to-image diffusion models to localize semantic inconsistencies in text and image pairs. These models, trained on large-scale datasets act as ``omniscient" agents that filter out irrelevant information and incorporate background knowledge to identify inconsistencies. In addition, D-TIIL uses text embeddings and modified image regions to visualize these inconsistencies. To evaluate D-TIIL's efficacy, we introduce a new TIIL dataset containing 14K consistent and inconsistent text-image pairs. Unlike existing datasets, TIIL enables assessment at the level of individual words and image regions and is carefully designed to represent various inconsistencies. D-TIIL offers a scalable and evidence-based approach to identifying and localizing text-image inconsistency, providing a robust framework for future research combating misinformation. Please refer \href{https://mingzhenhuang.com/projects/InconsisDet.html}{Project Page} for source code and dataset.
\end{abstract}

%%%%%%%%%%%%%%%%%%%%%%%%%%%%%%%%%%%%%%%%%%%%%%%%%%%%%%%%%%%%

\section{Introduction}

The widespread online misinformation~\citep{ali2020combatting} has become the bane of the Internet and social media. One simple means to create misinformation is to juxtapose images with texts that do not accurately reflect the image's original meaning or intention. In this work, we term this type of misinformation as text-image inconsistency~\citep{lee2019image, tan2020detecting, zeng2023correcting}. % \lsw{add real-world examples}.  
Exposing text-image inconsistency has become an important task in combating misinformation. Text-image inconsistency can be solved with binary classification, as in the recent works of MAIM~\citep{jaiswal2017multimedia}, COSMOS~\citep{aneja2021cosmos},  NewsCLIPpings~\citep{luo2021newsclippings}, and CCN~\citep{abdelnabi2022open}, which classifies an input text-image pair as contextual consistent or inconsistent. Although showing good classification performance on benchmark datasets, the classification-based methods output only the predicted categories, with little or no evidence to support the decision. On the other hand, humans often spot text-image inconsistency by locating image regions corresponding to objects or scenes inconsistent with the textual description, using knowledge of the world. In addition, humans often prefer more visual evidence of semantic inconsistency, as when a mis-contextualized text-image pair is explained to another human. However, when we need to analyze many text-image pairs, relying on human inspection is costly, time-consuming, and prone to mistakes and errors \citep{molina2021fake}. Our work aims to make this process automatic so it can scale up. 

Specifically, we aim to address two challenges in localizing text-image inconsistency intrinsic to the complex nature of semantic contents across the two modalities. First, there is unrelated information in text and images irrelevant to their semantic consistency. This is usually information only represented in one modality but not in the other. %(\ie, the non-overlapping portion of the Venn diagram illustration in \Fref{fig:teaser} (middle)). 
Unrelated information in one modality will not have a counterpart in the other, but it is not the cause of semantic mismatch and cannot be accounted for inconsistency. Furthermore, many cases of inconsistency are hard to identify due to limited background knowledge of humans or algorithms. For instance, to someone who is unaware that dolphins are mammals, a text stating ``a school of fish swimming in the ocean'' might seem consistent with an image showing dolphins swimming. Such missing information can be overcome by using a more knowledgeable human or incorporating background knowledge into the algorithm. 

\begin{wrapfigure}{!t}{0.5\textwidth}
\begin{center}
\vspace{-2em}
\includegraphics[width=0.48\textwidth]
{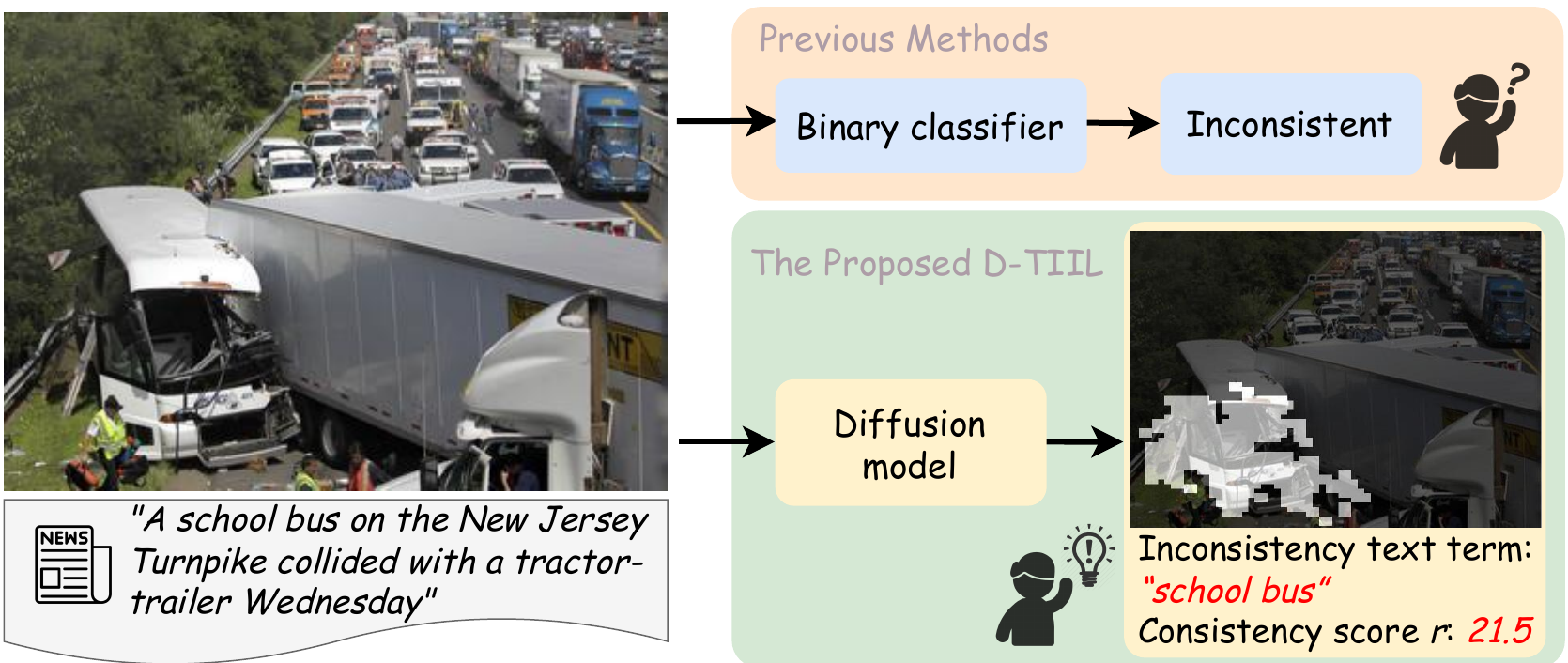}%Fig1_last.pdf}
\vspace{-1em}
\caption{\small \em Exposing text-image inconsistency based on previous methods and our method. Instead of employing a binary classification model, D-TIIL offers interpretable evidence by localizing word- and pixel-level inconsistencies and quantifying them through a consistency score.}%{\bf Left}: Examples in existing text-image inconsistency datasets, which use random swap to create inconsistent text-image pairs that may contradict with human intuitions. {\bf Right}: Examples in our TIIL dataset with pixel- and word-level inconsistency labels. The proposed D-TIIL method predicts the inconsistency region, localizes the inconsistency text term, and outputs a consistency score.} 
\label{fig:teaser}
\vspace{-2em}
\end{center}
\end{wrapfigure}

Both challenges are addressed in our work by leveraging the text-to-image diffusion models. Text-to-image diffusion models trained on large-scale datasets, such as DALL-E2~\citep{DALLE2ramesh2022hiera}, Stable Diffusion~\citep{rombach2022high}, Glide~\citep{nichol2021glide}, and GLIGEN~\citep{li2023gligen}, can generate realistic images with consistent semantic content in the text prompts. We can regard these large-scale text-to-image diffusion models as an ``omniscient'' agent with extensive background knowledge about any subject matter. Taking advantage of this knowledge representation, we describe a new method that {\em locates} the semantically inconsistent image regions and words, which is termed as {\it \underline{D}iffusion-based \underline{T}ext-\underline{I}mage \underline{I}nconsistency \underline{L}ocalization} (D-TIIL). D-TIIL proposes two different alignment steps that iteratively align the image and update text (in the form of vectorized embeddings) with diffusion models to (i) filter out the irrelevant semantic information in the text-image pairs and (ii) incorporate background knowledge that is not obvious in their shared semantic scope. The first alignment step employs diffusion models to generate aligned text embeddings from the input image. This is to filter out implicit semantics and establish textual consistency. The second alignment operation focuses on denoising the input text to be more relevant to the input image. To be more specific, we modify the input image based on the original input text to achieve semantic consistency, and then produce aligned text embeddings from this edited image. These two alignment steps yield modified yet knowledge-shared text embeddings from both the input image and text, making it easier to identify semantic inconsistencies. This approach sets our method apart from previous ones, as depicted in \Fref{fig:teaser}.%In the first alignment stage, the text embedding is aligned with the input image by diffusion models and filtered implicit semantics in the origin embedding, the aligned text is exactly consistent with the image; in the second， we edit the input image to be consistent with the input text and further align the input text with the edit image to filter irrelevant semantics. Different from previous methods, our proposed D-TIIL not only detects the inconsistency but also localizes the word- and pixel-level inconsistencies as illustrated in \Fref{fig:teaser}.}%[need more explanation]

% \begin{wrapfigure}{r}{0.5\textwidth}
%   \begin{center}
%     \includegraphics[width=0.48\textwidth]{Figure/Fig1.png}
%   \end{center}
%   \caption{Birds}
% \end{wrapfigure}

Existing datasets \citep{luo2021newsclippings, aneja2021cosmos} do not provide evidence of inconsistency at the level of image regions and words that can be used to evaluate D-TIIL. To this end, we create a new {\it \underline{T}ext-\underline{I}mage \underline{I}nconsistency \underline{L}ocalization} (TIIL) dataset, which contains $14$K text-image pairs. %TIIL includes 3.5K captions and images in the same news story or social media post from the VisualNews~\citep{liu2020visual} dataset as consistent pairs. 
Existing datasets construct inconsistent text-image pairs by randomly swapping texts \citep{luo2021newsclippings} or external search for the similar text of the swapped texts \citep{aneja2021cosmos} to match with the original image. These methods can create inconsistent text-image pairs that either conflict with human intuitions or are totally irrelevant (see more analysis in Sec.\ref{sec:data}). %(\ie, the examples shown in \Fref{fig:teaser} (left)). 
Differently, TIIL constructs inconsistent pairs by changing words in the text, and/or editing regions in the image (\eg, changing objects, attributes, or scene-texts)\footnote{TIIL has mixed both original and edited images; one cannot simply rely on forensic methods that identifying image editing to expose inconsistent pairs.}. The edited words and regions are manually selected. The image editing is made with the text-to-image diffusion models. Furthermore, all inconsistent text-image pairs undergo meticulous manual curating to reduce ambiguities in interpretation. 

The main contributions of our work can be summarized as follows:
\vspace{-0.3cm}
\begin{itemize}
\item  We develop a new method, D-TIIL, that leverages text-to-image diffusion models to expose text-image inconsistency with the location of inconsistent image regions and words;
\vspace{-0.1cm}
\item Text-to-image diffusion models are used as a latent and joint representation of the semantic contents of the text and image, where we can align text and image to discount irrelevant information, and we use the broad coverage of knowledge in the diffusion models to incorporate more extensive background; 
\vspace{-0.1cm}
\item We introduce a new dataset, TIIL, built on real-world image-text pairs from the Visual News dataset, for evaluating text-image inconsistency localization with pixel-level and word-level inconsistency annotations.
\end{itemize}

%Similarly, the same task also hinges on the implicit focus of the observer. A text saying ``Person A is at JFK airport'' might align with an image of an unidentified person for someone focused on the identity of the person but may seem inconsistent for someone more interested in the person's exact location. 

%This is performed in two semantic ``alignment'' steps of the LIM to gradually remove distracting and implicit information.
%Since the subjective nature of the problem makes it difficult to have a consensus on the semantic consistency/inconsistency across different evaluators, we formulate it as a regression and localization problem, which outputs a score in the range of $[0,100]$ to indicate the level of semantic consistency between $T$ and $I$, and the words/image regions that are mostly attributed to the inconsistency. 

%%%%%%%%%%%%%%%%%%%%%%%%%%%%%%%%%%%%%%%%%%%%%%%%%%%%%%%%%%%%
\vspace{-0.35cm}
\section{Backgrounds} \vspace{-0.2cm}
\subsection{Related Works}
  \vspace{-0.23cm}
% review LIMs

% review existing inconsistency methods
Text-image inconsistency detection has been the focus of several recent works~\citep{abdelnabi2022open, luo2021newsclippings, qi2021improving, aneja2021cosmos, abdelnabi2022open}. There are many methods~\citep{zlatkova2019fact,abdelnabi2022open, popat2018declare} using the reverse image search function provided by the Search Engine (\eg, Google Image Search) to gather textual evidence (articles or captions) from the Internet or external fact-checking sources on the Internet (e.g., Politifact\footnote{Politifact: https://www.factcheck.org} and Factcheck\footnote{Factcheck: https://www.politifact.com}). The resulting text is then compared with the original text in an embedding space such as BERT \citep{kenton2019bert} to determine their consistency. 
% These methods require manual examination of image-text pairs using web-based evidence that is both time-consuming and requires considerable reasoning effort. 
%Abdelnabi \etal \citep{abdelnabi2022open} proposed an automated image-text fact-checking method to retrieve evidence from the Internet. it first 
Although straightforward and intuitive, these methods rely solely on the results from the reverse image search and are limited by irrelevant or contradicting texts found online. Other inconsistency detection methods explore joint semantic representations of texts and images. For example, \cite{khattar2019mvae} designs a multimodal variational autoencoder for learning the relationship between textual and visual information for fake news detection. \cite{mccrae2021multi} detect semantic inconsistencies in video-caption posts by comparing visual features obtained from multiple video-understanding networks and textual features derived from the BERT~\citep{kenton2019bert} language model. \cite{aneja2021cosmos} employs a self-supervised training strategy to learn correlation from an image and two captions from different sources. 
More recently, neural vision-language models originally designed for other vision-language tasks (e.g., VQA, image-text retrieval) have also been applied to text-image inconsistency detection. For instance, CLIP~\citep{radford2021learning} is used in~\cite{luo2021newsclippings} and the VinVL~\citep{zhang2021vinvl} model introduced in~\cite{huang2022text}.

%While these image inconsistency detection methods have achieved automatic and promising results in their benchmarks, they have all concentrated on predicting the binary consistency of given image-text pairs and failed to provide interpretable evidence to explain the detection outcomes. The absence of localization and clarification for the identified inconsistency reduces the persuasiveness of the findings, consequently restricting its practicality in real-world situations.

% By processing and analyzing both textual and visual information simultaneously, these models can perform a wide range of tasks, such as image captioning~\cite{zhu2022exploring}, visual question answering~\cite{hu2023tifa}, and image-text matching~\cite{sheynin2022knn}, with remarkable performance. 
% Notable examples of LIMs include OpenAI's CLIP (Contrastive Language-Image Pretraining)~\cite{CLIP_radford2021} and Google's ALIGN (A Large-scale ImaGe and Noisy-Text Embedding)~\cite{hu2022scaling}. 

 % MAIM & 240K & 120K & 0 & Random swap   & $\times$\\ % ~\cite{jaiswal2017multimedia}
 %   MEIR & 58K &29K &  0 & Random swap   & $\times$\\ % ~\cite{sabir2018deep}
 %   NewsCLIPpings  & 988K &494K &  0 & Auto retrieval   & $\times$\\ % ~\cite{luo2021newsclippings}
 %   FacebookPost & 4K &2K &  0 & Random swap   & $\times$ \\ % ~\cite{mccrae2022multi}
 %   COSMOS   & 200K &850 &  1.7K & Manual    & $\times$ \\ \hline % ~\cite{aneja2021cosmos}
 
Several multi-modal datasets exist for image inconsistency detection. MAIM~\citep{jaiswal2017multimedia}, MEIR~\citep{sabir2018deep}, FacebookPost~\citep{mccrae2022multi} and COSMOS~\citep{aneja2021cosmos} are formed by swapping the original caption of an image with randomly selected ones to create inconsistent image-text pairs. The NewsCLIPpings dataset~\citep{luo2021newsclippings} utilizes CLIP~\citep{CLIP_radford2021} as a retrieve model to swap similar captions in the Visual News dataset~\citep{liu2020visual}. The main problem with these datasets is that the semantic relations among the labeled consistent or inconsistent pairs are not precise~\citep{huang2022text}, making them less reliable to be used as training data for text-image inconsistency detection methods. These methods and datasets do not report words or image regions that cause the inconsistency.
% For example, the research in \citep{huang2022text} uncovered that certain image and text pairs labeled as inconsistent exhibit high consistency. This finding complicates the task of automatic inconsistency detection methods, making it challenging for them to produce reliable predictions.
% notable challenge in these datasets lies in the ambiguous nature of the inconsistency and consistency definitions, which can often be a moving target. 
  \vspace{-0.13cm}
\subsection{Text-to-image Diffusion Models}
  \vspace{-0.23cm}
The diffusion model~\citep{ho2020denoising} %, recognized as a prominent LIM, 
has recently attained state-of-the-art performance in the field of text-to-image generation~\citep{DALLE2ramesh2022hiera, Imagen_saharia2022phot, rombach2022high}. As a category of likelihood-based models~\citep{nichol2021improved}, diffusion models perturb the data by progressively introducing Gaussian noise to the input data and train to restore the original data by reversing this noise application process. 
The key idea involves initialization with $\mathbf{x}_T \sim \mathcal{N}(0, \mathbf{I})$, which represents an iteratively noised image derived from the input image $\mathbf{x}_0$. At each timestep $t \in [0, T]$, the sample $\mathbf{x}_t$ is computed as $\mathbf{x}_t = \sqrt{\alpha_t} \mathbf{x}_0 + \sqrt{1-\alpha_t} \boldsymbol{\epsilon}_t$, where $\alpha_t \in (0, 1]$ defines the level of noise, and $\boldsymbol{\epsilon}_t \sim \mathcal{N}(0, \mathbf{I})$ represents the sampled noise.
Ultimately, the distribution of $\mathbf{x}_T$ approaches a Gaussian distribution. Diffusion models then iteratively reverse this process and denoise $\mathbf{x}_T$ to generate images given a text conditioning $c$ by minimizing a simple denoising objective:
\begin{align}
\vspace{-0.15cm}
    \mathcal{L}=\mathbb{E}_{t, \mathbf{x}_0,  \boldsymbol{\epsilon}}\left\|\boldsymbol{\epsilon}-\epsilon_\theta\left(\mathbf{x}_t, t, c\right)\right\|_2^2
\vspace{-0.15cm}
\end{align}
where $\epsilon_\theta$ is an UNet~\citep{ronneberger2015u} noise estimator that predicts $\epsilon_t$ from $\mathbf{x}_t$. Diffusion models have been widely used in various downstream applications, including image editing~\citep{couairon2023diffedit, kawar2023imagic} where a text-conditional diffusion model can be generalized for learning conditional distributions. When provided with different text conditionings, the model generates different noise estimates. Notably, the variation in noise across spatial locations reflects the semantic distinctions between the corresponding text conditions in the image space. This inspiration motivates us to utilize diffusion models for representing and exposing semantic inconsistencies in image inconsistency.

%%%%%%%%%%%%%%%%%%%%%%%%%%%%%%%%%%%%%%%%%%%%%%%%%%%%%%%%%%%%
  \vspace{-0.13cm}
  \section{Method}
  \vspace{-0.13cm}
\begin{figure}[t]
  \centering
  \includegraphics[width=1.0\linewidth]{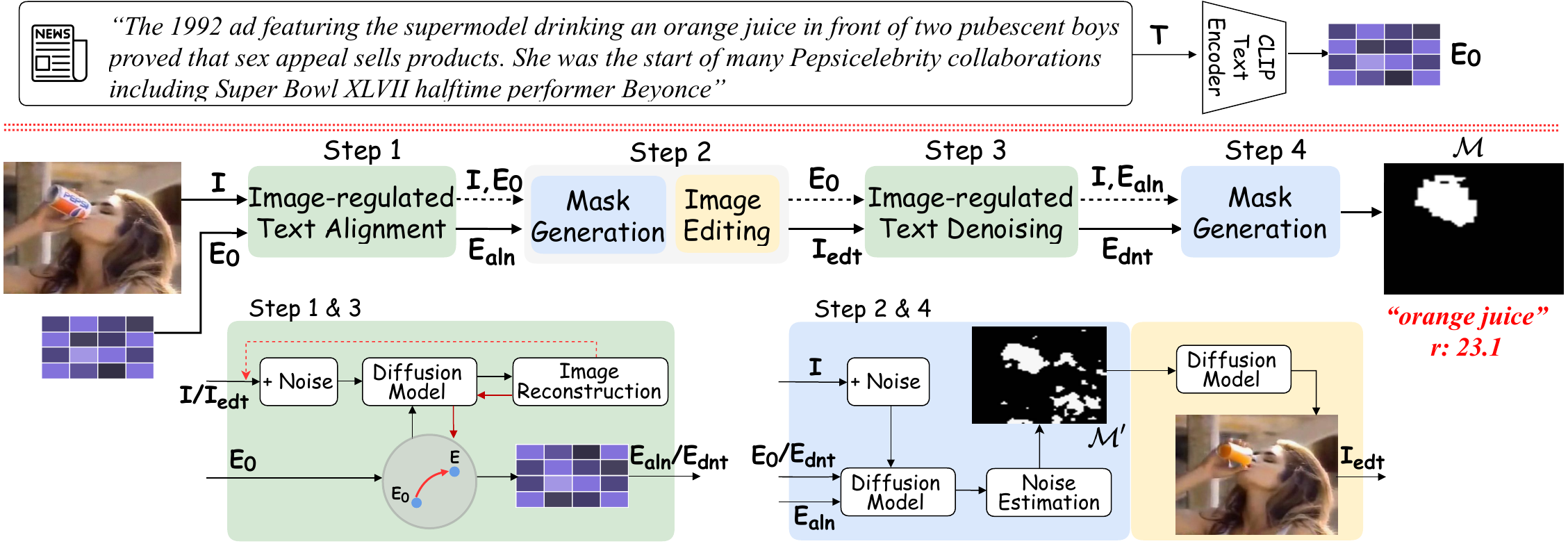}
\vspace{-1.5em}
  \caption{\small \em The overall pipeline of D-TIIL. See texts for details.}
  \label{fig:pipeline}
  \vspace{-0.5cm}
\end{figure}

\begin{figure}[t]
  \centering
  \includegraphics[width=1.0\linewidth]{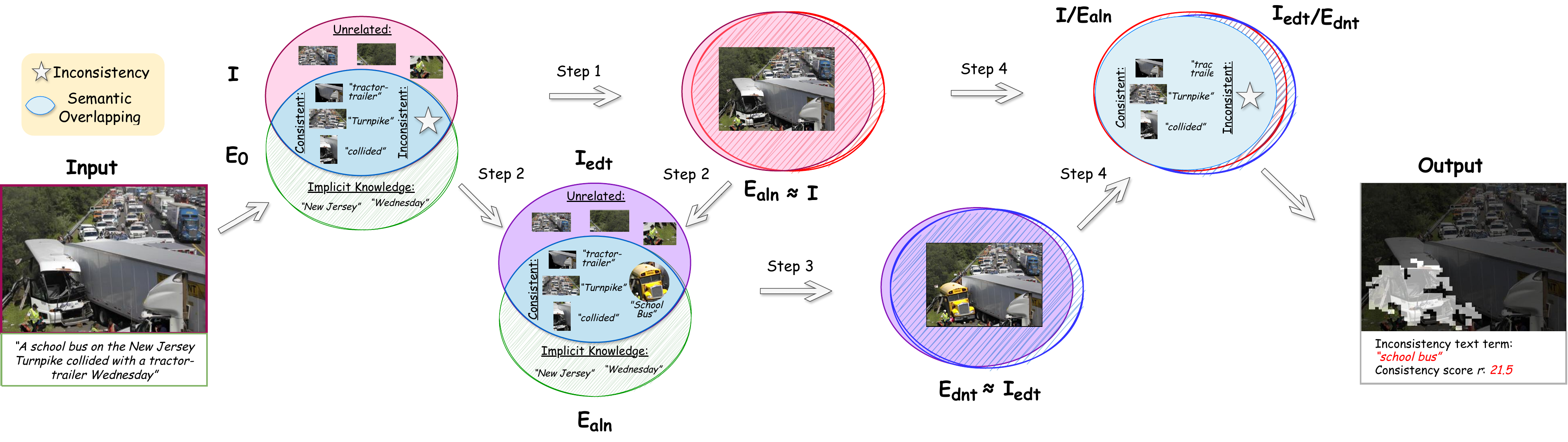}
\vspace{-0.61cm}
      \caption{\em \small The main process of D-TIIL is illustrated conceptually with Venn diagrams, where the semantic contents of text and image are represented as two circles. The four steps gradually align the semantic contents to facilitate exposure of inconsistency: given an initial image-text pair $(I, E_0)$, the proposed method first produces a text embedding $\E_{aln}$ that is aligned with $I$, and then an edited image $I_{edt}$ to filter the inconsistency. In Step 3, the model optimizes $E_0$ from the $I_{edt}$ to obtain a $E_{dnt}$ which is aligned with $I_{edt}$. Finally, in Step 4, the model produces the inconsistency mask from well-aligned pair $(I,E_{aln}, E_{dnt})$.}
  \label{fig:motivation}
  \vspace{-0.65cm}
\end{figure}

This section describes the proposed D-TIIL model in detail. The input to D-TIIL is a pair of image ${I}$ and text ${T}$, from which D-TIIL outputs the image region (as a binary mask $\mM$) and words in the text that exhibit semantic inconsistency. %\lsw{make Intro consistent} 
In addition, a consistency score ${r} \in [0,100]$, with $0$ being maximum inconsistent and $100$ being completely consistent, is also obtained based on the localization results. The overall process of D-TIIL is illustrated in \Fref{fig:pipeline} with four distinct steps. In those four steps, we iteratively align image-text semantic to produce final output as shown in \Fref{fig:motivation}

\myheading{Step 1: Align Text Embedding to Input Image.} 
We first obtain the CLIP~\citep{CLIP_radford2021} text embedding %\lsw{notation change} 
${E}_0$ of the input text ${T}$. Using a pre-trained Stable Diffusion model ${\mathcal G}$~\citep{rombach2022high}, we find another text embedding ${E}_{aln}$ that better aligns with the semantic content of the input image ${I}$ as shown in \Fref{fig:motivation} Step 1. Model ${\mathcal G}$ takes as input the text embedding $E_0$ and noised image $\mathbf{x}_T$ to generate an image, ${\mathcal G}(\mathbf{x}_T;E_0)$. It simulates a diffusion process that begins with the input noise and generates an image that exhibits similar semantic content to the text embedding $E_0$. The image-regulated text alignment in D-TIIL is to solve the following optimization problem:
\begin{align} \label{eq:constrain}
%\vspace{-0.1cm}
    {E}_{aln} = \arg\!\min_E \|I - {\mathcal G}(\mathbf{x}_T; E)\|_2 \quad s.t., \|E-E_0\|_F\le \gamma
%\vspace{-0.1cm}
\end{align}
where the learnable $E$ is initialized from $E_0$,  and $\gamma >0$ is a small constant determined as a hyper-parameter. $\|\cdot\|_2$ and $\|\cdot\|_F$ are the vector $\ell_2$ norm and matrix Frobenius norm, respectively. The constraint is used to control the deviation from the original embedding. This optimization problem can be solved using the gradient computation of $\mathcal{G}$ iteratively. The obtained ${E}_{aln}$ is semantically closer to the image when the original text $T$ has inconsistencies and distracting semantic information.

% Moreover, we synthesize an image from the input text and original image, modifying local content to align the image with the text. Subsequently, a new text embedding is created from this synthetic image to represent one that agrees with the original input text. This step ensures that the semantic content of the input text is projected into the image's relevant semantic subspace. Secondly, in the text-conditioned image denoising stage, we use the two embeddings from the first step to process the input image via diffusion. We generate two noise maps corresponding to the two embeddings from the input image. Their difference serves as the output inconsistency map, from which we derive an inconsistency score. \Fref{fig:motivation} provides a visual representation of our method's overall process.

\myheading{Step 2: Text-guided Image Editing.} \label{sec:noise_mask}
Next, we generate an edited image, ${I}_{edt}$, which is in alignment with the original text embedding, ${E}_0$. This is intended to materialize the original text, $T$, within a visual context, thereby minimizing the presence of extraneous and implicit data. It subsequently acts as the benchmark for estimating inconsistency. D-TIIL transposes the semantics of ${E}_0$ into the image space, with both ${E}_0$ and the aligned text embedding ${E}_{aln}$ guiding the editing process. Specifically, we introduce a noised version of image $I$ as the input and leverage the UNet architecture from the Stable Diffusion to derive two noise estimations, each corresponding to ${E_0}$ as target and ${E}_{aln}$ as reference. By examining the difference between these two noise estimates in the spatial domain, we can identify regions in image $I$ that are most prone to modifications due to the shift in conditioning text from ${E}_{aln}$ to ${E}_0$. This disparity is then transformed into a binary mask, denoted as $\mM^{\prime}$, by normalizing values within the [0, 1] range and subsequently employing a thresholding operation. After obtaining $\mM^{\prime}$, we use the Diffusion Inpainting model~\citep{lugmayr2022repaint} to yield the edited image ${I}_{edt}$, guided by the target text embedding ${E_0}$ and mask $\mM^{\prime}$. The process effectively transmutes the textual embedding ${E_0}$ into a visual equivalent, purging any distracting or implicit details, as shown in \Fref{fig:motivation} Step 2.

\myheading{Step 3: Align Text Embedding to Edited Image.} While the binary mask $\mM^{\prime}$ captures the regions of inconsistency between the input image $I$ and text embedding $E_0$, it may still include regions that are not directly related to semantic consistency, such as the objects or scenes in $I$ and $I_{edt}$ that does not correspond to a verbal description in $T$.
%The binary mask $\mM^{\prime}$ may still include regions that are not relevant to the semantic consistency, corresponding to objects or scenes in $I_0$ and $I_{edt}$ that does not have a verbal counterpart in $T_0$. 
This is a form of unrelated information that we use another round of operation involving the diffusion model $\mathcal{G}$ to reduce. Specifically, we formulate another optimization problem ${E}_{dnt} = \arg\!\min_E \|I_{edt} - {\mathcal G}(\mathbf{x}_T^{edt}; E)\|_2$, s.t., $\|E-E_0\|_F\le \gamma$ where $\mathbf{x}_T^{edt}$ is a noised image of $I_{edt}$. Compared with input text embedding ${E}_0$, the aligned text embedding ${E}_{dnt}$ includes extra implicit information from the images and excludes additional implicit information that only appears in the text as shown in \Fref{fig:motivation} Step 3 where we refer this operation as image-regulated text denoising.

\myheading{Step 4: Inconsistency Localization and Detection.} 
From the two text embeddings that are more closely aligned, $E_{aln}$ and $E_{dnt}$, we generate the difference in %noise patterns 
visual domain denoted as $\mM$. This is done by repeating the mask generation process outlined in Step 2. The $\mM$ represents the pixel-level inconsistent region within the image. To detect the corresponding inconsistent words, we compare the edited image ${I}_{edt}$ with the inconsistency mask $\mM$. Specifically, we leverage the CLIP image encoder to obtain the image embedding from the image ${I}_{edt}$. Next, we derive tokenized words from the rows in $E_0$ that exhibit the greatest cosine similarity with the image embedding, such as the example of ``an orange juice'' depicted in \Fref{fig:pipeline}. 
To further generate a regression score that quantifies the degree of image-text inconsistency, we extract the CLIP image embedding from the masked input image $I$ using $\mM$. We then compute the cosine similarity score between this CLIP image embedding of the masked image and the input text embedding $E_0$ as the consistency score. The resulting score is rescaled to the range of [0, 100], serving as the final consistent score $r$ of our D-TIIL model.
% The detected inconsist $\mM$
%%%%%%%%%%%%%%%%%%%%%%%%%%%%%%%%%%%%%%%%%%%%%%%%%%%%%%%%%%%%

  \vspace{-0.13cm}
\section{TIIL Dataset}  \label{sec:data}
  \vspace{-0.13cm}

We also construct TIIL as a more carefully curated dataset for text-image inconsistency analysis. Our approach to dataset creation is different from those of existing datasets that use randomly or algorithmically identified pairs as mismatched image-text pairs~\citep{jaiswal2017multimedia, sabir2018deep, aneja2021cosmos}. We leverage state-of-the-art text-to-image diffusion models to design text-guided inconsistencies within images and human annotation to improve the relevance of the inconsistent pairs.   

\begin{figure}[t]
  \centering
  \includegraphics[width=0.8\linewidth]{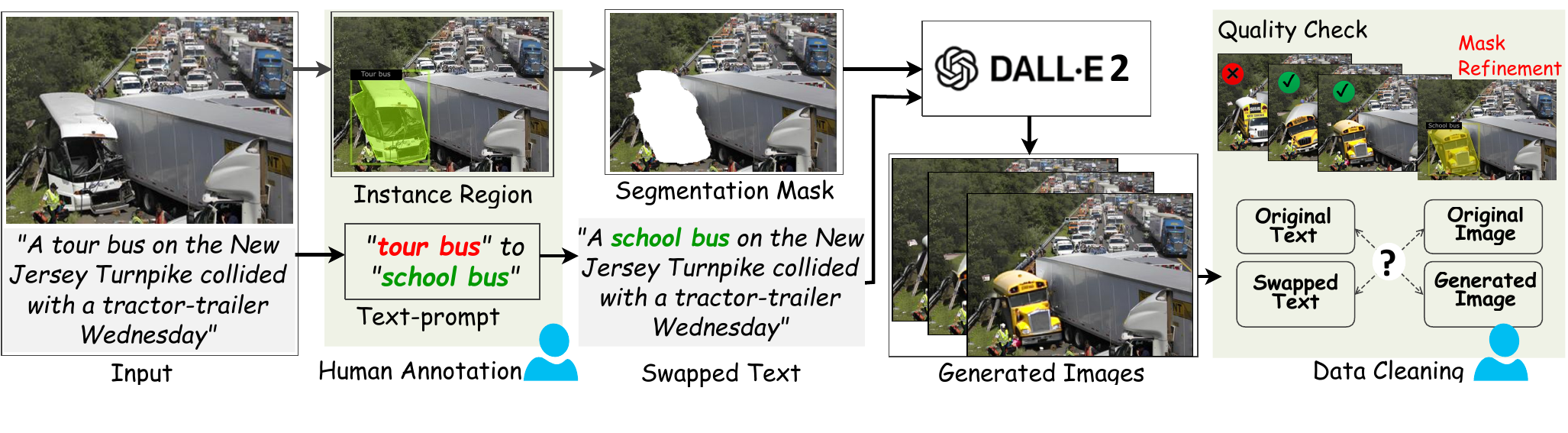}
  \vspace{-0.65cm}
  \caption{\em \small Pipeline depicting the generation and annotation process of the proposed TIIL dataset.} 
  \vspace{-0.33cm}
  \label{fig:dataset}
\end{figure}

\myheading{Data Generation.} Our methodology starts with a real-world image-text pair, $\left\{{I}, {T}\right\}$, obtained from the Visual News dataset~\citep{liu2020visual}, which offers a rich variety of news topics and sources, providing us with diverse real-world news data. The first step in our process involves creating an edited image, ${I}_e$. This is achieved by modifying a specific region in ${I}$ using an altered text ${T}_m$ by human annotators, where the text prompt corresponding to the object is replaced with a different generation term. The region to be manipulated is manually selected. Through this procedure, we can generate two consistent image-text pairs, namely $\left\{{I}, {T}\right\}$ and $\left\{{I}_e, {T}_m\right\}$, as well as two inconsistency pairs formed by $\left\{{I}, {T}_m\right\}$ and $\left\{{I}_e, {T}\right\}$. The DALL-E2 model~\citep{DALLE2ramesh2022hiera}, known for its capacity to generate images within specific regions based on text prompts, is leveraged to create this dataset. \Tref{tab:dalle_consis} demonstrates that the DALL-E2 model outperforms real-world image-text pairs in terms of CLIP similarity scores, indicating its superior ability to capture multi-modal connections. \Fref{fig:dataset} illustrates the entire generation pipeline of our TIIL dataset. The process starts with a real image-text pair. Human annotators then identify the corresponding visual region and textual term. Subsequently, these annotators provide a different text prompt to replace the chosen term, thereby creating an inconsistency with the selected object region. The mask of the selected object region and the swapped text with the new prompt are then fed into the DALL-E2 model for image generation. In the final step, human annotators carefully assess the quality of the generated images, evaluate the image-text inconsistency, and refine the region mask to provide the ground truth for the pixel-level inconsistency mask. The TIIL dataset consists of approximately 14K image-text pairs, encompassing a total of 7,138 inconsistencies and 7,101 consistent pairs. %Those inconsistencies are generated from 2,101 news image-text pairs from Visual News~\citep{liu2020visual}. 
All inconsistent instances in the dataset have been manually annotated. %Examples are shown in Fig.\ref{fig:datasetex}.

%\vspace{-0.31cm}
\myheading{Manual Annotations.} We also go through a manual meticulous data annotation process by a team of {six} professional annotators. The annotation process is carried out following a defined procedure. First, the annotators select object-term pairs that align with each other, and then, they input the target text prompt that corresponds to the selected object-term pairs. The final step of the process involves data cleaning to ensure the accuracy and coherence of the dataset. To maintain the highest quality, the annotations are cross-validated among the team members. This step allows for the detection and rectification of any potential errors or inconsistencies. Further details about our data annotation process are in the Supplementary Material.

\def\subFigSz{0.245\linewidth}
\def\subFigHb{0.18\linewidth}
\begin{figure}[t]
\centering

\begin{minipage}[c]{\subFigSz}
\vspace{-0.1cm}
    \caption*{\tiny{COSMOS}}
\end{minipage}
\begin{minipage}[c]{\subFigSz}
\vspace{-0.1cm}
    \caption*{\tiny{NewsCLIPpings}}
\end{minipage}
\begin{minipage}[c]{\subFigSz}
\vspace{-0.1cm}
    \caption*{\tiny{FacebookPost}}
\end{minipage}
\begin{minipage}[c]{\subFigSz}
\vspace{-0.1cm}
    \caption*{ \tiny{MAIM} }
\end{minipage}
\vspace{-0.2in}

\begin{minipage}[c]{1\linewidth}
\includegraphics[width=\subFigSz, height=\subFigHb]{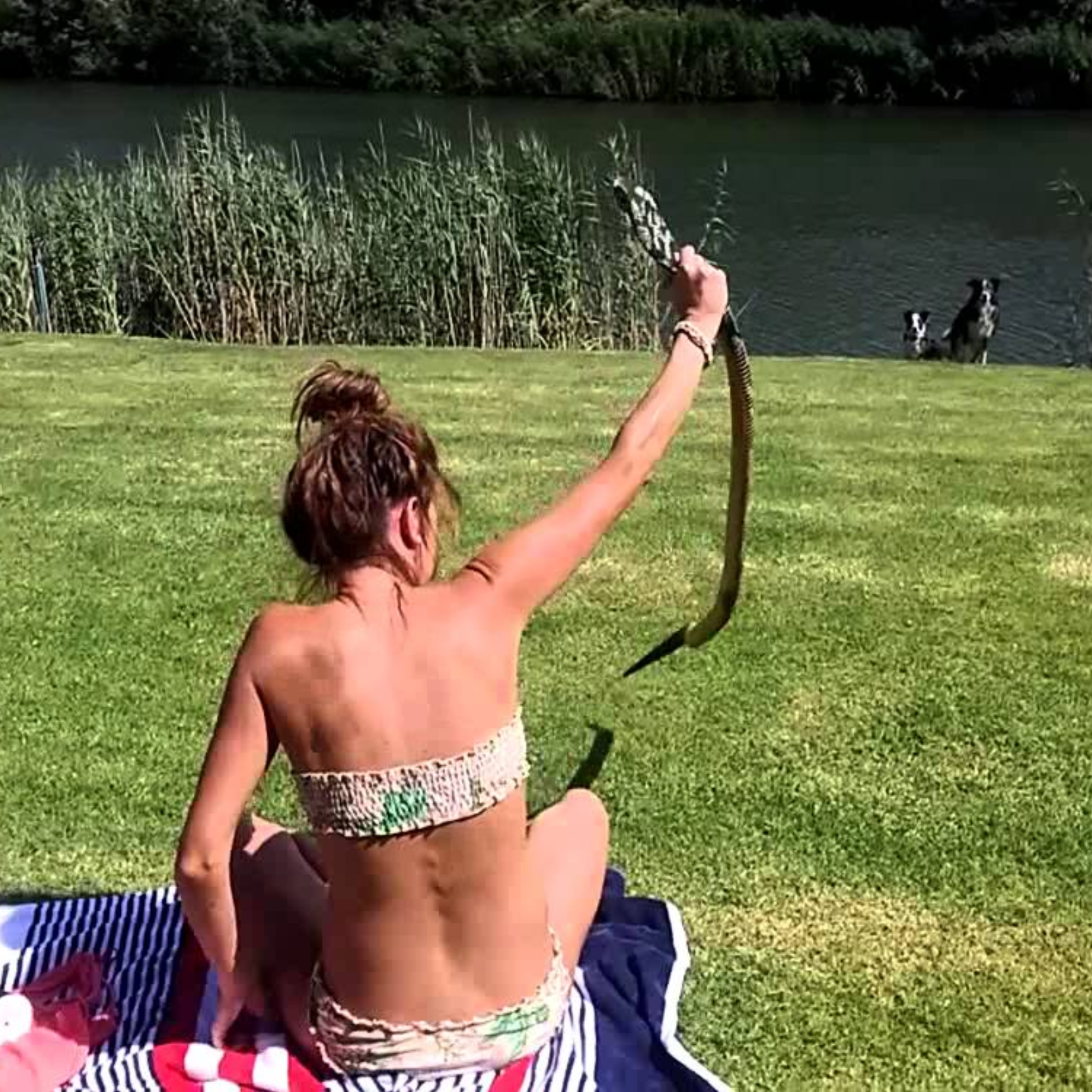}
\includegraphics[width=\subFigSz, height=\subFigHb]{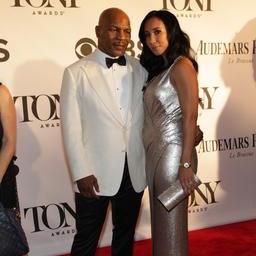}
\includegraphics[width=\subFigSz, height=\subFigHb]{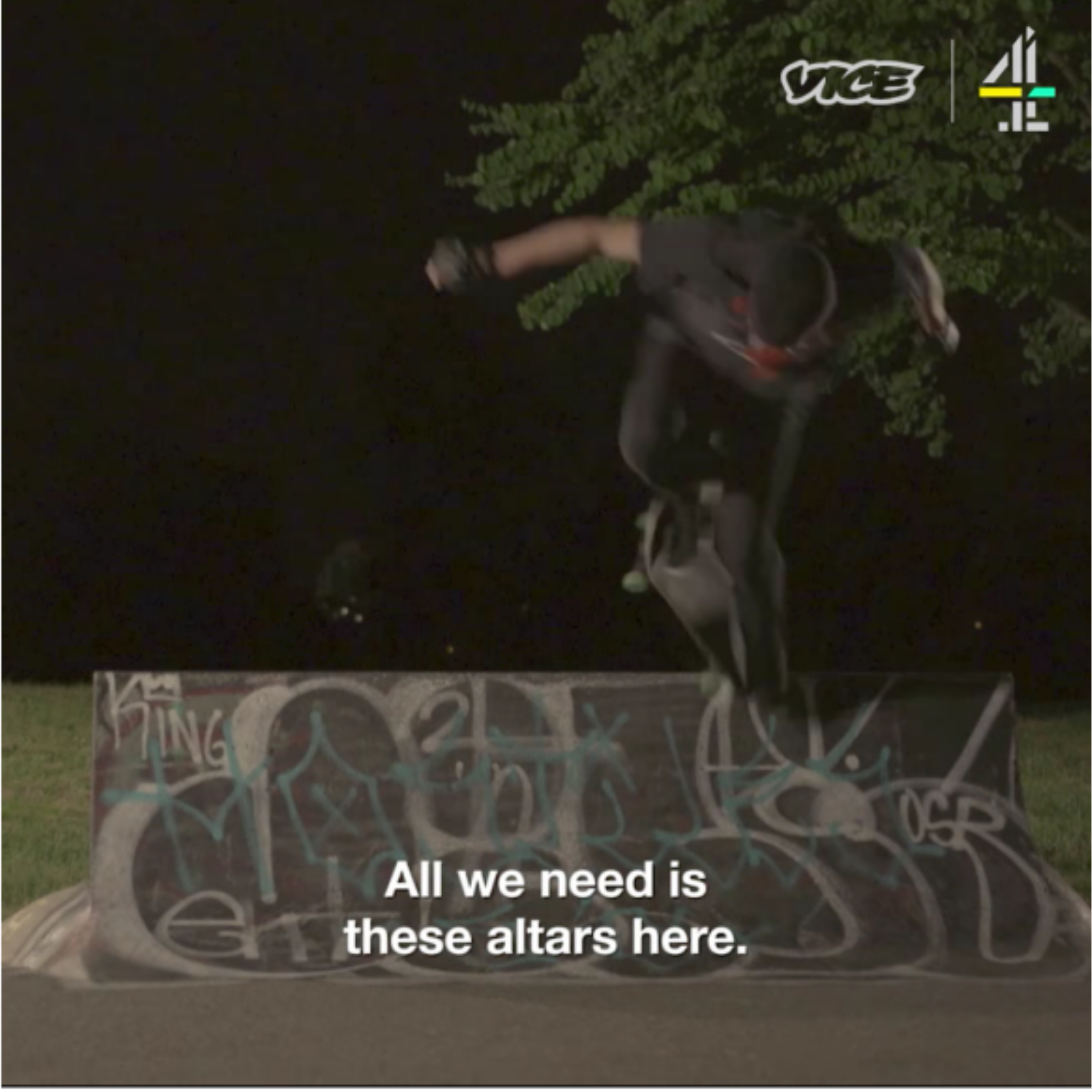}
\includegraphics[width=\subFigSz, height=\subFigHb]{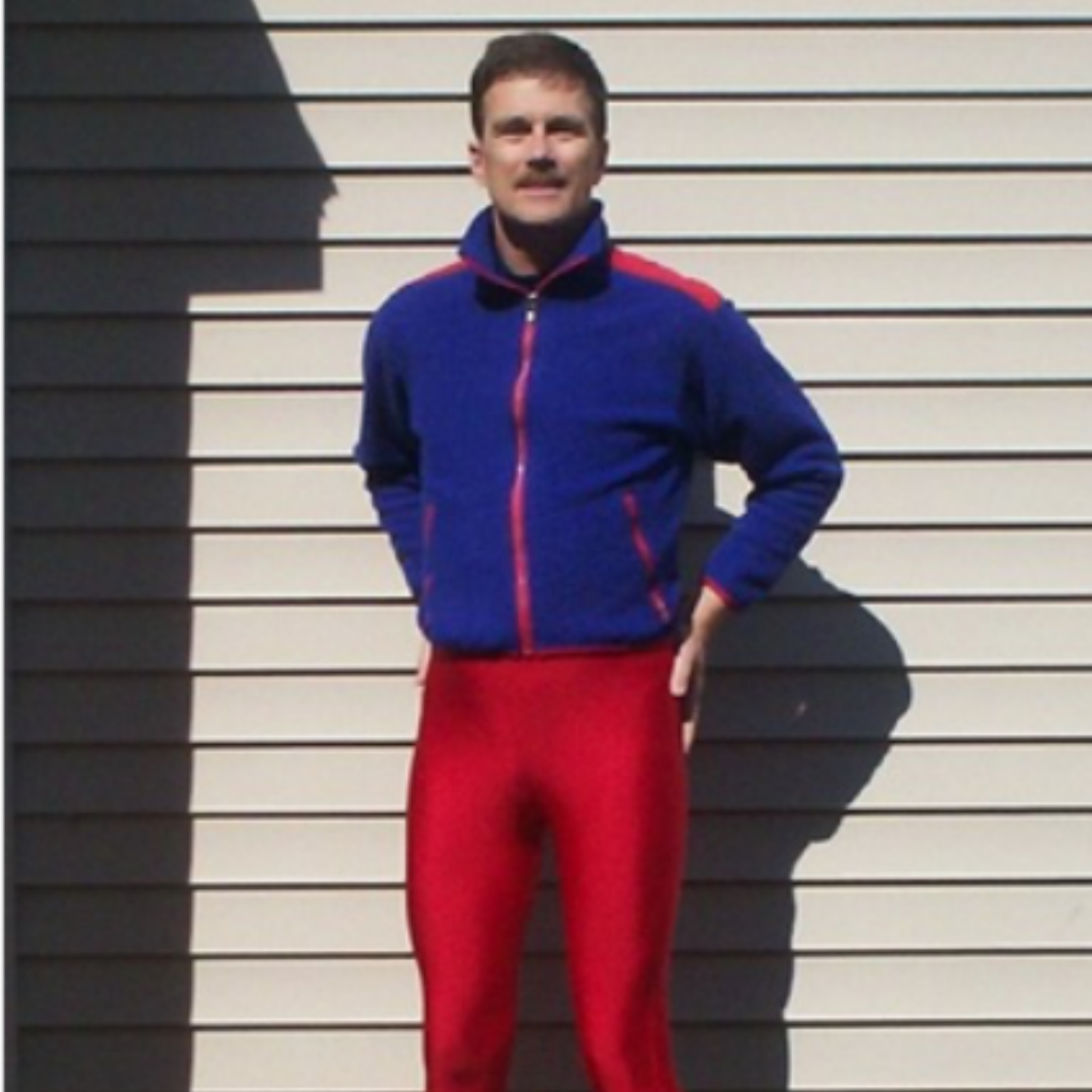}
 \\
\end{minipage}

\vspace{-0.15in}
\begin{minipage}[c]{\subFigSz}
\vspace{-0.1cm}
    \caption*{\tiny{``A sunbather catching a snake moments before it bites her"}}
\end{minipage}
\begin{minipage}[c]{\subFigSz}
\vspace{-0.1cm}
    \caption*{\tiny{``Mike Tyson and his wife Lakiha"}}
\end{minipage}
\begin{minipage}[c]{\subFigSz}
\vspace{-0.1cm}
    \caption*{\tiny{``House Minority Leader Nancy Pelosi on Kobe Bryant's last game for the LA Lakers"}}
\end{minipage}
\begin{minipage}[c]{\subFigSz}
\vspace{-0.1cm}
    \caption*{\tiny{``Budapest in August 2014"}}
\end{minipage}
\vspace{-0.3in}
\caption{\em \small Inconsistent examples from other datasets. Due to the constraints of using random swapping or auto-retrieval methods to produce inconsistent pairs, the resulting pairs could either be semantically consistent (as seen in the left two examples) or entirely unrelated (as illustrated by the right two examples).}
\label{fig:other_dataset}
\end{figure}

\def\subFigSz{0.24\linewidth}
\def\subFigH{0.15\linewidth}
\def\subFigHb{0.18\linewidth}

\begin{figure}[!t]
\centering
\vspace{-0.35cm}
\begin{minipage}[c]{1\linewidth}
\includegraphics[width=\subFigSz, height=0.13\linewidth]{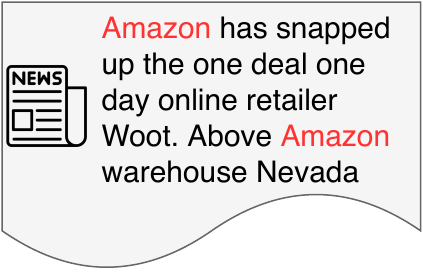}
\includegraphics[width=\subFigSz, height=\subFigH]{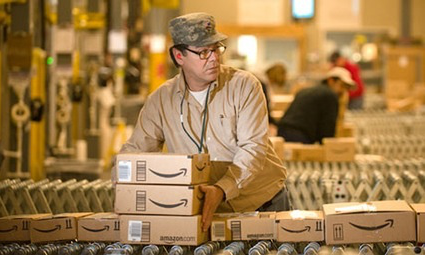}
\includegraphics[width=\subFigSz, height=\subFigH]{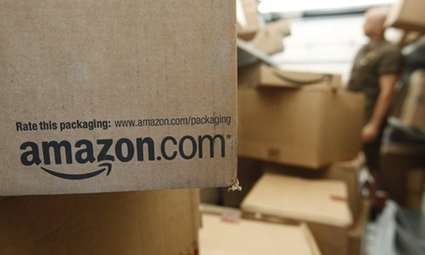}
\includegraphics[width=\subFigSz, height=\subFigH]{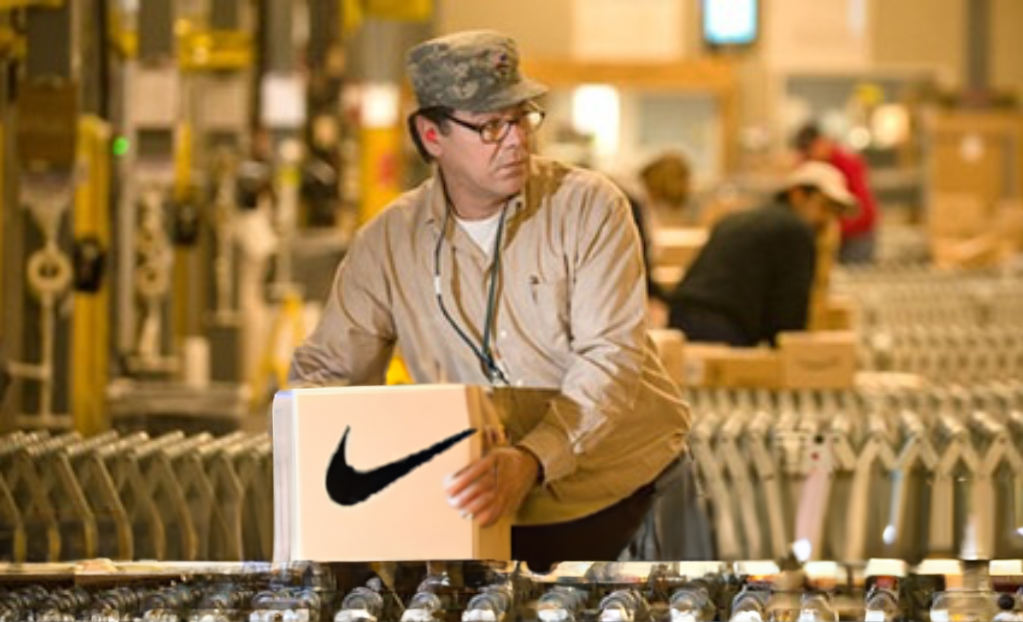}
 \\
\end{minipage}

\vspace{-0.1in}
\begin{minipage}[c]{\subFigSz}
\vspace{-0.1cm}
    \caption*{{\scriptsize{Text}}}
\end{minipage}
\begin{minipage}[c]{\subFigSz}
\vspace{-0.1cm}
    \caption*{{\scriptsize{Original image}}}
\end{minipage}
\begin{minipage}[c]{\subFigSz}
\vspace{-0.1cm}
    \caption*{{\scriptsize{NewsCLIPpings}}}
\end{minipage}
\begin{minipage}[c]{\subFigSz}
\vspace{-0.1cm}
    \caption*{{\scriptsize{TIIL: inconsistent image generated with edited term \em``Nike''}}}
\end{minipage}
% Amazon has snapped up the one deal one day online retailer Woot Above Amazon warehouse Nevada
\vspace{-0.62cm}
\caption{\em \small An inconsistent example in our TIIL and NewsCLIPpings. Note that the TIIL example agrees with a common viewer to be {\em inconsistent} while the one from NewsCLIPpings is consistent rather than inconsistent.
\label{fig:newsclipping}}
\vspace{-0.2in}
\end{figure}

\myheading{Comparison with Existing Datasets.}  As shown in \Tref{tab:dataset}, %TIIL has three advantages when compared with existing image inconsistency datasets. 
our TIIL dataset boasts several unique characteristics that set it apart from existing datasets. 
To the best of our knowledge, it is the first of its kind to feature both pixel-level and word-level inconsistencies, offering fine-grained and reliable inconsistency. 
We provide a comparison of the CLIP scores between the traditional random swap-based creation method and our diffusion-based approach in \Tref{tab:main}. The results demonstrate the superiority of our method in achieving higher levels of semantic similarity compared to randomly swapped image-text inconsistency pairs. This higher semantic similarity makes the inconsistencies more subtle and harder to detect, thereby enhancing the dataset's complexity and realism. In comparison to datasets that build inconsistency pairs through random swap (e.g., MAIM~\citep{jaiswal2017multimedia}, MEIR~\citep{sabir2018deep}, FacebookPost~\citep{mccrae2022multi} and COSMOS~\citep{aneja2021cosmos}) or automatic retrieval (such as NewsCLIPping~\citep{luo2021newsclippings}), TIIL offers more reliable consistent and inconsistent pairs, as demonstrated in \Fref{fig:other_dataset} and \Fref{fig:newsclipping}. Moreover, %our dataset captures realistic and plausible inconsistency scenarios, thereby enhancing the authenticity of the generated image-text pairs. 
our dataset is sourced from a wide array of news topics and sources, ensuring a diverse and rich collection of examples. For a more comprehensive understanding, we present examples of image-text pairs and annotated labels from the TIIL dataset in \Fref{fig:datasetex1}. These examples underline the range and complexity of the data within our novel dataset.

\setlength{\tabcolsep}{2pt}    
% \begin{table}[!t]
%  \small
% \begin{minipage}[]{0.34\linewidth} \centering
% \scalebox{0.9}{ 
% \begin{tabular}{lcc}
% \rowcolor{mygray}
%     &  Real-world & DALL-E2 \\
% %\rowcolor{mygray}  
% % &  Pairs &  Pairs \\
% \hline \thickhline\
%    % &    &  Pairs  \\   \hline \thickhline
%  Avg CLIP Score & 30.71 & 31.03\\
% \hline
% \end{tabular}}
% % \vskip -0.1in
% % \caption{{\bf } }
% \captionsetup{justification=centering}
% \vspace{0.05cm}
% \caption{{Comparison of consistent image-text pairs.} }
% \label{tab:dalle_consis}
% %
% \scalebox{0.9}{ \begin{tabular}{lcc}
% \rowcolor{mygray}
%    &  Random Swap & Ours \\   \hline \thickhline
%  Avg CLIP Score & 12.39  & 22.68\\
% \hline
% \end{tabular}}
% % \vskip -0.1in
% \captionsetup{justification=centering}
% \vspace{0.05cm}
% \caption{{Comparison of mis-\\contextualization generation.} }
% \label{tab:main}
% \end{minipage}

\vspace{-0.15in}
\setlength{\tabcolsep}{3pt}    
\begin{table}[!t]
 \small
\begin{minipage}[]{0.32\linewidth} \centering
\scalebox{0.9}{ 
\begin{tabular}{lc}
\rowcolor{mygray}
    &  Avg. CLIP Score \\
%\rowcolor{mygray}  
% &  Pairs &  Pairs \\
\hline \thickhline
Real-world  & 30.71 \\
DALL-E2 & 31.03 \\ 
\hline
Random Swap & 12.39 \\
Ours-Inconsistent & 22.68 \\
\hline
\end{tabular}}
\vspace{0.1cm}
\caption{\em \small{Comparison of the CLIP scores for different pairs.} }
\label{tab:main}
\label{tab:dalle_consis}
\end{minipage}
\vspace{-0.43cm}
%
%\hspace{0.05in}
\renewcommand\arraystretch{1.0}
\begin{minipage}[]{0.55\linewidth}\centering
\scalebox{0.82}{ 
\setlength{\tabcolsep}{3pt} 
\begin{tabular}{lccccc}
\rowcolor{mygray}
    & Total &Inconsistent & Anno & Generation & Pixel\&word-level    \\ 
\rowcolor{mygray}   
   Dataset & \#Pairs & \#Pairs  & \#Pairs & Method & Annotation
   \\   \hline \thickhline
    MAIM & 240K & 120K & 0 & Random swap   & $\times$\\ % ~\citep{jaiswal2017multimedia}
   MEIR & 58K &29K &  0 & Random swap   & $\times$\\ % ~\citep{sabir2018deep}
   NewsCLIPpings  & 988K &494K &  0 & Auto retrieval   & $\times$\\ % ~\citep{luo2021newsclippings}
   FacebookPost & 4K &2K &  0 & Random swap   & $\times$ \\ % ~\citep{mccrae2022multi}
   COSMOS   & 200K &850 &  1.7K & Manual    & $\times$ \\ \hline % ~\citep{aneja2021cosmos}
   TIIL (Ours) & 14K & 7K &  14K& Manual+Diffusion  & \checkmark \\
\hline
\end{tabular}}

\vspace{-0.2cm}
\captionsetup{justification=centering}
\caption{\em \small{Comparison with existing related datasets.} }
\label{tab:dataset}
\end{minipage}
\vspace{-0.2cm}
\end{table}

\section{Experiments} %
\vspace{-0.2cm}
This section presents a comprehensive analysis of our approach, including qualitative and quantitative results, comparisons with other methods, and ablation studies to evaluate different variations.
 
\def\subFigSz{0.24\linewidth} %width=0.24\linewidth
\def\subFigH{0.15\linewidth}
\def\subCapH{0.07\linewidth}
\begin{figure}[t]
\centering
\includegraphics[width=\subFigSz, height=\subFigH]{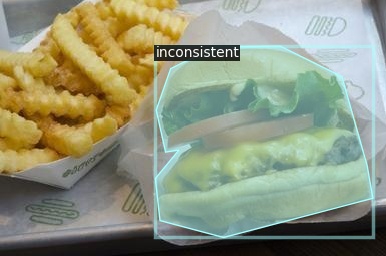}
\includegraphics[width=\subFigSz, height=\subFigH]{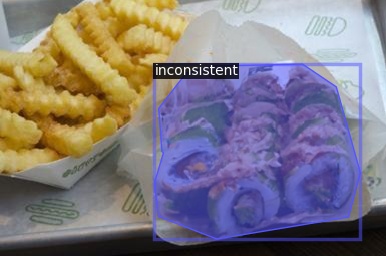}
\includegraphics[width=\subFigSz, height=\subFigH]{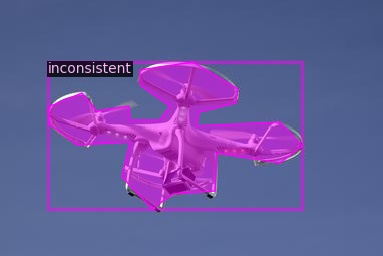}
\includegraphics[width=\subFigSz, height=\subFigH]{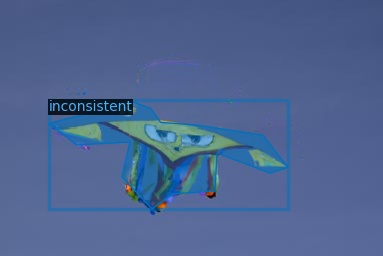}\\

\begin{minipage}[l]{0.48\linewidth}
    % \caption*{\scriptsize \it {``View photo gallery Washington Post food critic Tom Sietsema picked the Dupont Circle location of\colorbox{gray!10}{Sushi}/\colorbox{purple!30}{Shake Shack}among the city's best cheap eats''}}
    \caption*{\scriptsize \it {``View photo gallery Washington Post food critic Tom Sietsema picked the Dupont Circle location of \textcolor{gray!10}{Sushi}/\textcolor{purple}{Shake Shack}among the city's best cheap eats''}}
\vspace{-0.31cm}
\end{minipage}
\begin{minipage}[c]{0.48\linewidth}
    \caption*{\centering\scriptsize \it ``A \textcolor{magenta}{kite}/\textcolor{blue}{drone} flies at the International Consumer Electronics Show in January 2014 in Las Vegas''}
\vspace{-0.31cm}
\end{minipage}

\includegraphics[width=\subFigSz, height=\subFigH]{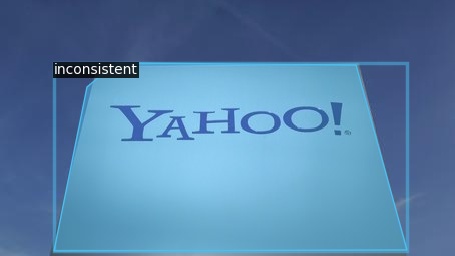}
\includegraphics[width=\subFigSz, height=\subFigH]{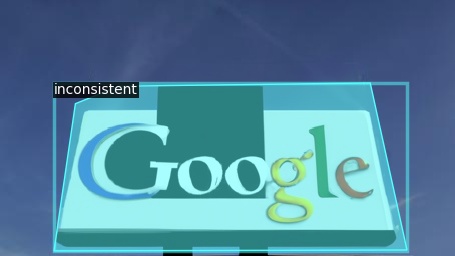}
\includegraphics[width=\subFigSz, height=\subFigH]{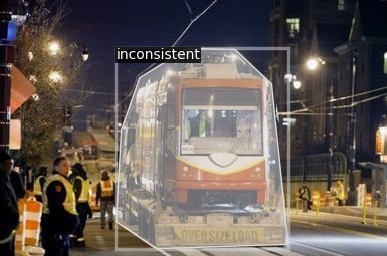}
\includegraphics[width=\subFigSz, height=\subFigH]{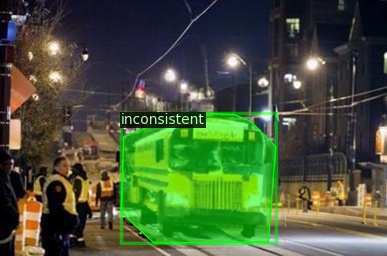}\\
\begin{minipage}[c]{0.48\linewidth}
    \caption*{{\scriptsize \it ``\textcolor{cyan}{Google}/\textcolor{green}{Yahoo} plans to spin off core web business''}}
\end{minipage}
\begin{minipage}[c]{0.48\linewidth}
    \caption*{\centering \it \scriptsize {``A \textcolor{gray!10}{school bus}/\textcolor{green!40}{new streetcar} is placed on the rails along H St NE''}}
\end{minipage}
\vspace{-0.25in}
\caption{\em \small Examples in TIIL dataset. Colored texts correspond to inconsistent regions of the same color in the image. The figure is best viewed in color.
\label{fig:datasetex1}}
\vspace{-0.16in}
\end{figure}

\def\subFigSz{0.19\linewidth}
\def\subFigH{0.16\linewidth}
% \captionsetup{type=figure} 
% \includegraphics[]{
\begin{figure}[!t]
\centering
\begin{minipage}[c]{0.02\linewidth}
    \caption*{\rotatebox{0}{\tiny{${I}$}}}
\end{minipage}\hfill
\begin{minipage}[c]{0.95\linewidth}
    \includegraphics[width=\subFigSz, height=\subFigH]{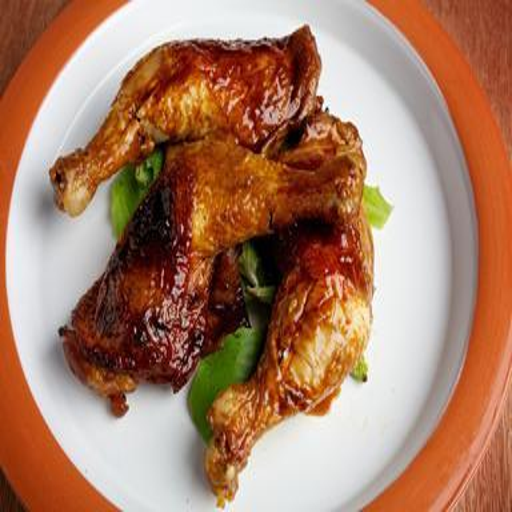}
    \includegraphics[width=\subFigSz, height=\subFigH]{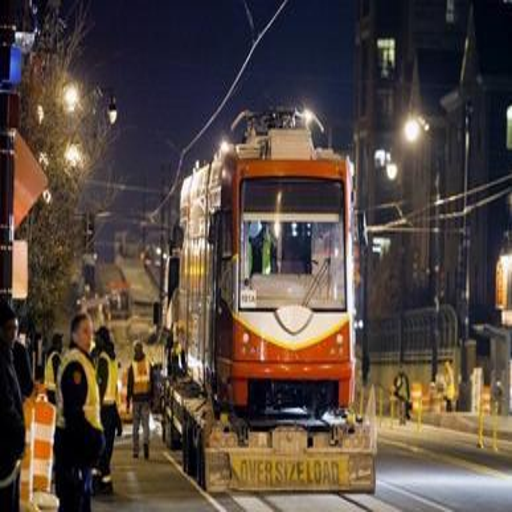}
    \includegraphics[width=\subFigSz, height=\subFigH]{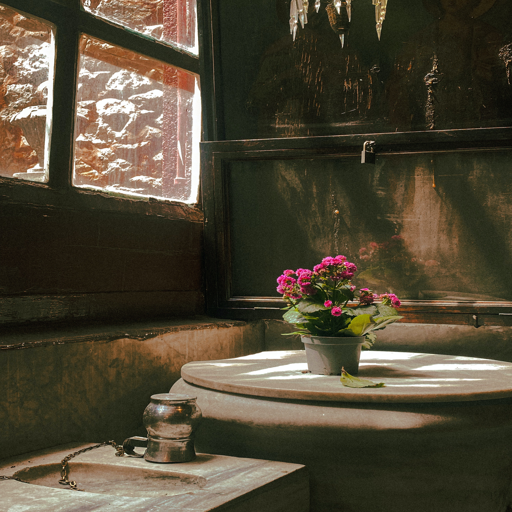}
    \includegraphics[width=\subFigSz, height=\subFigH]{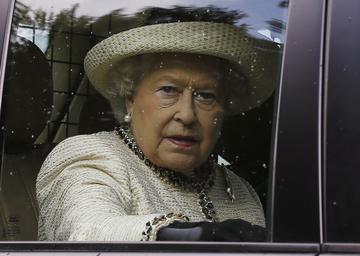}
    \includegraphics[width=\subFigSz, height=\subFigH]{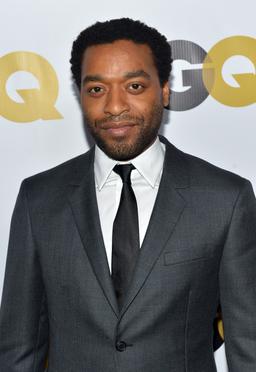}
\end{minipage}

\begin{minipage}[c]{0.02\linewidth}
    \caption*{\rotatebox{0}{\tiny{$\mM^{\prime}$}}}
\end{minipage}\hfill
\begin{minipage}[c]{0.95\linewidth}
    \includegraphics[width=\subFigSz, height=\subFigH]{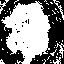}
    \includegraphics[width=\subFigSz, height=\subFigH]{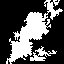}
    \includegraphics[width=\subFigSz, height=\subFigH]{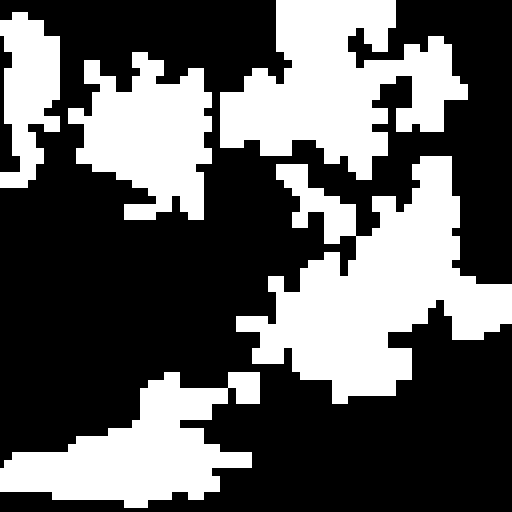}
    \includegraphics[width=\subFigSz, height=\subFigH]{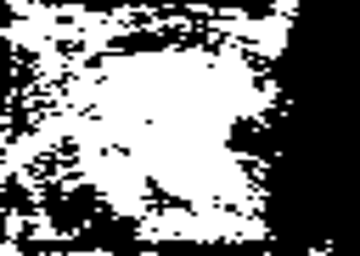}
    \includegraphics[width=\subFigSz, height=\subFigH]{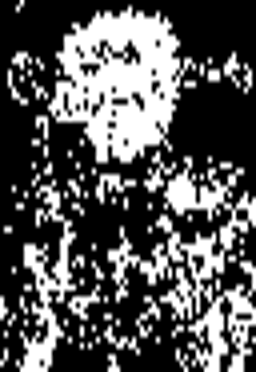}
\end{minipage}

\begin{minipage}[c]{0.02\linewidth}
    \caption*{\rotatebox{0}{\tiny{${I}_{edt}$}}}
\end{minipage}\hfill
\begin{minipage}[c]{0.95\linewidth}
    \includegraphics[width=\subFigSz, height=\subFigH]{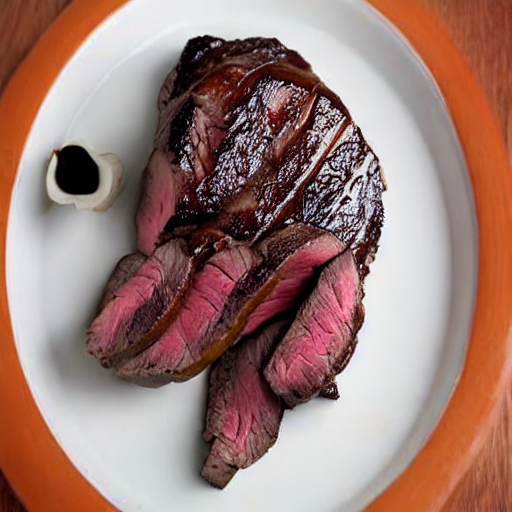}
    \includegraphics[width=\subFigSz, height=\subFigH]{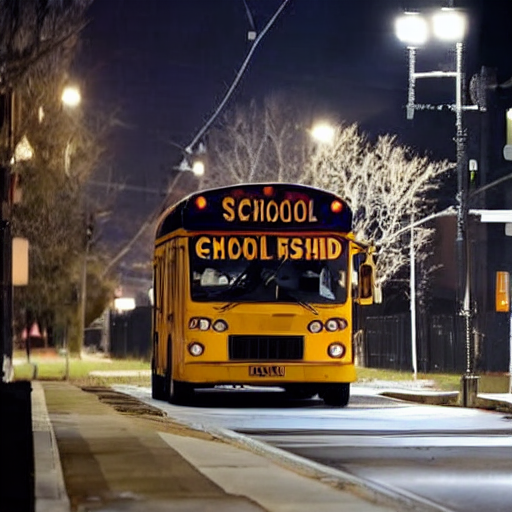}
    \includegraphics[width=\subFigSz, height=\subFigH]{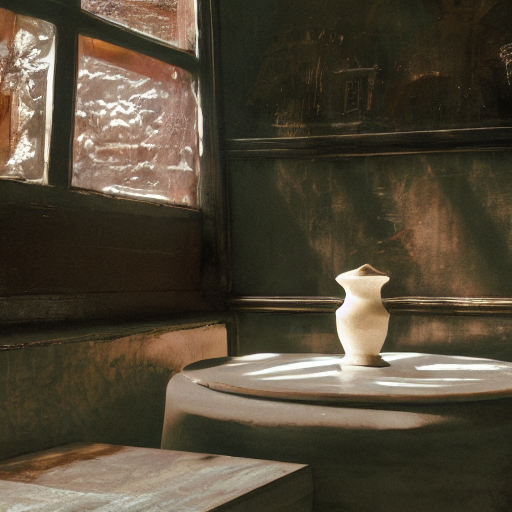}
    \includegraphics[width=\subFigSz, height=\subFigH]{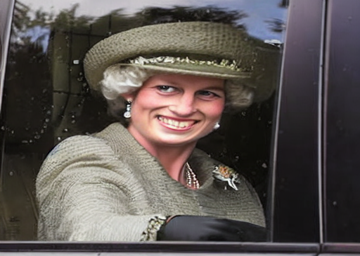}
    \includegraphics[width=\subFigSz, height=\subFigH]{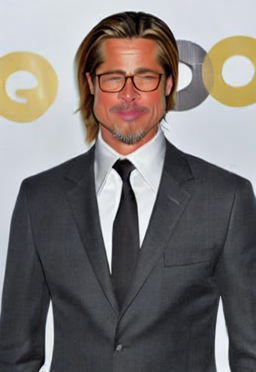}
\end{minipage}

\begin{minipage}[c]{0.02\linewidth}
    \caption*{\rotatebox{0}{\tiny{$\mM$}}}
\end{minipage}\hfill
\begin{minipage}[c]{0.95\linewidth}
    \includegraphics[width=\subFigSz, height=\subFigH]{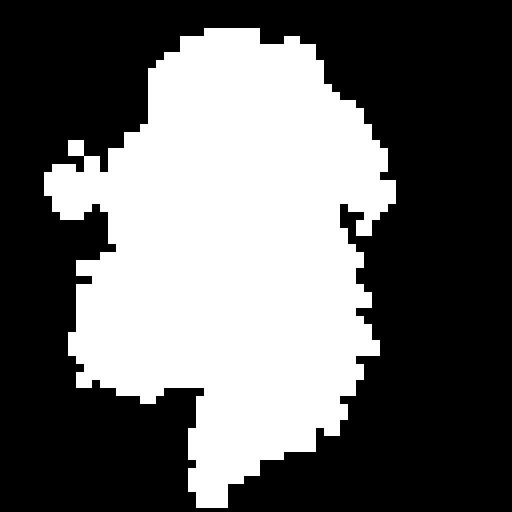}
    \includegraphics[width=\subFigSz, height=\subFigH]{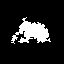}
    \includegraphics[width=\subFigSz, height=\subFigH]{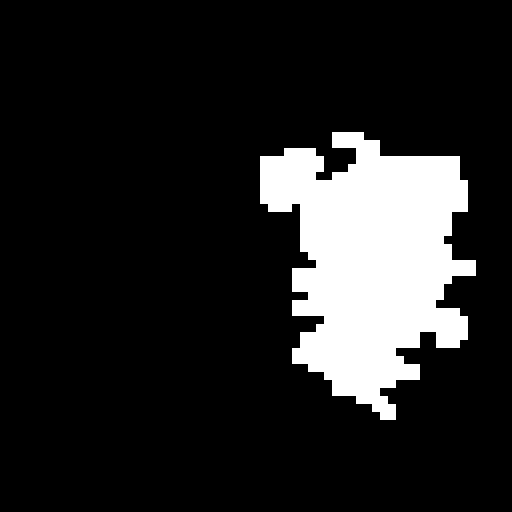}
    \includegraphics[width=\subFigSz, height=\subFigH]{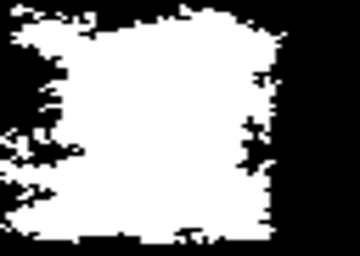}
    \includegraphics[width=\subFigSz, height=\subFigH]{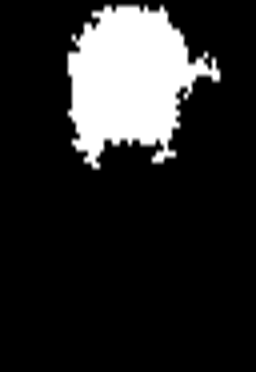}
\end{minipage}

\begin{minipage}[c]{0.02\linewidth}
    \caption*{\rotatebox{0}{\tiny{$r$}}}
\end{minipage}\hfill
% \hspace{0.01in}
\begin{minipage}[c]{0.95\linewidth}
\captionsetup{justification=centering}
\begin{minipage}[c]{\subFigSz}
    \caption*{\scriptsize \it Score: 13.2}
\end{minipage}
% \hspace{-0.01in}
\begin{minipage}[c]{\subFigSz}
    \caption*{\scriptsize \it Score: 26.3}
\end{minipage}
\begin{minipage}[c]{\subFigSz}
    \caption*{\scriptsize \it Score: 31.6}
\end{minipage}
\begin{minipage}[c]{\subFigSz}
    \caption*{\scriptsize \it Score: 11.8}
\end{minipage}
\begin{minipage}[c]{\subFigSz}
    \caption*{\scriptsize \it Score: 25.9}
\end{minipage}
\end{minipage}
\vspace{-0.3cm}

\begin{minipage}[c]{0.02\linewidth}
    \caption*{\rotatebox{0}{\tiny{${T}$}}}
\end{minipage}\hfill
% \hspace{0.01in}
\begin{minipage}[c]{0.95\linewidth}
\captionsetup{justification=centering}
\begin{minipage}[c]{\subFigSz}
\vspace{-0.1cm}
    \caption*{\scriptsize \it ``\textcolor{red}{Roast steak} in Pomegr-\\anate and Date Molasses''}
\end{minipage}
% \hspace{-0.01in}
\begin{minipage}[c]{\subFigSz}
\vspace{-0.1cm}
    \caption*{\scriptsize \it ``A \textcolor{red}{school bus} is placed on the rails along H St NE''}
\end{minipage}
\begin{minipage}[c]{\subFigSz}
\vspace{-0.1cm}
    \caption*{\scriptsize \it ``The \textcolor{red}{vase} sitting on the table is an object dating back to the previous century''}
\end{minipage}
\begin{minipage}[c]{\subFigSz}
\vspace{-0.1cm}
    \caption*{\scriptsize \it ``Britain's \textcolor{red}{Queen Diana} leaves the annual Braemar Highland Gathering in Braemar Scotland Sept 6 2014''}
\end{minipage}
\begin{minipage}[c]{\subFigSz}
\vspace{-0.1cm}
    \caption*{\scriptsize \it ``\textcolor{red}{Brad Pitt} 12 Years a Slave''}
\end{minipage}
\end{minipage}

\vspace{-0.63cm}
\caption{\em \small D-TIIL on TIIL examples. The detected inconsistent words are highlighted in \textcolor{red}{red}.} 
\label{fig:results}
\vspace{-0.25in}

\end{figure}

\begin{table}[t]
 \small
\renewcommand\arraystretch{1.0}
\begin{minipage}[b]{0.45\linewidth} \centering
\scalebox{0.9}{ 
\setlength{\tabcolsep}{8pt} 
\begin{tabular}{lccc}

\rowcolor{mygray}
   &  DetCLIP & GAE & Ours   \\   \hline \thickhline %GAE
  mIoU (\%) & 14.24 & 27.35  & 47.12\\
  % Pixel Acc && 22.40\\
\hline
\end{tabular}}
% \vskip -0.1in
\captionsetup{justification=centering}
\vspace{-0.2cm}
\caption{\small \it Comparison of text-image inconsistency localization.}
\vskip -0.13in
\label{tab:compare}
\end{minipage}
\begin{minipage}[b]{0.5\linewidth} \centering
\scalebox{0.9}{ 
\setlength{\tabcolsep}{6pt} 
\begin{tabular}{lcccc}
\rowcolor{mygray}
   &  CLIP & CLIP* & CCN & Ours \\   \hline \thickhline
  AUC (\%) & 84.18 & 78.58 & 82.80  & \textbf{87.00}\\ % 79.11 76.77 	82.80
  Accuracy (\%) & 76.77 &	71.87  & 74.10 & \textbf{79.46} \\
\hline
\end{tabular}}
% \vskip -0.1in
\captionsetup{justification=centering}
\vspace{-0.2cm}
\caption{\small \it Comparison of detection.}
\vskip -0.1in
\label{tab:det}
\end{minipage}
\vspace{-0.5cm}
\end{table}
 %--------------------------------------------
% \vskip -0.1in
% \setlength{\tabcolsep}{6pt} 
% \begin{table}[!ht]
%  \small
% \begin{minipage}[b]{0.38\linewidth}\centering
% \scalebox{0.9}{ 
% \setlength{\tabcolsep}{10pt} 
% \begin{tabular}{lcc}
% \rowcolor{mygray}
%    &  $\mM^{\prime}$ & $\mM$ \\   \hline \thickhline
%   mIoU (\%) & 38.43  & 47.12\\
% \hline
% \end{tabular}}
% % \vskip -0.1in
% \captionsetup{justification=centering}
% \vspace{0.1cm}
% \caption{Benefits of text denoising.}
% % \vskip -0.1in
% \label{tab:denoise}
% \end{minipage}
% \begin{minipage}[b]{0.6\linewidth} \centering
% \scalebox{1}{ 
% \setlength{\tabcolsep}{4pt} 
% \begin{tabular}{lccc}

% \rowcolor{mygray}
%    &  Random Init & w/o Emb Constrain & Ours   \\   \hline \thickhline %GAE
%   mIoU (\%)&  41.27&  43.47 & 47.50 \\
%   % Pixel Acc && 22.40\\
% \hline
% \end{tabular}} 
% % \vskip -0.1in
% \captionsetup{justification=centering}
% \vspace{0.1cm}
% \caption{Comparison of text embedding alignments.}
% \label{tab:emb}
% \end{minipage}
% \vspace{-2em}
% \end{table}

\vspace{-0.1cm}
\subsection{Settings}
%\vspace{-0.1cm}

\myheading{Implementation Details}.
We use the implementations of Stable Diffusion~\citep{rombach2022high} and CLIP~\citep{CLIP_radford2021} ViT-B/32 model available on \url{https://huggingface.co}. In the diffusion model, we use the denoising diffusion implicit model (DDIM)~\citep{song2020denoising} to sample noises, and classifier-free guidance~\citep{ho2022classifier} is set to the
recommended value of 7.5. 
%The input image is $512 \times 512$ pixels in size. All experiments were conducted on NVIDIA A6000 GPUs. 
We train both text embedding ${E}_{aln}$ and ${E}_{dnt}$ for $500$ iterations with a learning rate of $4e^{-6}$. The hyperparameter $\gamma$ is set to 8 in our experiment.  For noise estimation, we use the same random seed for two conditioned text embeddings, remove outlier values in noise predictions, and average spatial differences over a set of $10$ input noises. After obtaining the predicted inconsistent masks, we use a threshold to binarize them where the threshold is the average values among the mask. We only retain the top $3$ mask regions with the largest areas. To localize inconsistent words, we follow the previous work \citep{CLIP_radford2021} to use a prompt template ``A photo of \{words\}'' for CLIP~\citep{CLIP_radford2021} text embedding generation.

\vspace{0.2cm}
\myheading{Evaluation Metrics.} We report the mean of class-wise intersection over union (mIoU) \citep{everingham2015pascal} to evaluate the quality of the predicted inconsistency mask. mIoU is a metric that aligns with the per-pixel classification formulation, making it a commonly used standard metric in semantic segmentation tasks \citep{fan2021rethinking, klingner2021improving, gao2022large, xu2022multi}. 

\subsection{Comparison with Existing Methods} 
\vspace{-0.1cm}
%\myheading{Localization.} 
We first present qualitative results demonstrating the pixel-level and word-level detection of inconsistency achieved by the proposed D-TIIL model in \Fref{fig:results}. It can be observed that our D-TIIL achieves accurate results benefiting from the multi-step semantic alignment. Given the absence of prior work especially in addressing image inconsistency localization, we consider the following two relevant baseline approaches for comparison: (1) a straightforward solution that uses an object detector~\citep{zhou2022detecting} to detect all objects as segmentation mask in the image and then compares the CLIP~\citep{CLIP_radford2021} embedding similarity between the text and each object region. 
%Subsequently, the CLIP~\citep{CLIP_radford2021} text encoder is employed to encode the news text, while the CLIP image encoder is used to encode all object regions, resulting in image embeddings. Next, we calculate the cosine similarity between the text embedding and each object embedding. 
The object region with the highest dissimilarity is identified, and its corresponding mask is considered as the inconsistent mask. We denote this method as DetCLIP;
(2) an off-the-shelf method, GAE~\citep{chefer2021generic}, which provides explainability for bi-modal and encoder-decoder transformers by presenting co-attention maps. Specifically, it analyzes the classification relevancy of specific layers for CLIP~\citep{CLIP_radford2021} to provide a pixel-level attention heatmap. We further generate the inconsistent mask with the attention heatmap by applying a threshold.
The comparison results on the TIIL dataset are shown in \Tref{tab:compare} and \Fref{fig:gae}. The D-TIIL method demonstrates significant superiority over the baseline methods in both mIoU scores and qualitative evaluation.

\vspace{-0.12cm}
\subsection{Ablation Studies}
\vspace{-0.1cm}
%\vspace{-0.2cm}

\myheading{Text-image Inconsistency Detection.} %Combining with the CLIP~\citep{CLIP_radford2021}, 
While we have emphasized that binary classification may not be the best method for revealing inconsistencies in text image pairs, we have adapted the D-TIIL model into a binary framework. This allows us to compare D-TIIL with current text-image inconsistency detectors, such as the CLIP model~\citep{CLIP_radford2021}, its fine-tuned version on NewsClippings (referred to as CLIP*), and a recent detector that shows the best performance CCN~\citep{abdelnabi2022open}. We report the Area Under ROC (AUC) and Accuracy. As shown in \Tref{tab:det}, the D-TIIL model outperforms other models in terms of both AUC and Accuracy scores. This improvement can be attributed to the use of the detected inconsistency mask for image embedding extraction, which enables the exclusion of distracting and implicit regions from the image, thereby enhancing the classification performance. Moreover, compared with CCN, which requires inverse searches for inconsistency detection, our D-TIIL does not require information from external sources, especially showing superior performance on pairs with manipulated images that cannot be found on the Internet. Specifically, the experimental results demonstrate that CCN achieves an accuracy of 80.12\% on the subset composed entirely of original images/text sourced from the Internet, However, its accuracy drops to 68.0\% for subsets containing manipulated image/text. In comparison, our method obtains an accuracy of 82.79\% and 79.15\% on two subsets, respectively, indicating minimal impact on the online accessibility of data.

\vspace{0.2cm}
\myheading{Image-regulated Text Denoising.} We first highlight the benefits of performing additional image-regulated text denoising in Step 3 instead of using the $\mM^{\prime}$ in Step 2 as the inconsistency map. \Tref{tab:denoise} and \Fref{fig:results} reveal that $\mM^{\prime}$ exhibits a coarse pixel-level inconsistency map, whereas the detected inconsistency mask includes additional background areas due to the absence of denoising in the input text embeddings.

\begin{table}[!b]
 \small
 \vskip -0.2in
\begin{minipage}[b]{0.38\linewidth}\centering
\scalebox{1}{ 
\setlength{\tabcolsep}{14pt} 
\begin{tabular}{lcc}
\rowcolor{mygray}
   &  $\mM^{\prime}$ & $\mM$ \\   \hline \thickhline
  mIoU (\%) & 38.43  & 47.12\\
\hline
\end{tabular}}
% \vskip -0.1in
\captionsetup{justification=centering}
\vspace{-0.21cm}
\caption{\small \it Benefits of text denoising.}
% \vskip -0.1in
\label{tab:denoise}
\end{minipage}
\begin{minipage}[b]{0.6\linewidth} \centering
\scalebox{1}{ 
\setlength{\tabcolsep}{4pt} 
\begin{tabular}{lccc}

\rowcolor{mygray}
   &  Random Init & w/o Emb Constrain & Ours   \\   \hline \thickhline %GAE
  mIoU (\%)&  41.27&  43.47 & 47.50 \\
  % Pixel Acc && 22.40\\
\hline
\end{tabular}}
% \vskip -0.1in
\captionsetup{justification=centering}
\vspace{-0.21cm}
\caption{\small \it Comparison of text embedding alignments.}
\label{tab:emb}
\end{minipage}
% \vspace{-2em}
\end{table}

\myheading{Text Embedding Alignments.} We further compare our method with two variations to show the influence of the text embedding alignment process on inconsistency localization. This experiment is conducted in a randomly sampled subset of TIIL with 1000 image-text pairs. We consider two initialization variations, (1) random initialization of $E_0$ from random noise; %, such random initialization expands the embedding distance for optimization which leads to a slower converge and a downgraded performance. In the second method 
and (2) initialize $E$ with $E_0$ and is only supervised by the reconstruction $\mL_2$ loss without the text embedding constrain loss in \Eref{eq:constrain}. \Tref{tab:emb} demonstrates that the performance decreases when the aligned text differs from the initial text embedding $E_0$ too much. %Moreover, we also compare the cosine similarity between the consistent text embedding and inconsistent text embedding (80.98\%) with the cosine similarity between the consistent text embedding and random Gaussian text embedding (0.05\%), this also explains why initializing with $E_0$ is better than random ones. 

%\myheading{Limitations.} We found that leveraging LIMs to generate images provide more aligned consistent pairs. As shown in Table~\ref{tab:dalle_consis}, we compare the semantic consistency level of real-world image-text pairs and diffusion model (i.e., DALL-E2~\citep{DALLE2ramesh2022hiera}) aligned consistency pairs measured by CLIP~\citep{CLIP_radford2021} score on TIIL dataset. Note that the CLIP score corresponds to the cosine similarity between the visual CLIP embedding for an image and textual CLIP embedding for a text. The score is bound between 0 and 100, and the closer to 100, the better. Table~\ref{tab:dalle_consis} reveals that the diffusion model generation achieves a higher CLIP similarity score than real-world image-text pairs from VisualNews dataset~\citep{liu2020visual}.
\vspace{-0.1cm}
\subsection{Failure Cases}  
\vspace{-0.1cm}
\Fref{fig:fail} shows two examples that D-TIIL does not generate results consistent with human viewers. Such cases are attributed to the limitations of the underlying text-to-image diffusion models in understanding the entailment of semantic meanings of words and creating precise local edits of images. In \Fref{fig:fail}~(a), the word ``office'' is likely to be taken too literally by the model, even though a human viewer may extrapolate to this unusual setting. The case of \Fref{fig:fail}~(b) is a clear inconsistent pair, as the word ``stuffed animal'' is not reflected in the original image, but D-TIIL finds misaligned inconsistent regions.
% One reason 
% show knowledge-based cases
% \lsw{show failure cases}

\def\subFigSz{0.16\linewidth} %width=0.24\linewidth
\def\subFigH{0.12\linewidth}
\begin{figure}[!t]
\centering
\includegraphics[width=\subFigSz, height=\subFigH]{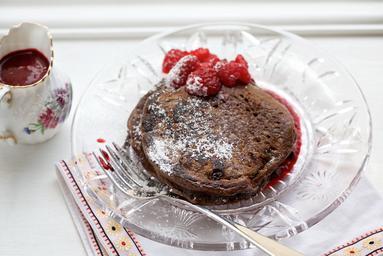}
\includegraphics[width=\subFigSz, height=\subFigH]{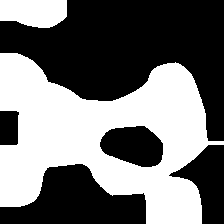}
\includegraphics[width=\subFigSz, height=\subFigH]{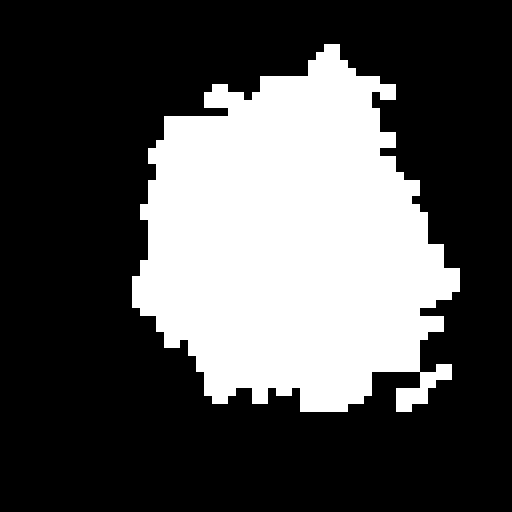}
\includegraphics[width=\subFigSz, height=\subFigH]{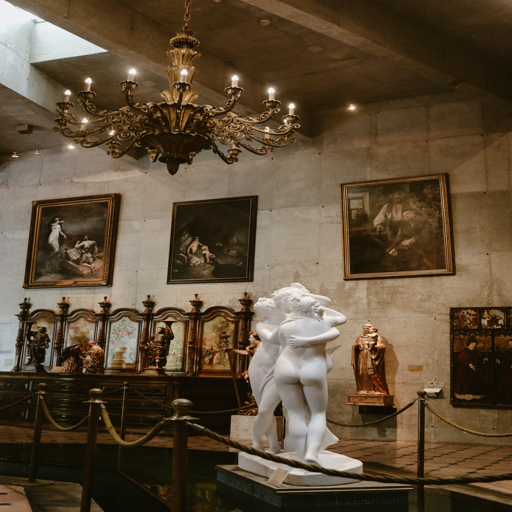}
\includegraphics[width=\subFigSz, height=\subFigH]{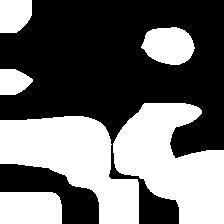}
\includegraphics[width=\subFigSz, height=\subFigH]{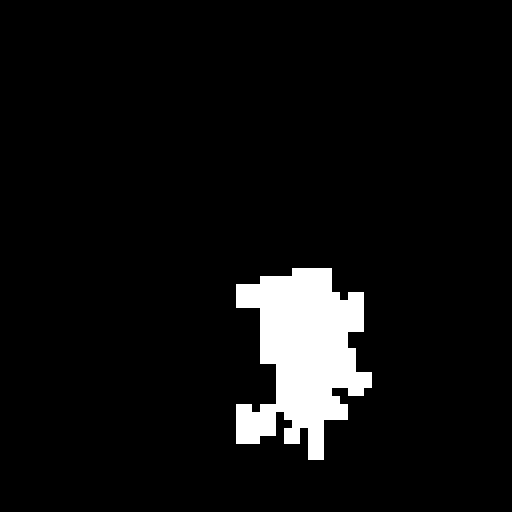}
% \includegraphics[width=\subFigSz, height=0.08\linewidth]{Figure/compare/taco.pdf}
% \begin{minipage}[c]{\subFigSz}
%     \caption*{{\footnotesize{Input text: Shrimp Tacos on a nice plate.}}}
% \end{minipage}
% \caption*{{\footnotesize{(d)}}}

\captionsetup{justification=centering}
\begin{minipage}[c]{\subFigSz}
    \caption*{{\scriptsize \it ``Shrimp Tacos\\ on a nice plate''}}
\end{minipage}
\begin{minipage}[c]{\subFigSz}
\caption*{{\scriptsize{\textcolor{red}{\it ``Tacos''}\\by GAE}}}
\end{minipage}
\begin{minipage}[c]{\subFigSz}
    \caption*{{\scriptsize{\textcolor{red}{\it ``Shrimp Tacos''}\\ by D-TIIL}}}
\end{minipage}
\begin{minipage}[c]{\subFigSz}
    \caption*{\scriptsize \it ``The History Museum\\displays a luxury car''}
\end{minipage}
\begin{minipage}[c]{\subFigSz}
    \caption*{{\scriptsize{ \textcolor{red}{\it ``car''}\\ by GAE}}}
\end{minipage}
\begin{minipage}[c]{\subFigSz}
    \caption*{{\scriptsize{\textcolor{red}{\it ``luxury car''}\\ by D-TIIL}}}
\end{minipage}
\vspace{-0.25in}
\caption{\em Comparison with GAE for detecting inconsistent mask.
\label{fig:gae}}
\vspace{-0.15in}
\end{figure}

\def\subFigSz{0.16\linewidth} %width=0.24\linewidth
\def\subFigH{0.16\linewidth}
\begin{figure}[t]
\centering
\includegraphics[width=\subFigSz, height=\subFigH]{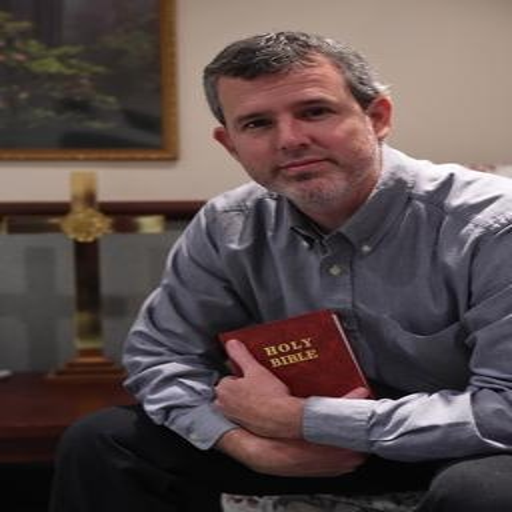}
\includegraphics[width=\subFigSz, height=\subFigH]{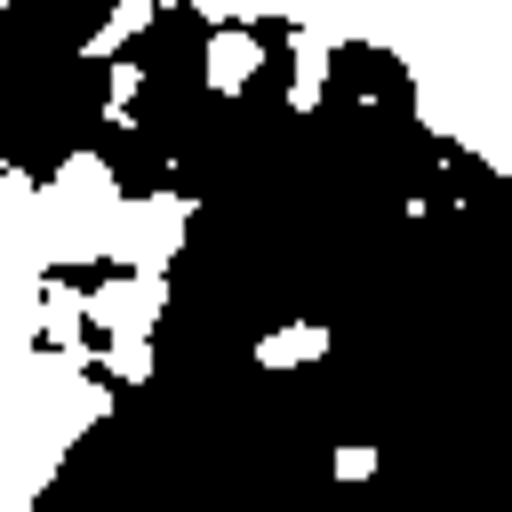}
\includegraphics[width=\subFigSz, height=\subFigH]{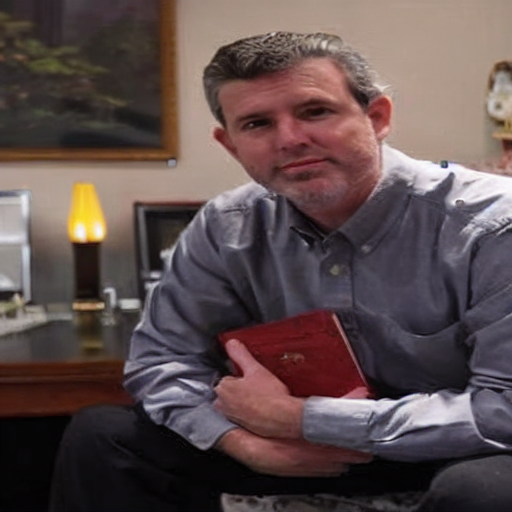}
\includegraphics[width=\subFigSz, height=\subFigH]{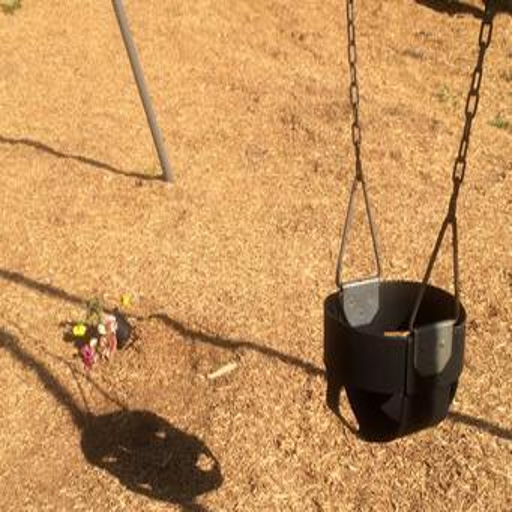}
\includegraphics[width=\subFigSz, height=\subFigH]{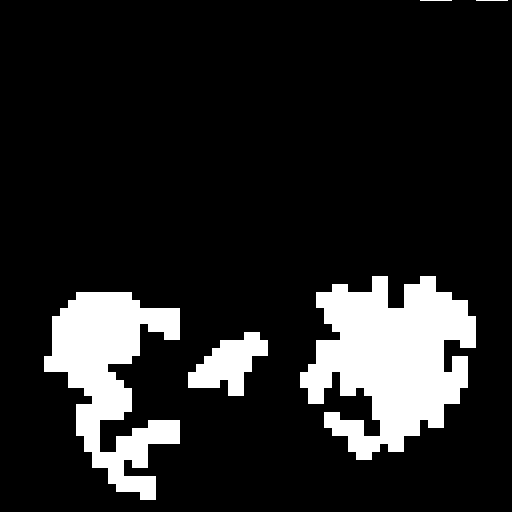}
\includegraphics[width=\subFigSz, height=\subFigH]{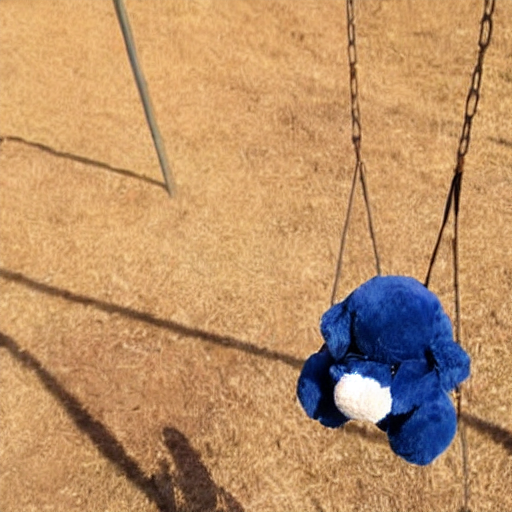}
% \vspace{-0.4in}
 \\
% \begin{minipage}[c]{\subFigSz}
%     \caption*{{\tiny{Consistent text: Rev Nace Lanier poses for a portrait in his office at Ronald Reagan National Airport. \\ \\ \\ \\ 
% }}}
% \end{minipage}

% \vspace{-0.4in}
\begin{minipage}[c]{\subFigSz}
    \caption*{\scriptsize (a) Original image}
\end{minipage}
\begin{minipage}[c]{\subFigSz}
    \caption*{{\scriptsize{ $\mM$}}}
\end{minipage}
\begin{minipage}[c]{\subFigSz}
    \caption*{{\scriptsize{ $I_{edt}$}}}
\end{minipage}
\begin{minipage}[c]{\subFigSz}
    \caption*{\scriptsize (b) Original image}
\end{minipage}
\begin{minipage}[c]{\subFigSz}
    \caption*{{\scriptsize{$\mM$}}}
\end{minipage}
\begin{minipage}[c]{\subFigSz}
    \caption*{{\scriptsize{  $I_{edt}$}}}
\end{minipage}
\vspace{-0.2cm}

\begin{minipage}[c]{0.45\linewidth}
\vspace{-0.2cm}
    \caption*{\scriptsize \it Text (a): ``Rev Nace Lanier poses for a portrait in his \textcolor{red}{office} at Ronald Reagan National Airport''}
\end{minipage}
\hspace{0.3in}
\begin{minipage}[c]{0.45\linewidth}
\vspace{-0.2cm}
    \caption*{\scriptsize \it (b): ``A \textcolor{red}{stuffed animal} and flowers rest near a white swing in Wills Memorial Park in La Plata MD where JiAire died last May''}
\end{minipage}
\vspace{-1.9em}
\caption{\em \small Failure cases of D-TIIL on TIIL dataset. (a) is a consistent image-text pair and (b) is an inconsistent image-text pair. Detected inconsistent words are highlighted in \textcolor{red}{red}.
\label{fig:fail}}
 \vspace{-2em}
\end{figure}

% \myheading{Consistency from }
%%%%%%%%%%%%%%%%%%%%%%%%%%%%%%%%%%%%%%%%%%%%%%%%%%%%%%%%%%%%
% mIoU ours 0.30531689523826433
\vspace{-0.1cm}
\section{Conclusion}
\vspace{-0.1cm}
%   \end{center}
In this work, we describe D-TIIL to expose text-image inconsistency by employing diffusion models as an omniscient, impartial evaluator to learn the semantic connections between textual and visual information. Instead of using a binary classification model, D-TIIL offers interpretable evidence by determining the inconsistency score of an image-text pair and pinpointing potential areas where the text and image semantics disagree. We also provide a new dataset, TIIL, built on real-world image-text pairs, for evaluating our D-TIIL method. Experimental evaluations of D-TIIL on this dataset demonstrate improved and more explainable results than the previous methods.
%In this work, we describe D-TIIL to detect image inconsistency using large language image models as an impartial, universal evaluator to assist in learning the semantic connections between textual and visual information. D-TIIL determines the inconsistency score of a image-text pair and pinpoints potential areas where the text and image semantics disagree. We also provide a new dataset, TIIL, built on real-world image-text pairs from the Visual News dataset, for evaluating our image inconsistency detection method. Experimental evaluations of D-TIIL on this dataset demonstrate improved and more explainable results than the previous methods.
There are a few directions we would like to enhance the current work in the future. 
% Current D-TIIL uses separate representations for texts and images. Recently there are new methods (\eg, ImageBind~\citep{girdhar2023imagebind}) to build joint embedding across different modalities including texts and images. It would be interesting to explore such joint representations for image inconsistency. 
Given the limited prior knowledge of the diffusion model we used, our model may not effectively handle the inconsistencies with respect to specific external knowledge. One potential solution is to replace our general foundation diffusion model with domain-specialized diffusion models to learn semantic connections specific to those domains (e.g., a text-to-image diffusion model trained on a Fashion dataset~\citep{sun2023sgdiff,karras2023dreampose} that generates fashion images would identify inconsistencies in fashion-related text-image pairs such as mismatched brands or styles). Furthermore, it is important to continue to enlarge our dataset with more recent text-prompt image generation models. 
%%%%%%%%%%%%%%%%%%%%%%%%%%%%%%%%%%%%%%%%%%%%%%%%%%%%%%%%%%%%

\vspace{0.1cm}
\myheading{\bf Ethics Statement}. This work is relevant to the fight against misinformation, which is a vexing problem that reduces the integrity of online information. While our method can effectively expose misinformation created with text-image inconsistency, there is a risk that the misinformation creator may use our method to select more deceptive text and image pairs, for instance, only use those that pass our algorithm. The mitigation to such abuse is to continue improving the algorithm and only provide its access to trustworthy parties. We will only release our code as open-source with the condition that it must not distribute harmful, offensive, dehumanizing content or otherwise harmful representations of people or their environments, cultures, religions, etc. produced with the model weights.

\myheading{Acknowledgement}. This work was supported in part by the US Defense Advanced Research Projects Agency (DARPA) Semantic Forensic (SemaFor) program, under Contract No. HR001120C0123, and National Science Foundation (NSF) Project SaTC-2153112. The views and conclusions contained herein are those of the authors and should not be interpreted as necessarily representing the official policies, either expressed or implied, of DARPA, NSF, or the U.S. Government.

\newpage
\bibliography{iclr2024_conference}

\begin{thebibliography}{44}
\providecommand{\natexlab}[1]{#1}
\providecommand{\url}[1]{\texttt{#1}}
\expandafter\ifx\csname urlstyle\endcsname\relax
  \providecommand{\doi}[1]{doi: #1}\else
  \providecommand{\doi}{doi: \begingroup \urlstyle{rm}\Url}\fi

\bibitem[Abdelnabi et~al.(2022)Abdelnabi, Hasan, and Fritz]{abdelnabi2022open}
Sahar Abdelnabi, Rakibul Hasan, and Mario Fritz.
\newblock Open-domain, content-based, multi-modal fact-checking of
  out-of-context images via online resources.
\newblock In \emph{Proceedings of the IEEE/CVF Conference on Computer Vision
  and Pattern Recognition}, pp.\  14940--14949, 2022.

\bibitem[Ali(2020)]{ali2020combatting}
Sana Ali.
\newblock Combatting against covid-19 \& misinformation: A systematic review.
\newblock \emph{Human Arenas}, pp.\  1--16, 2020.

\bibitem[Aneja et~al.(2021)Aneja, Bregler, and Nie{\ss}ner]{aneja2021cosmos}
Shivangi Aneja, Chris Bregler, and Matthias Nie{\ss}ner.
\newblock Cosmos: Catching out-of-context misinformation with
  self-supervisho2022classifiered learning.
\newblock \emph{arXiv preprint arXiv:2101.06278}, 2021.

\bibitem[Chefer et~al.(2021)Chefer, Gur, and Wolf]{chefer2021generic}
Hila Chefer, Shir Gur, and Lior Wolf.
\newblock Generic attention-model explainability for interpreting bi-modal and
  encoder-decoder transformers.
\newblock In \emph{Proceedings of the IEEE/CVF International Conference on
  Computer Vision}, pp.\  397--406, 2021.

\bibitem[Couairon et~al.(2023)Couairon, Verbeek, Schwenk, and
  Cord]{couairon2023diffedit}
Guillaume Couairon, Jakob Verbeek, Holger Schwenk, and Matthieu Cord.
\newblock Diffedit: Diffusion-based semantic image editing with mask guidance.
\newblock In \emph{The Eleventh International Conference on Learning
  Representations}, 2023.

\bibitem[Everingham et~al.(2015)Everingham, Eslami, Van~Gool, Williams, Winn,
  and Zisserman]{everingham2015pascal}
Mark Everingham, SM~Ali Eslami, Luc Van~Gool, Christopher~KI Williams, John
  Winn, and Andrew Zisserman.
\newblock The pascal visual object classes challenge: A retrospective.
\newblock \emph{International journal of computer vision}, 111:\penalty0
  98--136, 2015.

\bibitem[Fan et~al.(2021)Fan, Lai, Huang, Wei, Chai, Luo, and
  Wei]{fan2021rethinking}
Mingyuan Fan, Shenqi Lai, Junshi Huang, Xiaoming Wei, Zhenhua Chai, Junfeng
  Luo, and Xiaolin Wei.
\newblock Rethinking bisenet for real-time semantic segmentation.
\newblock In \emph{Proceedings of the IEEE/CVF conference on computer vision
  and pattern recognition}, pp.\  9716--9725, 2021.

\bibitem[Gao et~al.(2022)Gao, Li, Yang, Cheng, Han, and Torr]{gao2022large}
Shanghua Gao, Zhong-Yu Li, Ming-Hsuan Yang, Ming-Ming Cheng, Junwei Han, and
  Philip Torr.
\newblock Large-scale unsupervised semantic segmentation.
\newblock \emph{IEEE Transactions on Pattern Analysis and Machine
  Intelligence}, 2022.

\bibitem[Ho \& Salimans(2022)Ho and Salimans]{ho2022classifier}
Jonathan Ho and Tim Salimans.
\newblock Classifier-free diffusion guidance.
\newblock 2022.

\bibitem[Ho et~al.(2020)Ho, Jain, and Abbeel]{ho2020denoising}
Jonathan Ho, Ajay Jain, and Pieter Abbeel.
\newblock Denoising diffusion probabilistic models.
\newblock \emph{Advances in Neural Information Processing Systems},
  33:\penalty0 6840--6851, 2020.

\bibitem[Huang et~al.(2022)Huang, Jia, Chang, and Lyu]{huang2022text}
Mingzhen Huang, Shan Jia, Ming-Ching Chang, and Siwei Lyu.
\newblock Text-image de-contextualization detection using vision-language
  models.
\newblock In \emph{ICASSP 2022-2022 IEEE International Conference on Acoustics,
  Speech and Signal Processing (ICASSP)}, pp.\  8967--8971. IEEE, 2022.

\bibitem[Jaiswal et~al.(2017)Jaiswal, Sabir, AbdAlmageed, and
  Natarajan]{jaiswal2017multimedia}
Ayush Jaiswal, Ekraam Sabir, Wael AbdAlmageed, and Premkumar Natarajan.
\newblock Multimedia semantic integrity assessment using joint embedding of
  images and text.
\newblock In \emph{Proceedings of the 25th ACM international conference on
  Multimedia}, pp.\  1465--1471, 2017.

\bibitem[Karras et~al.(2023)Karras, Holynski, Wang, and
  Kemelmacher-Shlizerman]{karras2023dreampose}
Johanna Karras, Aleksander Holynski, Ting-Chun Wang, and Ira
  Kemelmacher-Shlizerman.
\newblock {Dreampose: Fashion image-to-video synthesis via stable diffusion}.
\newblock In \emph{ICCV}, 2023.

\bibitem[Kawar et~al.(2023)Kawar, Zada, Lang, Tov, Chang, Dekel, Mosseri, and
  Irani]{kawar2023imagic}
Bahjat Kawar, Shiran Zada, Oran Lang, Omer Tov, Huiwen Chang, Tali Dekel, Inbar
  Mosseri, and Michal Irani.
\newblock Imagic: Text-based real image editing with diffusion models.
\newblock In \emph{Proceedings of the IEEE/CVF Conference on Computer Vision
  and Pattern Recognition}, pp.\  6007--6017, 2023.

\bibitem[Kenton \& Toutanova(2019)Kenton and Toutanova]{kenton2019bert}
Jacob Devlin Ming-Wei~Chang Kenton and Lee~Kristina Toutanova.
\newblock Bert: Pre-training of deep bidirectional transformers for language
  understanding.
\newblock In \emph{Proceedings of naacL-HLT}, volume~1, pp.\ ~2, 2019.

\bibitem[Khattar et~al.(2019)Khattar, Goud, Gupta, and Varma]{khattar2019mvae}
Dhruv Khattar, Jaipal~Singh Goud, Manish Gupta, and Vasudeva Varma.
\newblock Mvae: Multimodal variational autoencoder for fake news detection.
\newblock In \emph{The world wide web conference}, pp.\  2915--2921, 2019.

\bibitem[Klingner et~al.(2021)Klingner, Bar, Mross, and
  Fingscheidt]{klingner2021improving}
Marvin Klingner, Andreas Bar, Marcel Mross, and Tim Fingscheidt.
\newblock Improving online performance prediction for semantic segmentation.
\newblock In \emph{Proceedings of the IEEE/CVF Conference on Computer Vision
  and Pattern Recognition}, pp.\  1--11, 2021.

\bibitem[Lee \& Choi(2019)Lee and Choi]{lee2019image}
Kiljae Lee and Jungsil Choi.
\newblock Image-text inconsistency effect on product evaluation in online
  retailing.
\newblock \emph{Journal of Retailing and Consumer Services}, 49:\penalty0
  279--288, 2019.

\bibitem[Li et~al.(2023)Li, Liu, Wu, Mu, Yang, Gao, Li, and Lee]{li2023gligen}
Yuheng Li, Haotian Liu, Qingyang Wu, Fangzhou Mu, Jianwei Yang, Jianfeng Gao,
  Chunyuan Li, and Yong~Jae Lee.
\newblock Gligen: Open-set grounded text-to-image generation.
\newblock In \emph{Proceedings of the IEEE/CVF Conference on Computer Vision
  and Pattern Recognition}, pp.\  22511--22521, 2023.

\bibitem[Liu et~al.(2020)Liu, Wang, Wang, and Ordonez]{liu2020visual}
Fuxiao Liu, Yinghan Wang, Tianlu Wang, and Vicente Ordonez.
\newblock Visual news: Benchmark and challenges in news image captioning.
\newblock \emph{arXiv preprint arXiv:2010.03743}, 2020.

\bibitem[Lugmayr et~al.(2022)Lugmayr, Danelljan, Romero, Yu, Timofte, and
  Van~Gool]{lugmayr2022repaint}
Andreas Lugmayr, Martin Danelljan, Andres Romero, Fisher Yu, Radu Timofte, and
  Luc Van~Gool.
\newblock Repaint: Inpainting using denoising diffusion probabilistic models.
\newblock In \emph{Proceedings of the IEEE/CVF Conference on Computer Vision
  and Pattern Recognition}, pp.\  11461--11471, 2022.

\bibitem[Luo et~al.(2021)Luo, Darrell, and Rohrbach]{luo2021newsclippings}
Grace Luo, Trevor Darrell, and Anna Rohrbach.
\newblock Newsclippings: Automatic generation of out-of-context multimodal
  media.
\newblock \emph{arXiv preprint arXiv:2104.05893}, 2021.

\bibitem[McCrae et~al.(2021)McCrae, Wang, and Zakhor]{mccrae2021multi}
Scott McCrae, Kehan Wang, and Avideh Zakhor.
\newblock Multi-modal semantic inconsistency detection in social media news
  posts.
\newblock \emph{arXiv preprint arXiv:2105.12855}, 2021.

\bibitem[McCrae et~al.(2022)McCrae, Wang, and Zakhor]{mccrae2022multi}
Scott McCrae, Kehan Wang, and Avideh Zakhor.
\newblock Multi-modal semantic inconsistency detection in social media news
  posts.
\newblock In \emph{MultiMedia Modeling: 28th International Conference, MMM
  2022, Phu Quoc, Vietnam, June 6--10, 2022, Proceedings, Part II}, pp.\
  331--343. Springer, 2022.

\bibitem[Molina et~al.(2021)Molina, Sundar, Le, and Lee]{molina2021fake}
Maria~D Molina, S~Shyam Sundar, Thai Le, and Dongwon Lee.
\newblock “fake news” is not simply false information: a concept
  explication and taxonomy of online content.
\newblock \emph{American behavioral scientist}, 65\penalty0 (2):\penalty0
  180--212, 2021.

\bibitem[Nichol et~al.(2021)Nichol, Dhariwal, Ramesh, Shyam, Mishkin, McGrew,
  Sutskever, and Chen]{nichol2021glide}
Alex Nichol, Prafulla Dhariwal, Aditya Ramesh, Pranav Shyam, Pamela Mishkin,
  Bob McGrew, Ilya Sutskever, and Mark Chen.
\newblock Glide: Towards photorealistic image generation and editing with
  text-guided diffusion models.
\newblock \emph{arXiv preprint arXiv:2112.10741}, 2021.

\bibitem[Nichol \& Dhariwal(2021)Nichol and Dhariwal]{nichol2021improved}
Alexander~Quinn Nichol and Prafulla Dhariwal.
\newblock Improved denoising diffusion probabilistic models.
\newblock In \emph{International Conference on Machine Learning}, pp.\
  8162--8171. PMLR, 2021.

\bibitem[Popat et~al.(2018)Popat, Mukherjee, Yates, and
  Weikum]{popat2018declare}
Kashyap Popat, Subhabrata Mukherjee, Andrew Yates, and Gerhard Weikum.
\newblock Declare: Debunking fake news and false claims using evidence-aware
  deep learning.
\newblock \emph{arXiv preprint arXiv:1809.06416}, 2018.

\bibitem[Qi et~al.(2021)Qi, Cao, Li, Liu, Sheng, Mi, He, Lv, Guo, and
  Yu]{qi2021improving}
Peng Qi, Juan Cao, Xirong Li, Huan Liu, Qiang Sheng, Xiaoyue Mi, Qin He,
  Yongbiao Lv, Chenyang Guo, and Yingchao Yu.
\newblock Improving fake news detection by using an entity-enhanced framework
  to fuse diverse multimodal clues.
\newblock In \emph{Proceedings of the 29th ACM International Conference on
  Multimedia}, pp.\  1212--1220, 2021.

\bibitem[Radford et~al.(2021{\natexlab{a}})Radford, Kim, Hallacy, Ramesh, Goh,
  Agarwal, Sastry, Askell, Mishkin, Clark, et~al.]{CLIP_radford2021}
Alec Radford, Jong~Wook Kim, Chris Hallacy, Aditya Ramesh, Gabriel Goh,
  Sandhini Agarwal, Girish Sastry, Amanda Askell, Pamela Mishkin, Jack Clark,
  et~al.
\newblock Learning transferable visual models from natural language
  supervision.
\newblock In \emph{International conference on machine learning}, pp.\
  8748--8763. PMLR, 2021{\natexlab{a}}.

\bibitem[Radford et~al.(2021{\natexlab{b}})Radford, Kim, Hallacy, Ramesh, Goh,
  Agarwal, Sastry, Askell, Mishkin, Clark, et~al.]{radford2021learning}
Alec Radford, Jong~Wook Kim, Chris Hallacy, Aditya Ramesh, Gabriel Goh,
  Sandhini Agarwal, Girish Sastry, Amanda Askell, Pamela Mishkin, Jack Clark,
  et~al.
\newblock Learning transferable visual models from natural language
  supervision.
\newblock \emph{arXiv preprint arXiv:2103.00020}, 2021{\natexlab{b}}.

\bibitem[Ramesh et~al.(2022)Ramesh, Dhariwal, Nichol, Chu, and
  Chen]{DALLE2ramesh2022hiera}
Aditya Ramesh, Prafulla Dhariwal, Alex Nichol, Casey Chu, and Mark Chen.
\newblock Hierarchical text-conditional image generation with clip latents.
\newblock \emph{arXiv preprint arXiv:2204.06125}, 2022.

\bibitem[Rombach et~al.(2022)Rombach, Blattmann, Lorenz, Esser, and
  Ommer]{rombach2022high}
Robin Rombach, Andreas Blattmann, Dominik Lorenz, Patrick Esser, and Bj{\"o}rn
  Ommer.
\newblock High-resolution image synthesis with latent diffusion models.
\newblock In \emph{Proceedings of the IEEE/CVF Conference on Computer Vision
  and Pattern Recognition}, pp.\  10684--10695, 2022.

\bibitem[Ronneberger et~al.(2015)Ronneberger, Fischer, and
  Brox]{ronneberger2015u}
Olaf Ronneberger, Philipp Fischer, and Thomas Brox.
\newblock U-net: Convolutional networks for biomedical image segmentation.
\newblock In \emph{Medical Image Computing and Computer-Assisted
  Intervention--MICCAI 2015: 18th International Conference, Munich, Germany,
  October 5-9, 2015, Proceedings, Part III 18}, pp.\  234--241. Springer, 2015.

\bibitem[Sabir et~al.(2018)Sabir, AbdAlmageed, Wu, and
  Natarajan]{sabir2018deep}
Ekraam Sabir, Wael AbdAlmageed, Yue Wu, and Prem Natarajan.
\newblock Deep multimodal image-repurposing detection.
\newblock In \emph{Proceedings of the 26th ACM international conference on
  Multimedia}, pp.\  1337--1345, 2018.

\bibitem[Saharia et~al.(2022)Saharia, Chan, Saxena, Li, Whang, Denton,
  Ghasemipour, Ayan, Mahdavi, Lopes, et~al.]{Imagen_saharia2022phot}
Chitwan Saharia, William Chan, Saurabh Saxena, Lala Li, Jay Whang, Emily
  Denton, Seyed Kamyar~Seyed Ghasemipour, Burcu~Karagol Ayan, S~Sara Mahdavi,
  Rapha~Gontijo Lopes, et~al.
\newblock Photorealistic text-to-image diffusion models with deep language
  understanding.
\newblock \emph{arXiv preprint arXiv:2205.11487}, 2022.

\bibitem[Song et~al.(2020)Song, Meng, and Ermon]{song2020denoising}
Jiaming Song, Chenlin Meng, and Stefano Ermon.
\newblock Denoising diffusion implicit models.
\newblock \emph{arXiv preprint arXiv:2010.02502}, 2020.

\bibitem[Sun et~al.(2023)Sun, Zhou, He, and Mok]{sun2023sgdiff}
Zhengwentai Sun, Yanghong Zhou, Honghong He, and PY~Mok.
\newblock Sgdiff: A style guided diffusion model for fashion synthesis.
\newblock 2023.

\bibitem[Tan et~al.(2020)Tan, Plummer, and Saenko]{tan2020detecting}
Reuben Tan, Bryan~A Plummer, and Kate Saenko.
\newblock Detecting cross-modal inconsistency to defend against neural fake
  news.
\newblock \emph{arXiv preprint arXiv:2009.07698}, 2020.

\bibitem[Xu et~al.(2022)Xu, Ouyang, Bennamoun, Boussaid, and Xu]{xu2022multi}
Lian Xu, Wanli Ouyang, Mohammed Bennamoun, Farid Boussaid, and Dan Xu.
\newblock Multi-class token transformer for weakly supervised semantic
  segmentation.
\newblock In \emph{Proceedings of the IEEE/CVF Conference on Computer Vision
  and Pattern Recognition}, pp.\  4310--4319, 2022.

\bibitem[Zeng et~al.(2023)Zeng, Wu, Li, Li, Huang, and Sha]{zeng2023correcting}
Zhi Zeng, Mingmin Wu, Guodong Li, Xiang Li, Zhongqiang Huang, and Ying Sha.
\newblock Correcting the bias: Mitigating multimodal inconsistency contrastive
  learning for multimodal fake news detection.
\newblock In \emph{2023 IEEE International Conference on Multimedia and Expo
  (ICME)}, pp.\  2861--2866. IEEE, 2023.

\bibitem[Zhang et~al.(2021)Zhang, Li, Hu, Yang, Zhang, Wang, Choi, and
  Gao]{zhang2021vinvl}
Pengchuan Zhang, Xiujun Li, Xiaowei Hu, Jianwei Yang, Lei Zhang, Lijuan Wang,
  Yejin Choi, and Jianfeng Gao.
\newblock Vinvl: Revisiting visual representations in vision-language models.
\newblock In \emph{Proceedings of the IEEE/CVF Conference on Computer Vision
  and Pattern Recognition}, pp.\  5579--5588, 2021.

\bibitem[Zhou et~al.(2022)Zhou, Girdhar, Joulin, Kr{\"a}henb{\"u}hl, and
  Misra]{zhou2022detecting}
Xingyi Zhou, Rohit Girdhar, Armand Joulin, Philipp Kr{\"a}henb{\"u}hl, and
  Ishan Misra.
\newblock Detecting twenty-thousand classes using image-level supervision.
\newblock In \emph{Computer Vision--ECCV 2022: 17th European Conference, Tel
  Aviv, Israel, October 23--27, 2022, Proceedings, Part IX}, pp.\  350--368.
  Springer, 2022.

\bibitem[Zlatkova et~al.(2019)Zlatkova, Nakov, and Koychev]{zlatkova2019fact}
Dimitrina Zlatkova, Preslav Nakov, and Ivan Koychev.
\newblock Fact-checking meets fauxtography: Verifying claims about images.
\newblock \emph{arXiv preprint arXiv:1908.11722}, 2019.

\end{thebibliography}
\bibliographystyle{iclr2024_conference}

\newpage

\appendix

\section*{Appendix}
In the Appendix, we provide more details about TIIL dataset and conduct additional ablation experiments.
% \section{Appendix}
\section{TIIL Dataset}
\vskip -0.2cm
We provide more annotation details and statistics about the TIIL dataset in this section. More examples in the dataset are shown in \Fref{fig:datasetex}.

\myheading{Annotation Details.}
The annotation process was carried out by a team of six annotators with professional background, including 1 postdoc from the research team, 3 graduate volunteers, and 2 undergraduate volunteers. These annotators possess a comprehensive understanding of the data annotation task, which comprised three steps: (1) selecting matched object-term pairs and inputting the target text prompt, (2) manipulating the image or text with exact instruction and (3) conducting data cleaning as the final step.
The generation of the TIIL dataset starts with a real image-text pair. The first annotation process involves identifying the corresponding visual regions and textual terms through human involvement. Initially, we automatically extract separate text terms with spaCy~\footnote{https://github.com/explosion/spaCy}, human annotators then select and annotate the visual region that corresponds to the matched text term. The mask is annotated with the CVAT annotation platform~\footnote{https://www.cvat.ai/}. Additionally, the annotators provide a target text prompt with the instruction that i) it should be inconsistent with the original text but match the context that may mislead the readers; ii) it should not share semantic overlap with the original term, for instance, replacing "Chiwetel Ejiofor" with "Brad Pitt" rather than "a male actor".
% The annotation process for each text-image pair takes approximately 40 seconds on average. In total, we invested 37 hours to complete this annotation step. 
By utilizing the selected object region and annotated prompt as inputs to the DALL-E2 model, we obtain three manipulated images for each image-text pair. The second phase of the annotation process focuses on data cleaning to ensure the accuracy and coherence of the dataset. Each human annotator carefully follows three steps for quality checking purposes: (1) assessing the image quality of the generated images, (2) evaluating image-text inconsistencies, and (3) refining the region masks to establish ground truth for pixel-level inconsistency masks as the generated objects may have different shapes. To maintain the highest annotation quality, a rigorous cross-validation procedure was implemented within the team. This involved multiple assessments of each image-text pair by different annotators. 
\myheading{Dataset Statistics.}
The images in TIIL dataset have resolutions ranging from 256x396 to 3744x3744 pixels, and on average, the consistent mask covers 44.73\% of the entire image area. There are a total of 7,101 consistent pairs in the dataset. Among them, 2,101 are composed of original images and text sourced from the Visual News dataset~\citep{liu2020visual} (referred to as $\left\{{I}, {T}\right\}$ in Section 4 of the main paper). The remaining pairs consist of images generated by DALL-E2~\citep{DALLE2ramesh2022hiera} along with their corresponding text (referred to as $\left\{{I_e}, {T_m}\right\}$). 
The dataset also includes 7,138 inconsistent pairs, out of which 2,101 pairs consist of original images with manipulated text (referred to as $\left\{{I}, {T_m}\right\}$), while the remaining pairs consist of images generated by DALL-E2~\citep{DALLE2ramesh2022hiera} paired with the original text (referred to as$\left\{{I_e}, {T}\right\}$). Regarding manipulation regions, the dataset contains 3,015 annotated inconsistent regions of large size (greater than $200\times200$), 1,548 annotated inconsistent regions of medium size (ranging between $100\times 100$ and $200\times 200$), and 474 annotated inconsistent regions of small size (smaller than $100\times 100$).
% \begin{table}[hb]
% \centering
% \setlength{\tabcolsep}{12pt} 
% \begin{tabular}{lcccc}
% \rowcolor{mygray}
%    &  Thres1  & Thres2 & Thres3 & Thres4 \\   \hline \thickhline
%   CLIP Score & 80.98 &  1.05 & 22.68 & 0.19 \\
% \hline
% \end{tabular}
% % \vskip -0.1in
% \label{tab:random}
% \captionsetup{justification=centering}
% \caption{\small \it Additional statistic of TIIL.}
% \vspace{0.1cm}
% \end{table}
\vskip -0.2cm

\def\subFigSz{0.24\linewidth} %width=0.24\linewidth
\def\subFigH{0.15\linewidth}
\def\subCapH{0.07\linewidth}
\begin{figure}[h]
\centering
\includegraphics[width=\subFigSz, height=\subFigH]{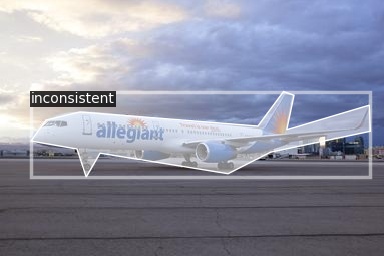}
\includegraphics[width=\subFigSz, height=\subFigH]{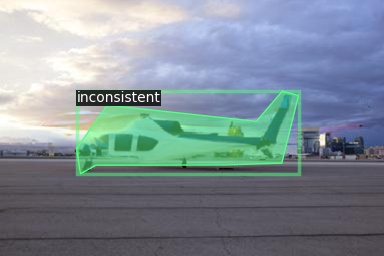}
\includegraphics[width=\subFigSz, height=\subFigH]{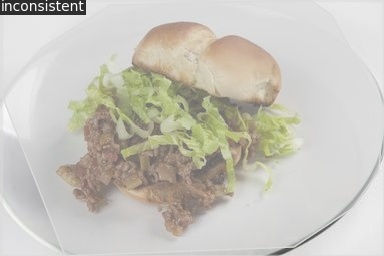}
\includegraphics[width=\subFigSz, height=\subFigH]{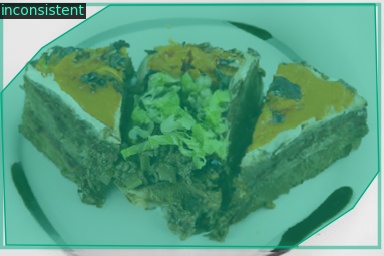}\\

\begin{minipage}[l]{0.48\linewidth}
    \caption*{\scriptsize \it {``An \textcolor{gray!20}{helicopter}/\textcolor{green!30}{Allegiant} Air is offering new flights at BWI''}}
\end{minipage}
\begin{minipage}[c]{0.48\linewidth}
    \caption*{\centering\scriptsize \it ``An \textcolor{gray!20}{cake}/\textcolor{green!20}{Asian Sloppy Joes}''}
\end{minipage}
\vskip -0.1in

\includegraphics[width=\subFigSz, height=\subFigH]{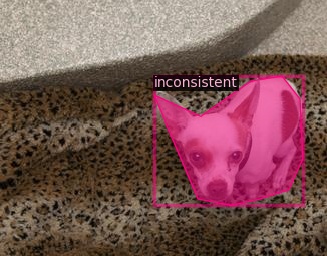}
\includegraphics[width=\subFigSz, height=\subFigH]{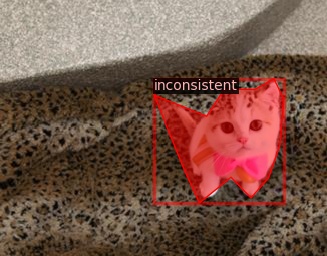}
\includegraphics[width=\subFigSz, height=\subFigH]{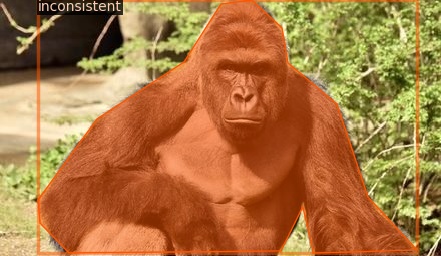}
\includegraphics[width=\subFigSz, height=\subFigH]{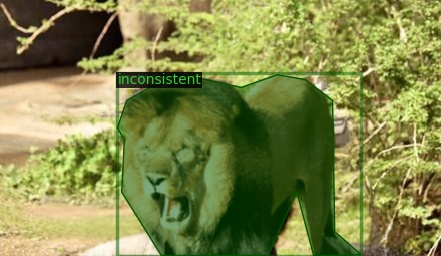}\\
\begin{minipage}[c]{0.48\linewidth} 
\captionsetup{justification=centering}
    \caption*{{\scriptsize \it ``One of the rescued cute\textcolor{magenta}{cat}/\textcolor{red!60}{dog} from a South Carolina \\ shelter that was in the path of Hurricane Matthew''}}
\end{minipage}
\begin{minipage}[c]{0.48\linewidth}
    \caption*{\centering \it \scriptsize {``Harambe was a 17-year-old \textcolor{orange}{lion}/\textcolor{lime}{gorilla} at the Cincinnati Zoo''}}
\end{minipage}
\vskip -0.1in

\includegraphics[width=\subFigSz, height=\subFigH]{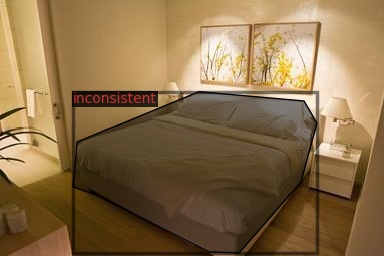}
\includegraphics[width=\subFigSz, height=\subFigH]{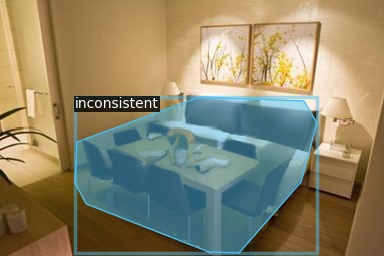}
\includegraphics[width=\subFigSz, height=\subFigH]{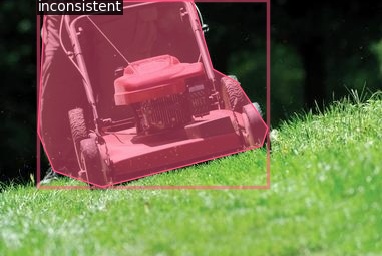}
\includegraphics[width=\subFigSz, height=\subFigH]{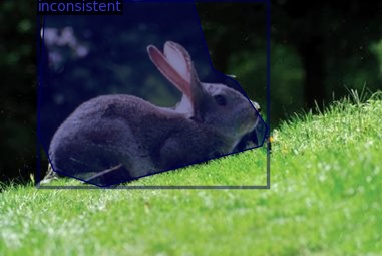}\\
\begin{minipage}[c]{0.48\linewidth} 
    \captionsetup{justification=centering}
    \caption*{{\scriptsize \it ``Of the 216 condos 45 percent of them have one\\ \textcolor{lightgray}{dining room}/\textcolor{cyan}{bedroom} The rest have two bedrooms''}}
\end{minipage}
\begin{minipage}[c]{0.48\linewidth}
\captionsetup{justification=centering}
    \caption*{\centering \it \scriptsize {``A \textcolor{red}{rabbit}/\textcolor{lightgray}{Montgomery lawnlovers} are up in arms''}}
\end{minipage}
\vskip -0.1in

\includegraphics[width=\subFigSz, height=\subFigH]{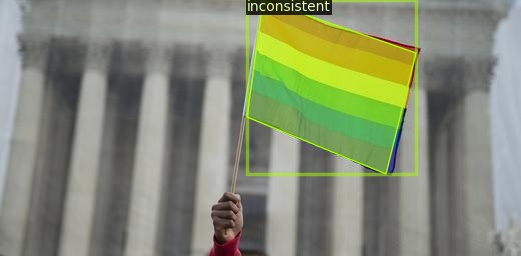}
\includegraphics[width=\subFigSz, height=\subFigH]{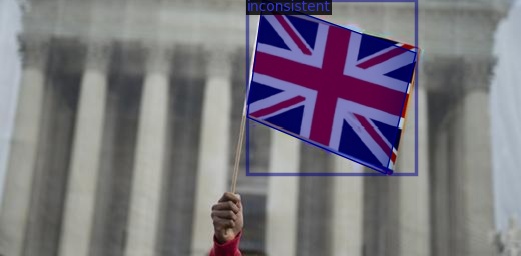}
\includegraphics[width=\subFigSz, height=\subFigH]{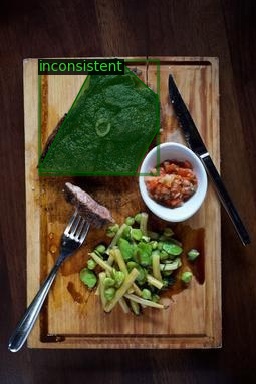}
\includegraphics[width=\subFigSz, height=\subFigH]{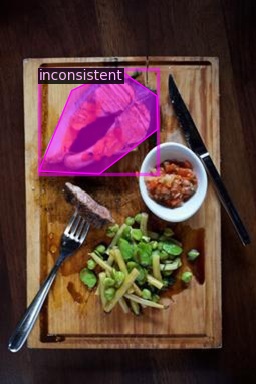}\\
\begin{minipage}[c]{0.48\linewidth} 
\captionsetup{justification=centering}
    \caption*{{\scriptsize \it ``A samesex marriage supporter waves a \textcolor{yellow}{UK Flag}/\textcolor{purple}{a rainbow flag} \\ in front of the US Supreme Court''}}
\end{minipage}
\begin{minipage}[c]{0.48\linewidth}
    \caption*{\centering \it \scriptsize {``On the menu at Wm Mulherin's Sons \textcolor{olive}{grilled fried salmon}/\textcolor{magenta}{lamb steak} with a salad of shell beans''}}
\end{minipage}

% \vspace{-0.1in}
\caption{\em \small Examples in TIIL dataset. Colored texts correspond to inconsistent regions of the same color in the image. The figure is best viewed in color.
\label{fig:datasetex}}
%\vspace{-0.2in}
\end{figure}

\section{Additional Ablation Studies}
% \vskip -0.2cm
To demonstrate the impact of mask binarization threshold and text embedding alignment initialization methods on the performance, we offer additional ablation studies in this section.

% \vskip 0.1in

% \vspace{-0.5cm}

\myheading{Text Embedding Alignment Initialization.} 
Instead of initializing the text embedding with random noise, we choose to initialize it with the embedding $E_0$ of the input text. In Table \ref{tab:emb} of the main paper, we have provided a comparison of different embedding initialization methods in terms of mIoU. Here, we further present a comparison between random initialization and our $E_0$-based initialization using CLIP scores to explain our motivation. 
Table \ref{tab:random} demonstrates that the semantic similarity between inconsistent text embeddings (i.e., the input $E_0$ and the text embedding corresponding to the input inconsistent image) is significantly higher than the similarity of random noise embeddings (i.e., the input $E_0$ and the randomly initialized embedding). This highlights the advantage of initializing with $E_0$, as it not only serves as a reference for optimizing $E$, but can also expedite the convergence during the alignment process.

\def\subFigSz{0.23\linewidth}
\def\subFigH{0.16\linewidth}
\captionsetup{type=figure} 
\begin{figure}[!ht]
\centering
\begin{minipage}[c]{0.02\linewidth}
    \caption*{\rotatebox{0}{\tiny{${I}$}}}
\end{minipage}\hfill
\begin{minipage}[c]{0.95\linewidth}
    \includegraphics[width=\subFigSz, height=\subFigH]{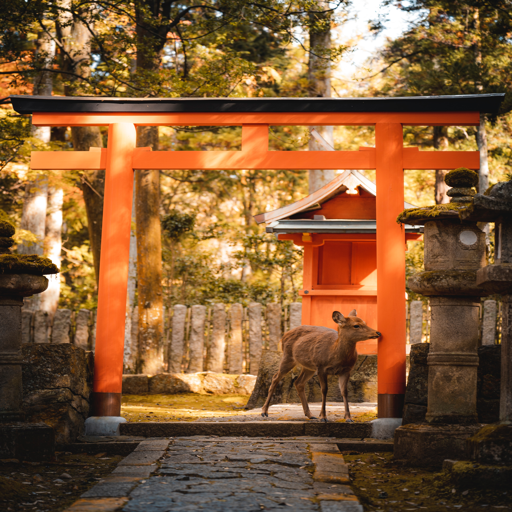}
    \includegraphics[width=\subFigSz, height=\subFigH]{Figure/thres/img13_ori.png}
    \includegraphics[width=\subFigSz, height=\subFigH]{Figure/thres/img13_ori.png}
    \includegraphics[width=\subFigSz, height=\subFigH]{Figure/thres/img13_ori.png}
\end{minipage}

\begin{minipage}[c]{0.02\linewidth}
    % \caption*{\rotatebox{0}{\tiny{$\theta$}}}
\end{minipage}\hfill
% \hspace{0.01in} 
\begin{minipage}[c]{0.95\linewidth}
\captionsetup{justification=centering}
\begin{minipage}[c]{\subFigSz}
    \caption*{\scriptsize \it Threshold: 0.1}
\end{minipage}
% \hspace{-0.01in}
\begin{minipage}[c]{\subFigSz}
    \caption*{\scriptsize \it Threshold: 0.2}
\end{minipage}
\begin{minipage}[c]{\subFigSz}
    \caption*{\scriptsize \it Threshold: 0.4}
\end{minipage}
\begin{minipage}[c]{\subFigSz}
    \caption*{\scriptsize \it Average threshold ($\sim$ 0.3)}
\end{minipage}
\end{minipage}
\vspace{-0.2cm}

\begin{minipage}[c]{0.02\linewidth}
    \caption*{\rotatebox{0}{\tiny{$\mM^{\prime}$}}}
\end{minipage}\hfill
\begin{minipage}[c]{0.95\linewidth}
    \includegraphics[width=\subFigSz, height=\subFigH]{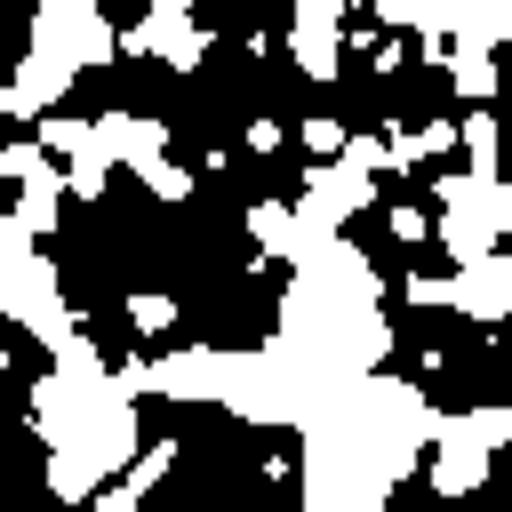}
    \includegraphics[width=\subFigSz, height=\subFigH]{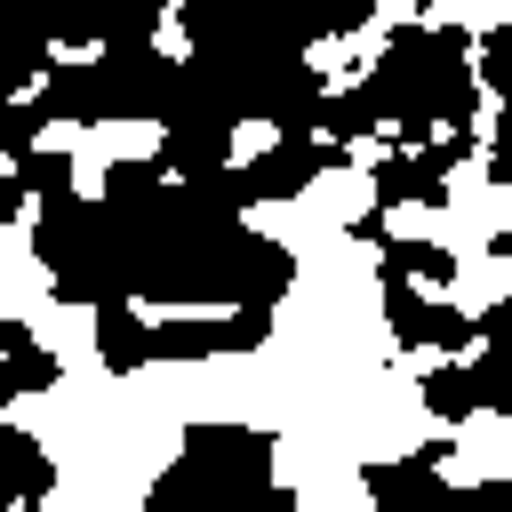}
    \includegraphics[width=\subFigSz, height=\subFigH]{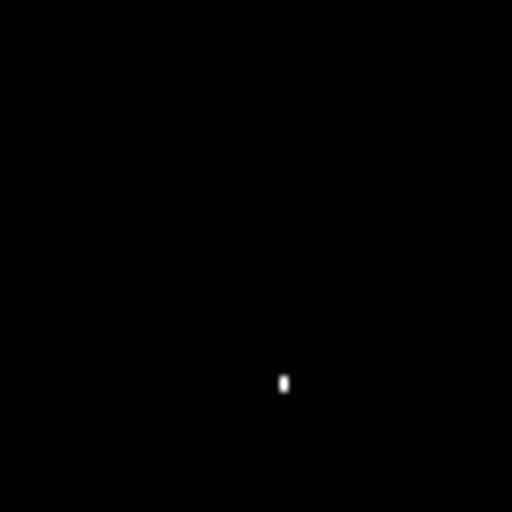}
    \includegraphics[width=\subFigSz, height=\subFigH]{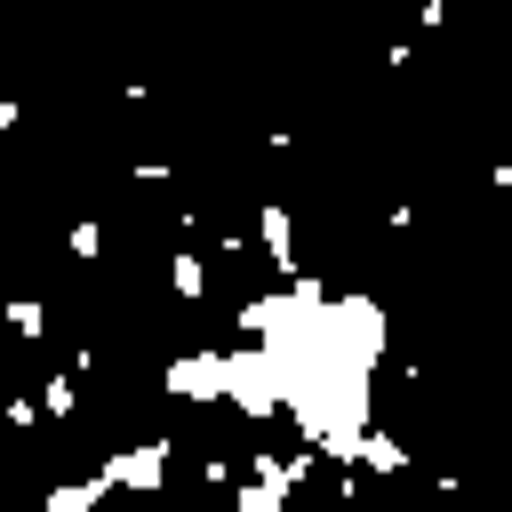}
\end{minipage}

\begin{minipage}[c]{0.02\linewidth}
    \caption*{\rotatebox{0}{\tiny{${I}_{edt}$}}}
\end{minipage}\hfill
\begin{minipage}[c]{0.95\linewidth}
    \includegraphics[width=\subFigSz, height=\subFigH]{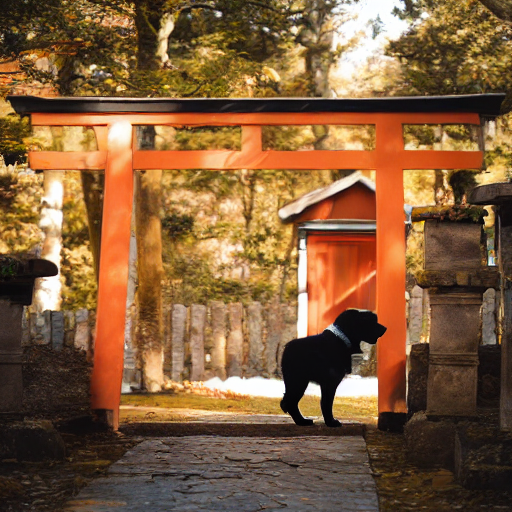}
    \includegraphics[width=\subFigSz, height=\subFigH]{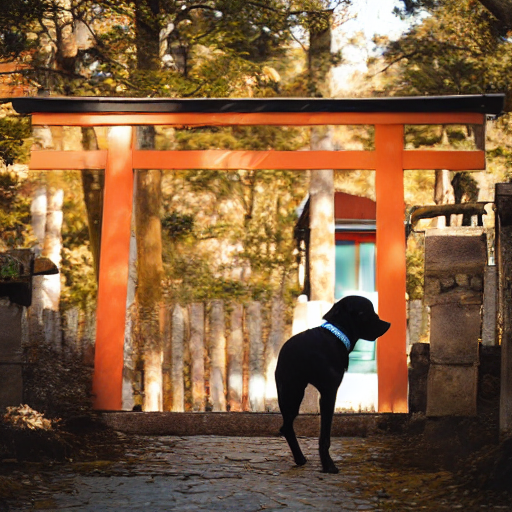}
    \includegraphics[width=\subFigSz, height=\subFigH]{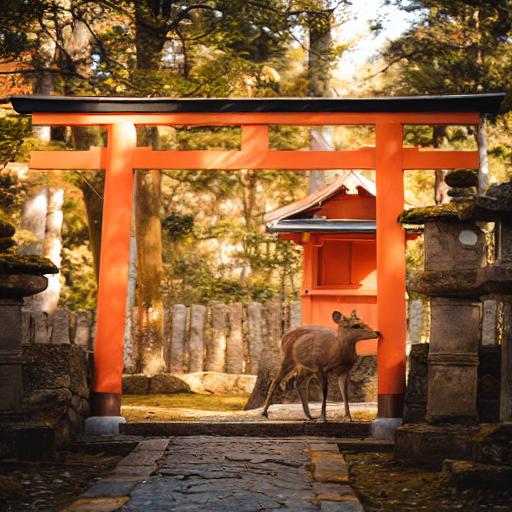}
    \includegraphics[width=\subFigSz, height=\subFigH]{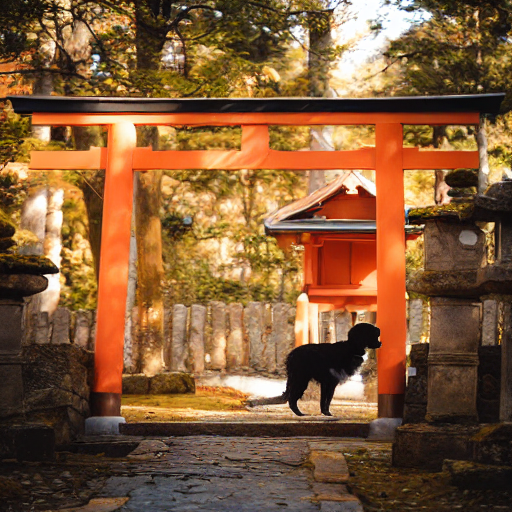}
\end{minipage}

\begin{minipage}[c]{0.02\linewidth}
    \caption*{\rotatebox{0}{\tiny{$\mM$}}}
\end{minipage}\hfill
\begin{minipage}[c]{0.95\linewidth}
    \includegraphics[width=\subFigSz, height=\subFigH]{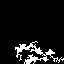}
    \includegraphics[width=\subFigSz, height=\subFigH]{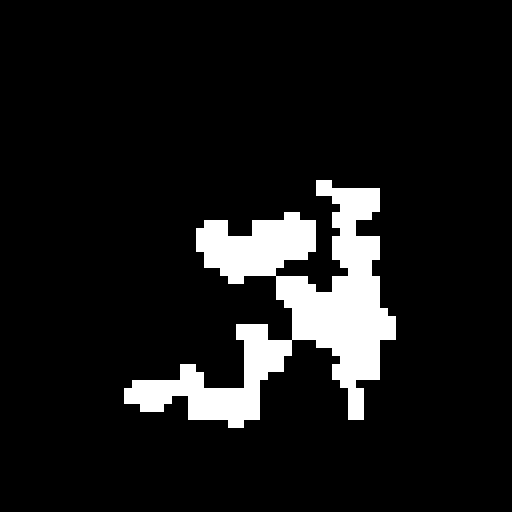}
    \includegraphics[width=\subFigSz, height=\subFigH]{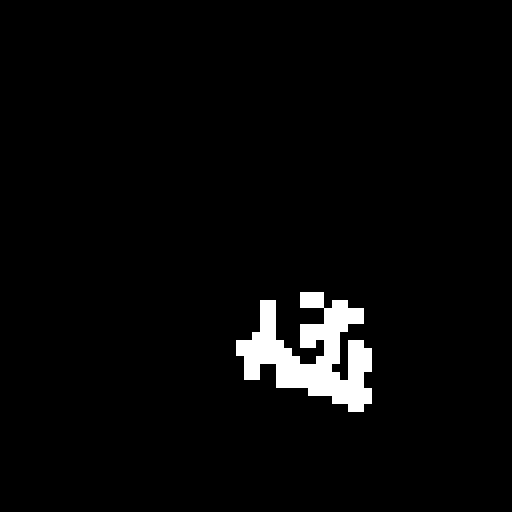}
    \includegraphics[width=\subFigSz, height=\subFigH]{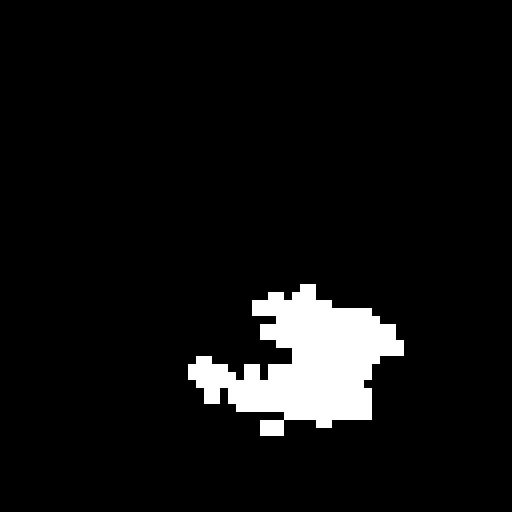}
\end{minipage}

\begin{minipage}[c]{0.02\linewidth}
    \caption*{\rotatebox{0}{\tiny{$r$}}}
\end{minipage}\hfill
% \hspace{0.01in}
\begin{minipage}[c]{0.95\linewidth}
\captionsetup{justification=centering}
\begin{minipage}[c]{\subFigSz}
    \caption*{\scriptsize \it Score: 24.3}
\end{minipage}
% \hspace{-0.01in}
\begin{minipage}[c]{\subFigSz}
    \caption*{\scriptsize \it Score: 20.1}
\end{minipage}
\begin{minipage}[c]{\subFigSz}
    \caption*{\scriptsize \it Score: 70.8}
\end{minipage}
\begin{minipage}[c]{\subFigSz}
    \caption*{\scriptsize \it Score: 11.9}
\end{minipage}
\end{minipage}
% \vspace{-0.3cm}

\begin{minipage}[c]{0.02\linewidth}
    \caption*{\rotatebox{0}{\tiny{${T}$}}}
\end{minipage}\hfill
% \hspace{0.01in}
\begin{minipage}[c]{0.95\linewidth}
\captionsetup{justification=centering}
\begin{minipage}[c]{\linewidth}
\vspace{-0.45cm}
    \caption*{\footnotesize \it ``A \textcolor{red}{dog} stood proudly, surrounded by the crisp autumn air''}
\end{minipage}
\end{minipage}

\begin{minipage}[c]{0.02\linewidth}
    \caption*{\rotatebox{0}{\tiny{${I}$}}}
\end{minipage}\hfill
\begin{minipage}[c]{0.95\linewidth}
    \includegraphics[width=\subFigSz, height=\subFigH]{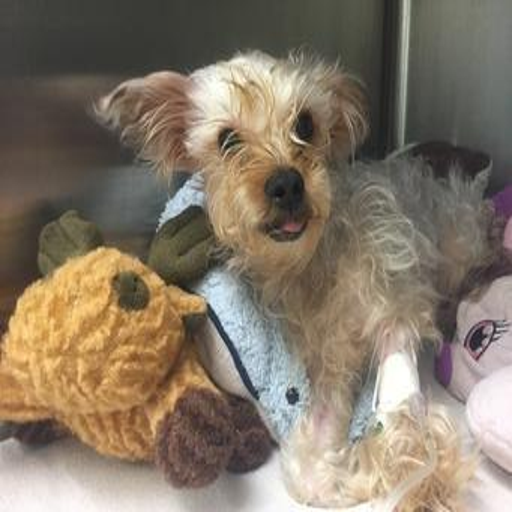}
    \includegraphics[width=\subFigSz, height=\subFigH]{Figure/thres/62133_ori.png}
    \includegraphics[width=\subFigSz, height=\subFigH]{Figure/thres/62133_ori.png}
    \includegraphics[width=\subFigSz, height=\subFigH]{Figure/thres/62133_ori.png}
\end{minipage}

\begin{minipage}[c]{0.02\linewidth}
    % \caption*{\rotatebox{0}{\tiny{$\theta$}}}
\end{minipage}\hfill
% \hspace{0.01in}
\begin{minipage}[c]{0.95\linewidth}
\captionsetup{justification=centering}
\begin{minipage}[c]{\subFigSz}
    \caption*{\scriptsize \it Threshold: 0.1}
\end{minipage}
% \hspace{-0.01in}
\begin{minipage}[c]{\subFigSz}
    \caption*{\scriptsize \it Threshold: 0.3}
\end{minipage}
\begin{minipage}[c]{\subFigSz}
    \caption*{\scriptsize \it Threshold: 0.4}
\end{minipage}
\begin{minipage}[c]{\subFigSz}
    \caption*{\scriptsize \it Average threshold ($\sim$ 0.2)}
\end{minipage}
\end{minipage}
\vspace{-0.2cm}

\begin{minipage}[c]{0.02\linewidth}
    \caption*{\rotatebox{0}{\tiny{$\mM^{\prime}$}}}
\end{minipage}\hfill
\begin{minipage}[c]{0.95\linewidth}
    \includegraphics[width=\subFigSz, height=\subFigH]{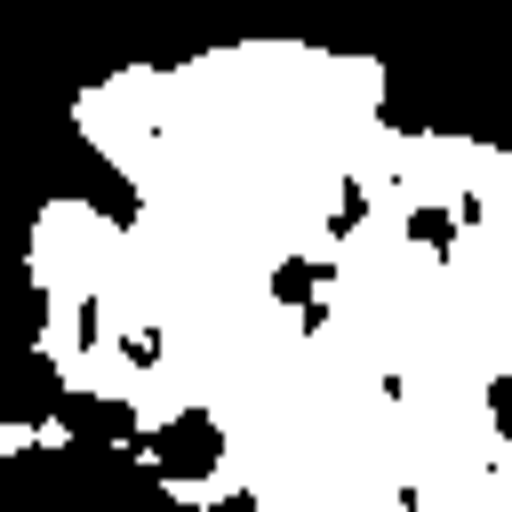}
    \includegraphics[width=\subFigSz, height=\subFigH]{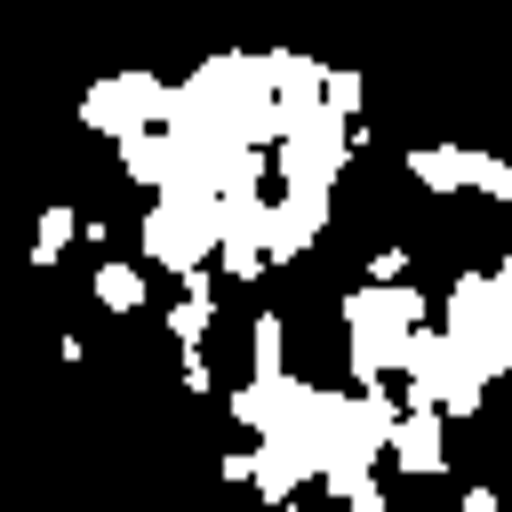}
    \includegraphics[width=\subFigSz, height=\subFigH]{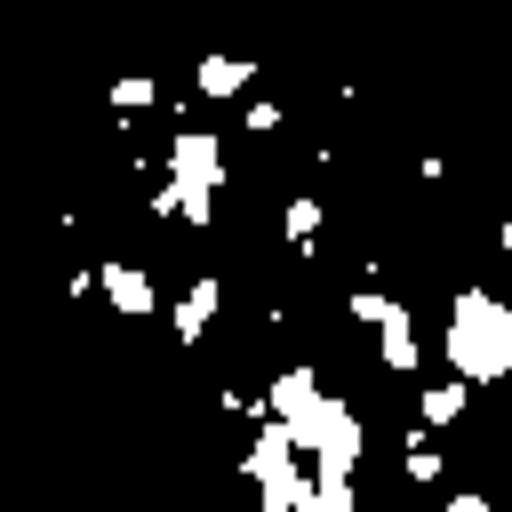}
    \includegraphics[width=\subFigSz, height=\subFigH]{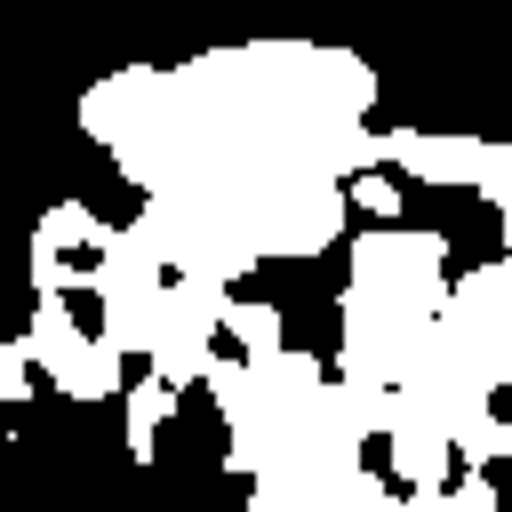}
\end{minipage}

\begin{minipage}[c]{0.02\linewidth}
    \caption*{\rotatebox{0}{\tiny{${I}_{edt}$}}}
\end{minipage}\hfill
\begin{minipage}[c]{0.95\linewidth}
    \includegraphics[width=\subFigSz, height=\subFigH]{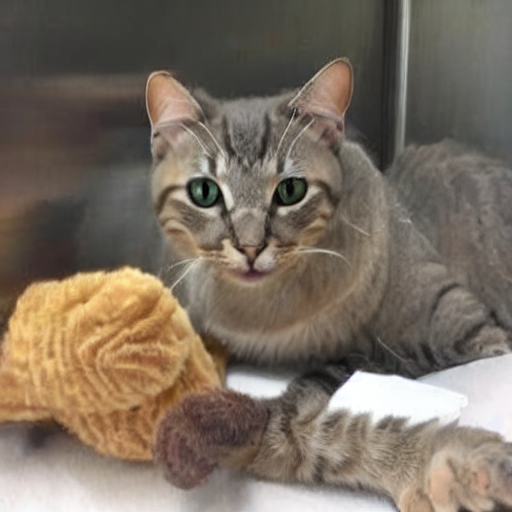}
    \includegraphics[width=\subFigSz, height=\subFigH]{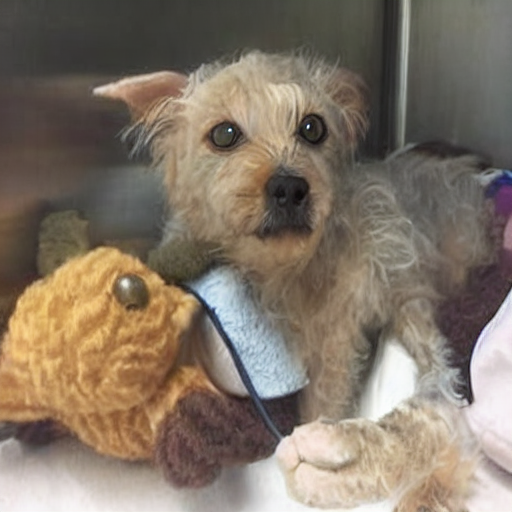}
    \includegraphics[width=\subFigSz, height=\subFigH]{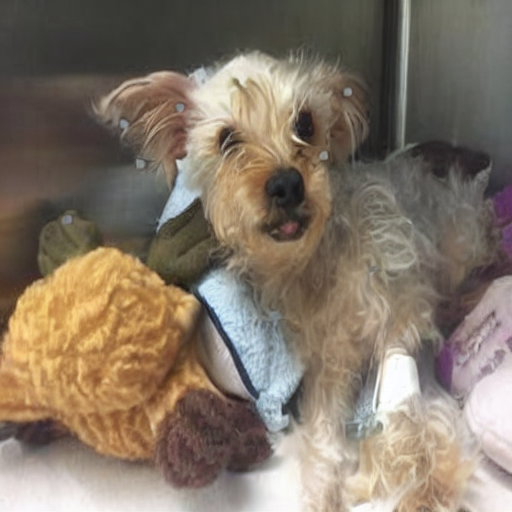}
    \includegraphics[width=\subFigSz, height=\subFigH]{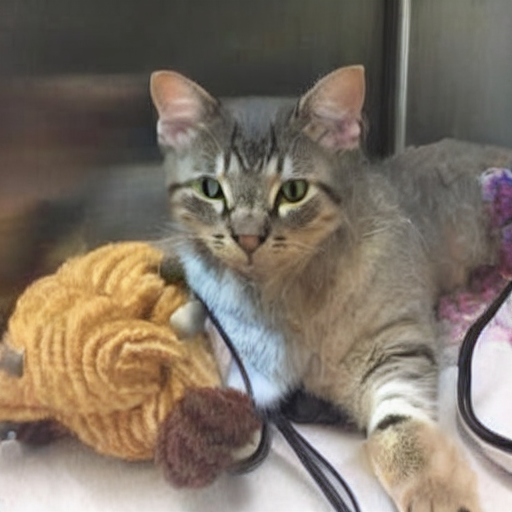}
\end{minipage}

\begin{minipage}[c]{0.02\linewidth}
    \caption*{\rotatebox{0}{\tiny{$\mM$}}}
\end{minipage}\hfill
\begin{minipage}[c]{0.95\linewidth}
    \includegraphics[width=\subFigSz, height=\subFigH]{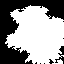}
    \includegraphics[width=\subFigSz, height=\subFigH]{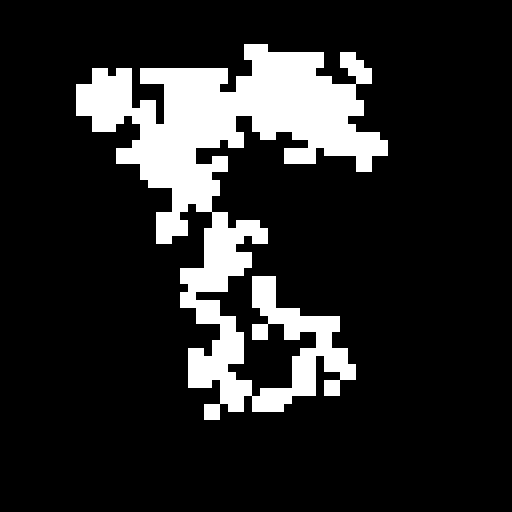} 
    \includegraphics[width=\subFigSz, height=\subFigH]{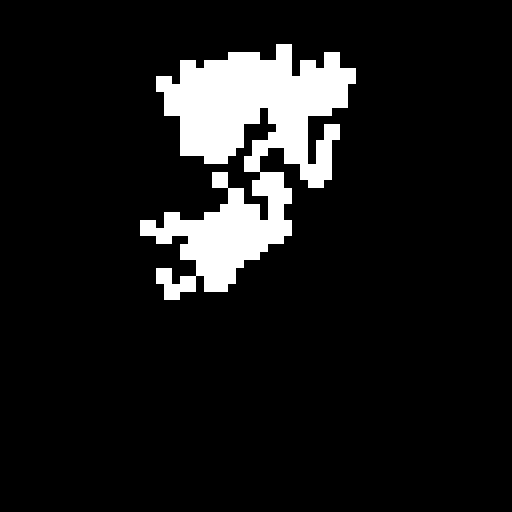}
    \includegraphics[width=\subFigSz, height=\subFigH]{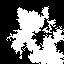}  
\end{minipage}

\begin{minipage}[c]{0.02\linewidth}
    \caption*{\rotatebox{0}{\tiny{$r$}}}
\end{minipage}\hfill
% \hspace{0.01in}
\begin{minipage}[c]{0.95\linewidth}
\captionsetup{justification=centering}
\begin{minipage}[c]{\subFigSz}
    \caption*{\scriptsize \it Score: 13.5}
\end{minipage}
% \hspace{-0.01in}
\begin{minipage}[c]{\subFigSz}
    \caption*{\scriptsize \it Score: 58.7}
\end{minipage}
\begin{minipage}[c]{\subFigSz}
    \caption*{\scriptsize \it Score: 66.1}
\end{minipage}
\begin{minipage}[c]{\subFigSz}
    \caption*{\scriptsize \it Score: 12.3}
\end{minipage}
\end{minipage}
% \vspace{-0.3cm}

\begin{minipage}[c]{0.02\linewidth}
    \caption*{\rotatebox{0}{\tiny{${T}$}}}
\end{minipage}\hfill
% \hspace{0.01in}
\begin{minipage}[c]{0.95\linewidth}
\captionsetup{justification=centering} 
\begin{minipage}[c]{\linewidth}
\vspace{-0.45cm}
    \caption*{\footnotesize \it ``A \textcolor{red}{cat} shown recovering will be adopted by a veterinary technician''}
\end{minipage}
\end{minipage}

\vspace{-0.3cm}
\caption{\em \small Comparison of D-TIIL on TIIL examples with different thresholds.} 
\label{fig:thres}
\vspace{-0.2in}
\end{figure}

\begin{table}[!ht]
\centering
\setlength{\tabcolsep}{12pt} 
\begin{tabular}{lccc}
\rowcolor{mygray}
  & Random Init  & Ours\\   \hline \thickhline
  CLIP Score  & 0.19  & 80.98 \\
\hline
\end{tabular}
% \vskip -0.1in
\captionsetup{justification=centering}
% \vspace{0.05cm}
\caption{\small \it Comparison of different text embedding initialization methods.}
\vspace{-0.3cm}
\label{tab:random}
\end{table}

\myheading{Comparison Between Image-manipulated and Text-manipulated Subsets.} As TIIL dataset is able to be categorized by two different kinds of manipulations: image-manipulated set and text-manipulated set, we further provide the performance of different methods on those two subsets for both localization (in \Tref{tab:loc_subset}) and detection (in \Tref{tab:det_subset}).
The text-changed samples achieved slightly worse performance than image-manipulated data, since the text changes (such as replacing a word or phrase) are less impactful than image changes (where various region sizes are manipulated), therefore more difficult to be detected.

\begin{table}[!ht]
\centering
\setlength{\tabcolsep}{12pt} 
\begin{tabular}{lccc}
\rowcolor{mygray}
  & Image-changed  & Text-changed\\   \hline \thickhline
  % CLIP Score  & 0.19  & 80.98 \\
  DetCLIP	&11.93	&15.21 \\
    GAE&28.57	&26.85 \\
    Ours&	50.28	&45.57 \\
\hline
\end{tabular}
% \vskip -0.1in
\captionsetup{justification=centering}
% \vspace{0.05cm}
\caption{\small \it Comparison of the localization performance in different samples.}
\label{tab:loc_subset}
\vspace{-0.3cm}
\end{table}

% Detection	
% \myheading{}
\begin{table}[!ht]
\centering
\setlength{\tabcolsep}{4pt} 
\begin{tabular}{lcccc}
\rowcolor{mygray}
  & AUC (image-changed)	&ACC (image-changed)	&AUC (text-changed)	&ACC (text-changed) \\ \thickhline
CLIP&	86.50	&79.11	&78.58	&71.87 \\
Ours	&88.80	&81.25	&82.29	&74.70 \\
\hline
\end{tabular}
% \vskip -0.1in
\captionsetup{justification=centering}
% \vspace{0.05cm}
\caption{\small \it Comparison of the detection performance in different samples.}
\label{tab:det_subset}
% \vspace{-0.3cm}
\end{table}
% We find that the CCN achieves an accuracy of 80.12\% on the subset composed solely of original images/text, which was sourced from the Internet, while a much lower accuracy of 68.07% in the subset with manipulated image/text.
% Moreover, we also compare the cosine similarity between the consistent text embedding and inconsistent text embedding (80.98\%) with the cosine similarity between the consistent text embedding and random Gaussian text embedding (0.05\%), this also explains why initializing with $E_0$ is better than random ones.
% \begin{minipage}[b]{\linewidth} \centering
% \scalebox{1}{ 

\begin{table}[t]
\centering
\setlength{\tabcolsep}{18pt} 
\begin{tabular}{lccccc}
\rowcolor{mygray}
   &  0.1  & 0.2 & 0.3  & 0.4 & Average (Ours) \\   \hline \thickhline
  mIoU (\%) & 39.66 &  42.81 & 43.23 &  42.66 & 47.50\\ %  4274 4325  4144
\hline
\end{tabular}
% \vskip -0.1in
\captionsetup{justification=centering}
\vspace{0.1cm}
\caption{\small \it Comparison of different mask thresholds on a subset of TIIL with 1,000 image-text pairs.}
% \vspace{-0.1cm}
\label{tab:thres}
\end{table}

\myheading{Mask Threshold.} We compare four fixed threshold strategies with our average-based method, which uses the average values among the mask as the threshold for $\mM^{\prime}$ and $\mM$.  
% This experiment is conducted on a randomly sampled subset of TIIL with 1000 image-text pairs. 
The results in \Tref{tab:thres} and \Fref{fig:thres} show that using a relatively smaller threshold results in a lower mIoU score and a larger predicted area, wherein the generated image $I_{edt}$ includes more "implicit" backgrounds. On the contrary, when using a larger threshold, there is an increase in the mIoU, but it can also lead to a smaller predicted inconsistency area. This is particularly evident when setting the threshold to 0.4, as depicted in \Fref{fig:thres}.
Our method surpasses the performance of fixed threshold strategies by utilizing an adaptive threshold for mask binarization.

\def\subFigSz{0.23\linewidth}
\def\subFigH{0.16\linewidth}
\captionsetup{type=figure} 
\begin{figure}[!ht]
\centering

\begin{minipage}[c]{1\linewidth}
    \includegraphics[width=\subFigSz, height=\subFigH]{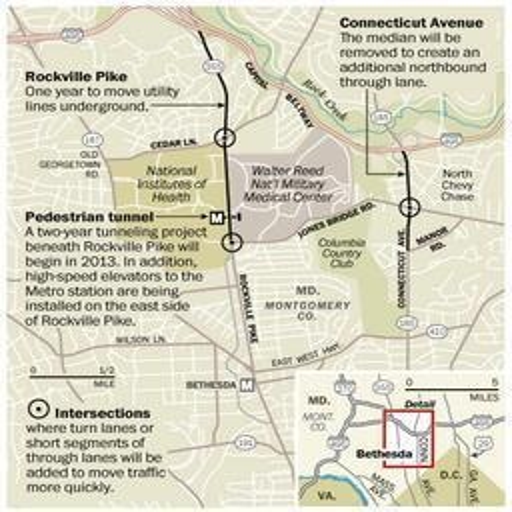}
    \includegraphics[width=\subFigSz, height=\subFigH]{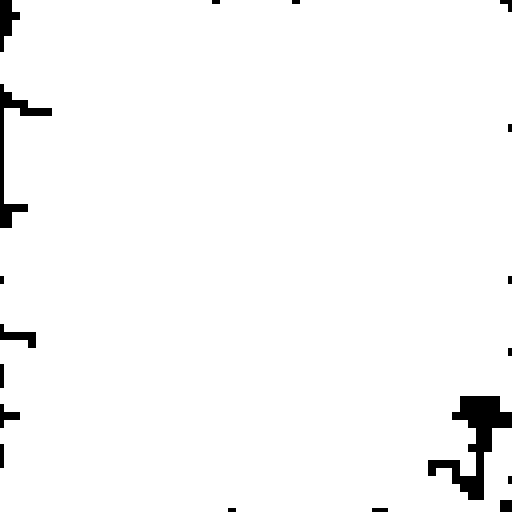}
    \hspace{0.02\linewidth}
    \includegraphics[width=\subFigSz, height=\subFigH]{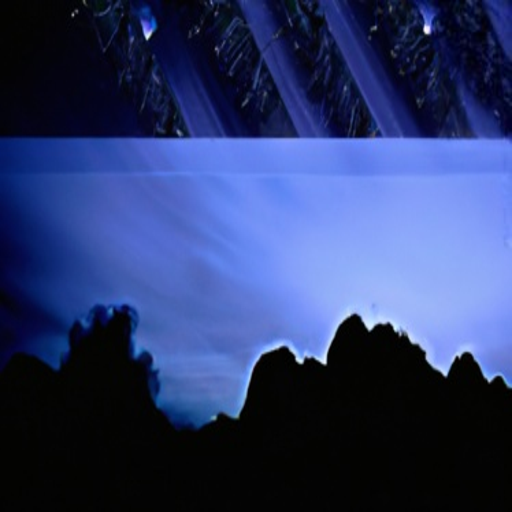}
    \includegraphics[width=\subFigSz, height=\subFigH]{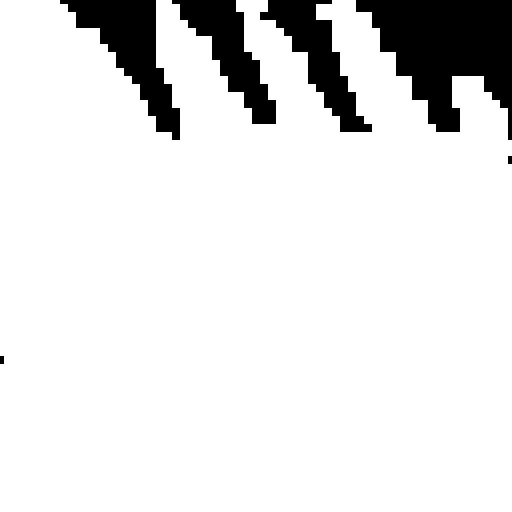}
\end{minipage}

\begin{minipage}[c]{0.45\linewidth}
    \caption*{\scriptsize \it ``A \textcolor{red}{oil painting} for Rockville Pike and Connecticut Avenue in Bethesda''}
\end{minipage}
\hspace{0.02\linewidth}
\begin{minipage}[c]{0.45\linewidth}
    \caption*{\scriptsize \it ``Icicles hang from the \textcolor{red}{caves} in northern Wisconsin''}
\end{minipage}

\begin{minipage}[c]{\linewidth}
\vspace{-0.45cm}
    \caption*{ \it (a) Completely inconsistent image descriptions}
\end{minipage}

\begin{minipage}[c]{1\linewidth}
    \includegraphics[width=\subFigSz, height=\subFigH]{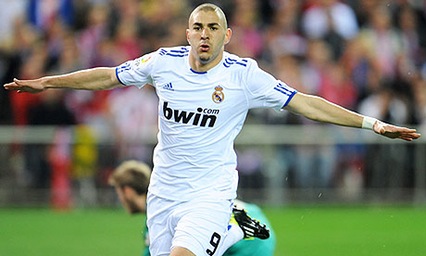}
    \includegraphics[width=\subFigSz, height=\subFigH]{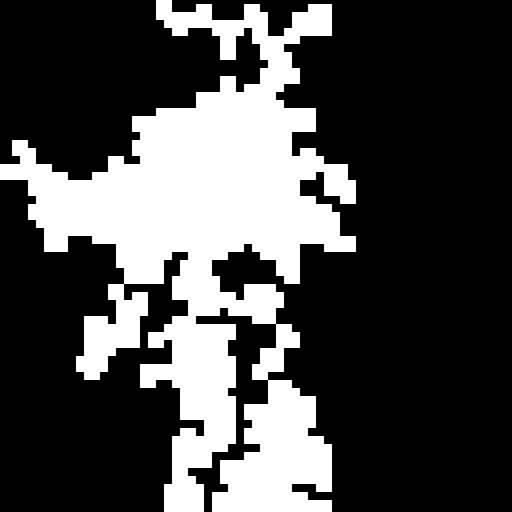}
    \hspace{0.02\linewidth}
    \includegraphics[width=\subFigSz, height=\subFigH]{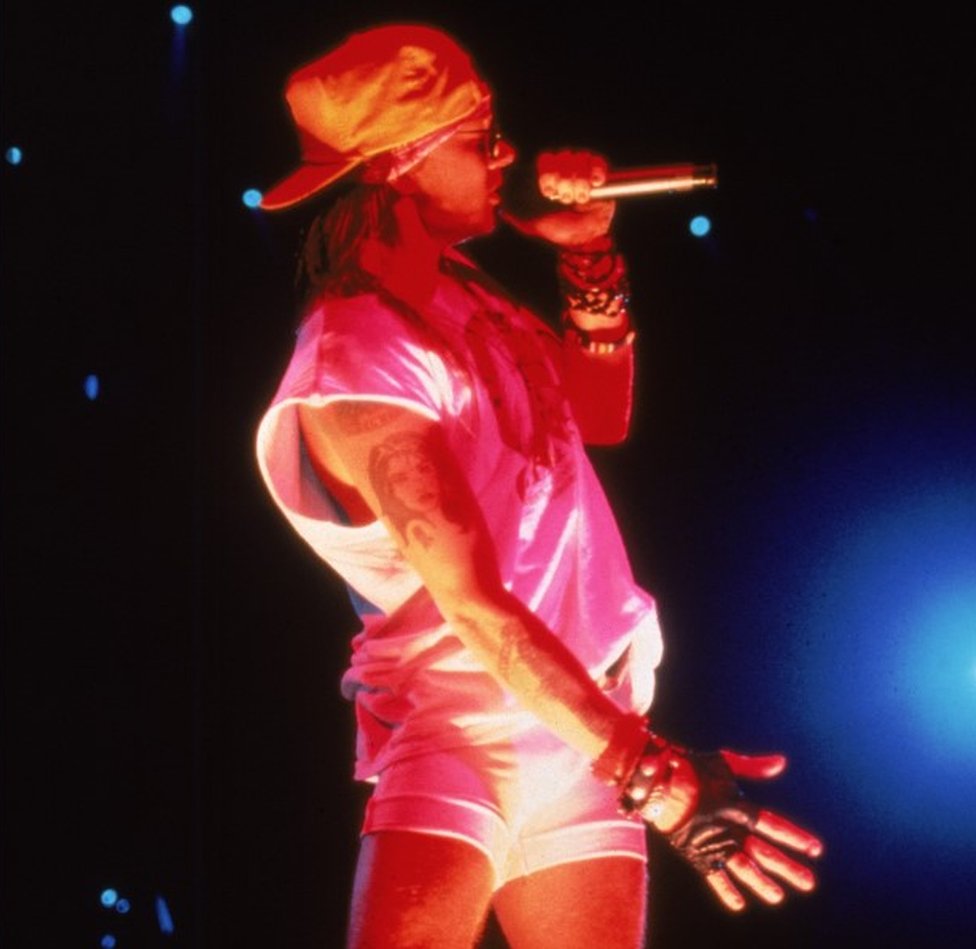}
    \includegraphics[width=\subFigSz, height=\subFigH]{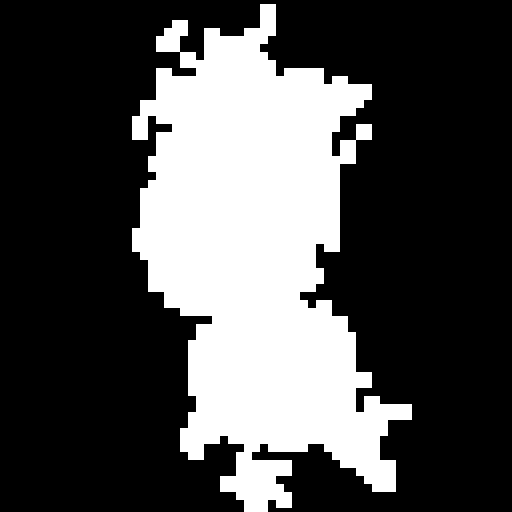}
\end{minipage}
\begin{minipage}[c]{0.45\linewidth}
    \caption*{\scriptsize \it ``\textcolor{red}{Karim Benzema from Atletico Madrid} celebrates his goal in the Madrid derby''}
\end{minipage}
\hspace{0.02\linewidth}
\begin{minipage}[c]{0.45\linewidth}
    \caption*{\scriptsize \it ``Roses \textcolor{red}{playing basketball} in October 1987''}
\end{minipage}
\vspace{-0.3cm}

\begin{minipage}[c]{1\linewidth}
    \includegraphics[width=\subFigSz, height=\subFigH]{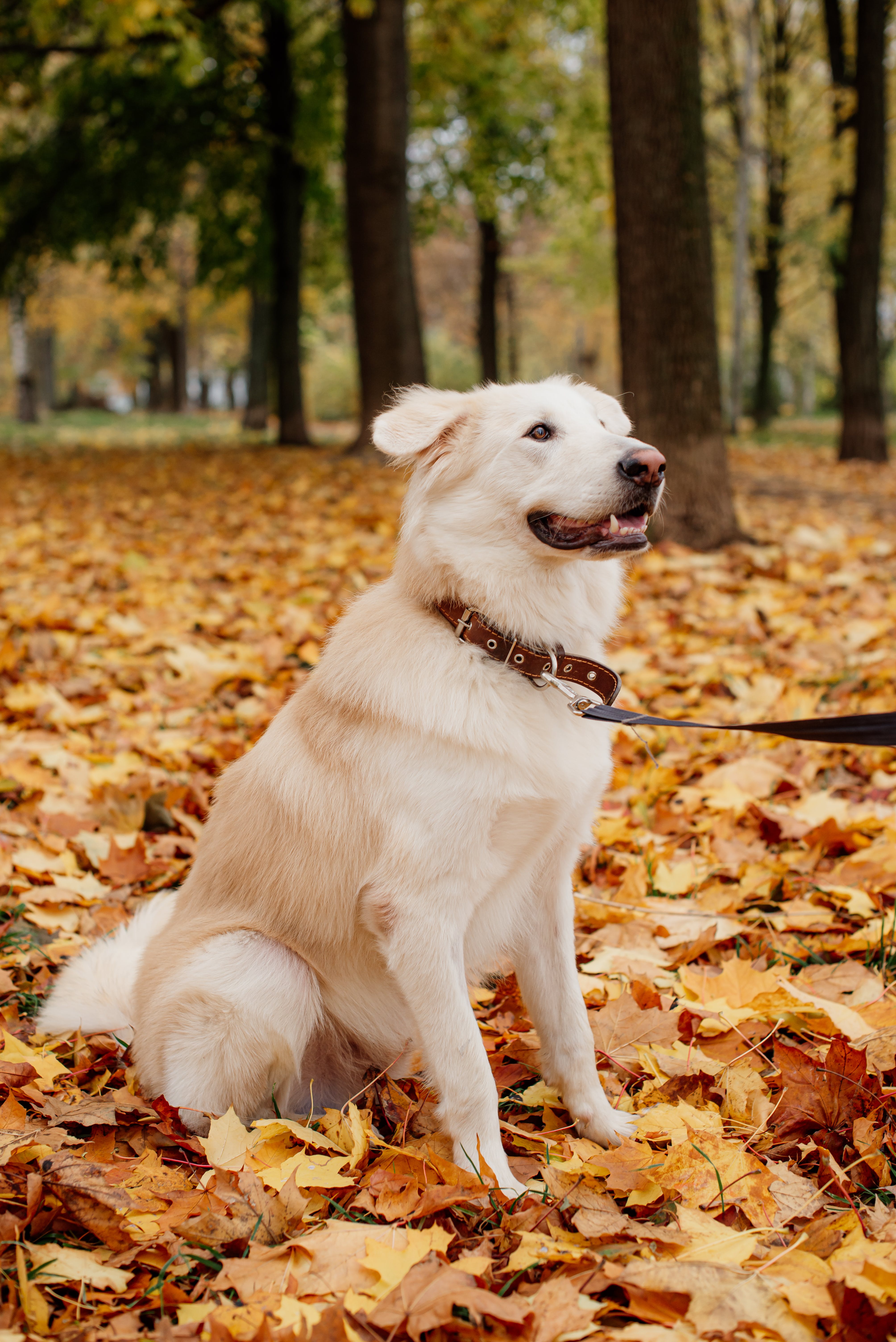}
    \includegraphics[width=\subFigSz, height=\subFigH]{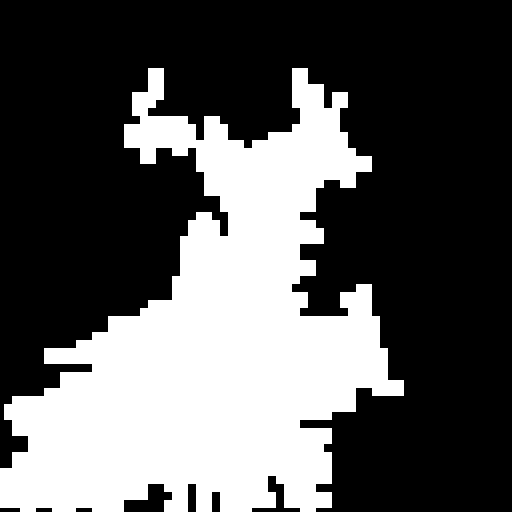}
    \hspace{0.02\linewidth}
    \includegraphics[width=\subFigSz, height=\subFigH]{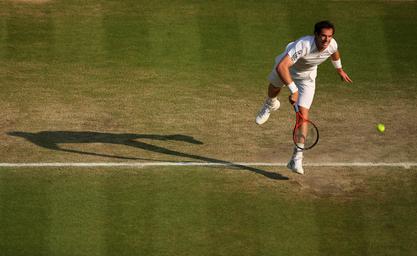}
    \includegraphics[width=\subFigSz, height=\subFigH]{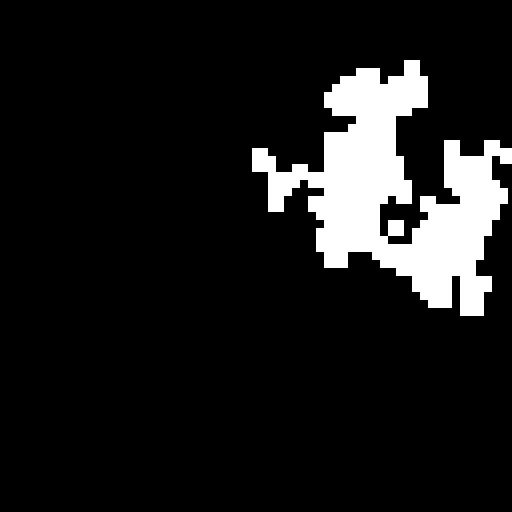}
\end{minipage}
\begin{minipage}[c]{0.45\linewidth}
    \caption*{\scriptsize \it ``A lovely dog plays \textcolor{red}{ball} on the autumn leaves''}
\end{minipage}
\hspace{0.02\linewidth}
\begin{minipage}[c]{0.45\linewidth}
    \caption*{\scriptsize \it ``Andy Murray kicking \textcolor{red}{a football} to Jerzy Janowicz during their semifinal match''}
\end{minipage}

\begin{minipage}[c]{\linewidth}
\vspace{-0.4cm}
    \caption*{ \it (b) Objects are aligned but attributes/predictes are not.}
\end{minipage}

\begin{minipage}[c]{1\linewidth}
    \includegraphics[width=\subFigSz, height=\subFigH]{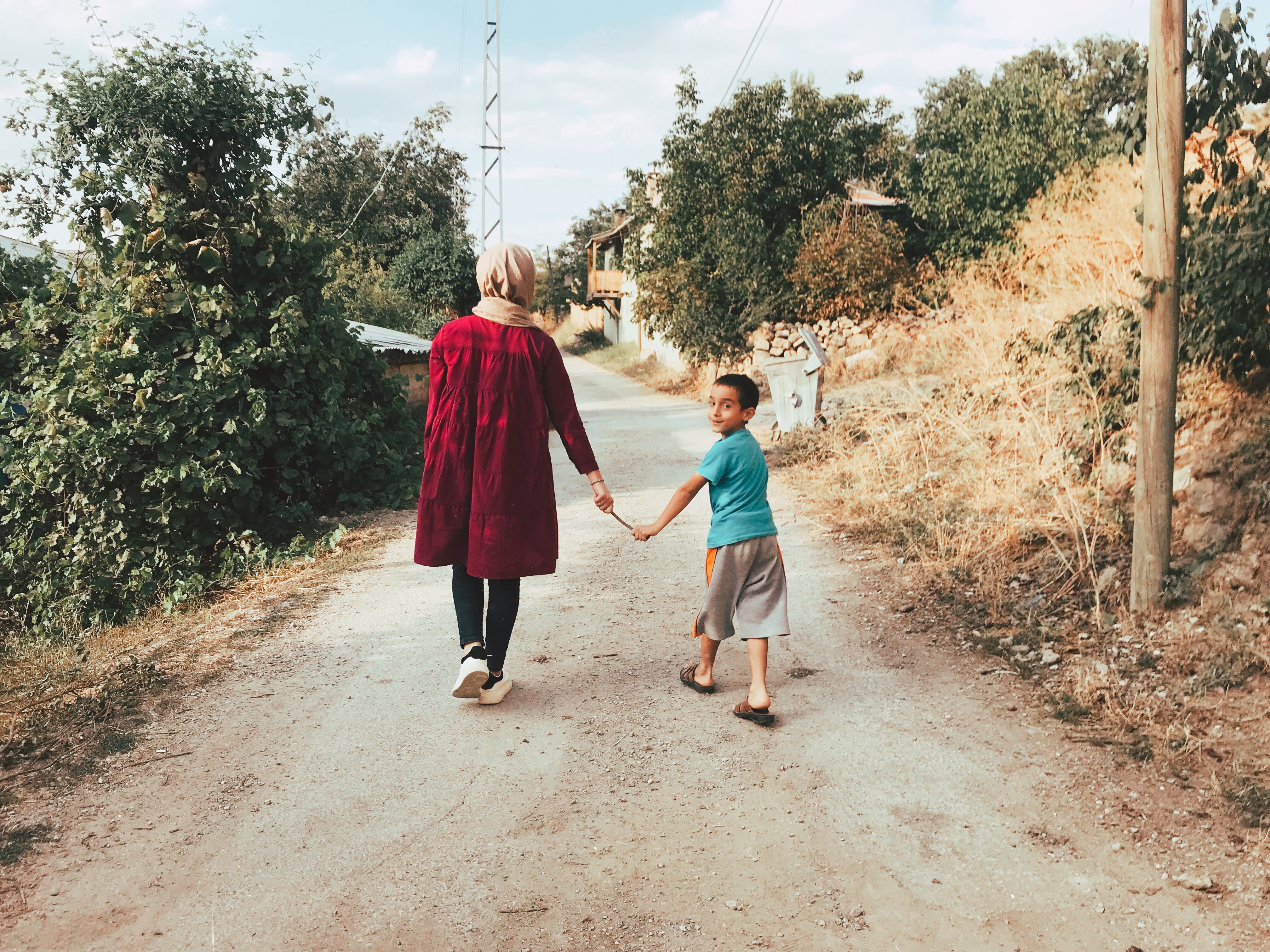}
    \includegraphics[width=\subFigSz, height=\subFigH]{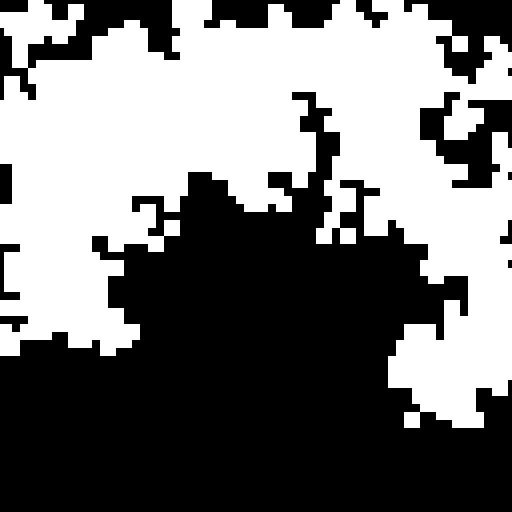}
    \hspace{0.02\linewidth}
    \includegraphics[width=\subFigSz, height=\subFigH]{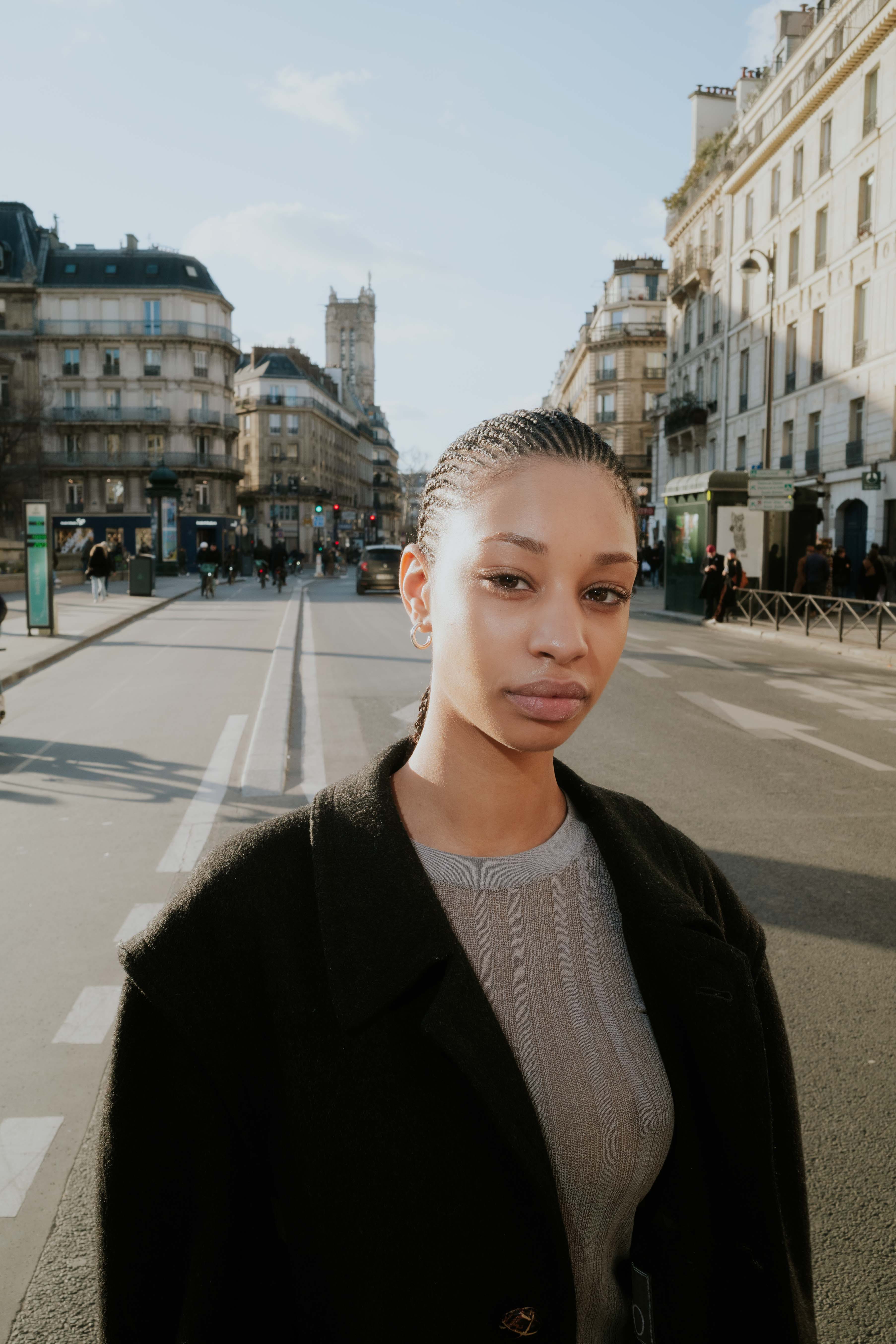}
    \includegraphics[width=\subFigSz, height=\subFigH]{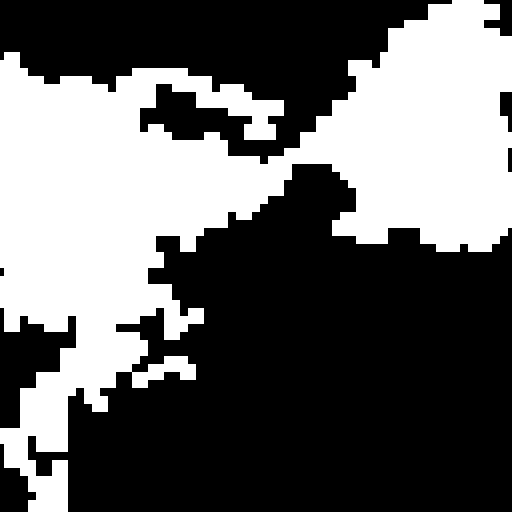}
\end{minipage}
\begin{minipage}[c]{0.45\linewidth}
    \caption*{\scriptsize \it ``Woman and boy holding a stick while walking through the \textcolor{red}{city}''}
\end{minipage}
\hspace{0.02\linewidth}
\begin{minipage}[c]{0.45\linewidth}
    \caption*{\scriptsize \it ``Portrait of woman on \textcolor{red}{the forest}''}
\end{minipage}

\begin{minipage}[c]{\linewidth}
\vspace{-0.4cm}
    \caption*{ \it (c) Backgrounds or scenarios are inconsistent.}
\end{minipage}

\caption{\em \small Additional examples from D-TIIL on real-world news image-text pairs. The detected inconsistent text is highlighted as \textcolor{red}{red}.} 
\label{fig:addi}
\vspace{-0.1in}
\end{figure}

{
\myheading{Additional examples.} We include additional examples to cover different scenarios of inconsistencies in \Fref{fig:addi}. \Fref{fig:addi} (a) shows the case that text and images are completely misaligned where the most area of the image is supposed to be masked; \Fref{fig:addi} (b) shows the case that the objects shared in the image and textual semantic space are well aligned but their attributes (e.g., actions, adjective) are inconsistent, the masks are supposed to cover the whole objects. \Fref{fig:addi} (c) contains more complex semantics inconsistent cases where the semantics inconsistency occurs in the background or the scene. }

{
\myheading{Analysis of the learned representation.}
D-TIIL has two alignment steps to iteratively align the image/text embeddings and filter out relevant semantic information with diffusion models. The well-aligned representations make it easier to identify and localize the semantic inconsistencies. Specifically, the learned representations are two parts, including $E_{aln}$ with aligned semantic space with the input image $I$, and $E_{dnt}$ with aligned semantic space with the input text embedding $E_0$. To show the effectiveness of our two alignment steps in learning the representation, we have provided the comparison of averaged cosine similarity scores on a subset of our dataset. We observed that the similarity between $E_{aln}$ and $I$ has increased by 9.2\% compared to the similarity between $E_0$ and $I$ after the first step alignment. The similarity between $E_{dnt}$ and $I_{edt}$ is increased by 2.4\% compared to the similarity between $E_0$ and $I_{edt}$ by the second step alignment. Note that 2.4\% is not subtle since it only comes from the exclusion of distracting semantics from text embeddings.

}

\end{document}